\def\year{2022}\relax
\documentclass[letterpaper]{article} 
\usepackage{aaai22}  
\usepackage{times}  
\usepackage{helvet}  
\usepackage{courier}  
\usepackage[hyphens]{url}  
\usepackage{graphicx} 
\def\UrlFont{\rm}  
\usepackage{natbib}  
\usepackage{caption} 
\DeclareCaptionStyle{ruled}{labelfont=normalfont,labelsep=colon,strut=off} 
\usepackage{algorithm}
\usepackage{algorithmic}

\usepackage{newfloat}
\usepackage{listings}
\floatstyle{ruled}
\newfloat{listing}{tb}{lst}{}
\floatname{listing}{Listing}
\usepackage{comment}
\usepackage[utf8]{inputenc} 
\usepackage[T1]{fontenc}    
\usepackage{hyperref}       
\usepackage{url}            
\usepackage{booktabs}       
\usepackage{amsfonts}       
\usepackage{nicefrac}       
\usepackage{microtype}      
\usepackage{appendix}
\usepackage{microtype}      
\usepackage{xcolor}         

\usepackage{svg}
\usepackage{graphicx}
\usepackage{subfigure}
\usepackage{amsmath}
\usepackage{amsthm}
\usepackage{mathrsfs}
\usepackage{algorithm}
\usepackage{algorithmic}
\usepackage{amssymb}

\DeclareMathOperator*{\argmax}{arg\,max}
\DeclareMathOperator*{\argmin}{arg\,min}

\nocopyright

\title{A Review for Deep Reinforcement Learning in Atari: \\
Benchmarks, Challenges, and Solutions}
\author {
    Jiajun Fan
}
\affiliations{
    Tsinghua Shenzhen International Graduate School\\
    Tsinghua Univsersity,Shenzhen, China\\
    fanjj21@mails.tsinghua.edu.cn
}

\def\GDIHmeanhns{9620.98}
\def\GDIHmeanhnsle{4.81E-07}

\def\GDIHmedianhns{1146.39}
\def\GDIHmedianhnsle{5.73E-08}

\def\GDIHHWRB{22}
\def\GDIHHWRBle{1.10E-07}

\def\GDIHmeanHWRNS{154.27}
\def\GDIHmeanHWRNSle{7.71E-09}
\def\GDIHmedianHWRNS{50.63}
\def\GDIHmedianHWRNSle{2.53E-09}

\def\GDIHmeanSABER{71.26}
\def\GDIHmeanSABERle{3.56E-09}

\def\GDIHmedianSABER{50.63}
\def\GDIHmedianSABERle{2.53E-09}

\def\GDIHnumframes{2E+8}
\def\GDIHgametime{0.114}
\def\GDIImeanhns{7810.6}
\def\GDIImedianhns{832.5}
\def\GDIIHWRB{17}
\def\GDIImeanHWRNS{117.99}
\def\GDIImedianHWRNS{35.78}
\def\GDIImeanSABER{61.66}
\def\GDIImedianSABER{35.78}
\def\GDIInumframes{2E+8}
\def\GDIIgametime{0.114}

\newcommand{\best}[1]{\textbf{#1}}

\usepackage{bibentry}

\begin{document}
\maketitle

\begin{abstract}
The Arcade Learning Environment (ALE) is proposed as an evaluation platform for empirically assessing the generality of agents across dozens of Atari 2600 games. ALE offers various challenging problems and has drawn significant attention from the deep reinforcement learning (RL) community. From Deep Q-Networks (DQN) to Agent57, RL agents seem to achieve superhuman performance in ALE. However, is this the case? In this paper, to explore this problem, we first review the current evaluation metrics in the Atari benchmarks and then reveal that the current evaluation criteria of achieving superhuman performance are inappropriate, which underestimated the human performance relative to what is possible. To handle those problems and promote the development of RL research, we propose a novel Atari benchmark based on human world records (HWR), which puts forward higher requirements for RL agents on both final performance and learning efficiency.
Furthermore, we summarize the state-of-the-art (SOTA) methods in Atari benchmarks and provide benchmark results over new evaluation metrics based on human world records. We concluded that at least four open challenges hinder RL agents from achieving superhuman performance from those new benchmark results. Finally, we also discuss some promising ways to handle those problems. 
\end{abstract}

\section{Introduction}

The Arcade Learning Environment \citep[ALE]{ale} was proposed as a platform for empirically assessing agents designed for general competency across a wide range of Atari games. ALE offers an interface to a diverse set of Atari 2600 game environments designed to be engaging and challenging for human players \citep{atarihuman}. Most games of Atari have not been entirely conquered by humans, making the human world records breakthrough a symbol of RL agents to achieve superhuman performance. As \citet{ale} put it, the Atari 2600 games are well suited for evaluating general competency in AI agents for three main reasons: 

\begin{enumerate}
    \item ALE provides multiple different tasks, requiring enough generality.
    \item As some of ALE has not been broken through, and they are also challenging for humans.
    \item Developed by an independent party, ALE is free of the experimenter’s bias.
\end{enumerate}


Agents are expected to perform well in as many games as possible \textbf{making minimal assumptions about the domain at hand and without the use of game-specific information}. In recent reinforcement learning advances, researchers \citep{agent57,muesli,rainbow} are seeking agents that can achieve superhuman performance. Deep Q-Networks \citep[DQN]{dqn} was the first algorithm to achieve human-level control in a large number of the Atari 2600 games, measured by human normalized scores (HNS). Subsequently, using HNS to assess performance on Atari games has become one of the most widely used benchmarks in deep reinforcement learning (RL). Current state-of-the-art(SOTA) algorithms such as \citep[Agent]{agent57} claimed that they had achieved superhuman performance when they outperformed the human baseline uniformly over all Atari 57 games.

It seems that reinforcement learning agents have been able to reach the superhuman level. However, is this the case? In this paper, we argue that the performance of current reinforcement learning agents is far from the superhuman level from several aspects.

As \citet{atarihuman} put it, the human baseline scores potentially \textbf{underestimating human performance} relative to what is possible. Thus, we argue that this human baseline is far from representative of the best human player, which means that using it to claim superhuman performance is misleading. This paper will propose more comprehensive and reasonable evaluation metrics for the Atari benchmark to test the real superhuman reinforcement learning algorithms.

Learning efficiency \citep{ale2} is one of the metrics to evaluate the learning ability of RL agents. However, many SOTA algorithms (e.g., \citep{agent57}) always overemphasize the final performance they obtain but ignore the computational cost required to obtain that performance. It has led more algorithms to improve the final performance at the expense of more training samples. This paper argues that a superhuman agent should surpass humans in both final performance and learning efficiency. In this paper, we will propose several measures to evaluate the learning efficiency of RL agents.

In this work, we first discuss current evaluation metrics in Atari benchmarks. We then propose a more comprehensive evaluation system, which perfects the work of \citet{atarihuman}. We advocate introducing the human world records baseline into the evaluation system of Atari benchmarks. As an illustration of our new benchmark, we provide benchmark results for several representative algorithms in model-free RL, model-based RL, and other fields. From those benchmark results, we find out that SOTA RL algorithms are far from superhuman performance on ALE, which means ALE is still a challenging problem. Finally, we conclude the challenges for obtaining superhuman agents in ALE and propose some promising solutions.

The main contributions of this work are:

\begin{enumerate}
    \item \textbf{Review of Evaluation Metrics for Atari Benchmark.} We reviewed the most used evaluation metrics for ALE and thoroughly discussed the advantages and disadvantages while using those metrics.
    \item \textbf{Introduction of the measure of learning efficiency for RL agents on ALE.} We argue the importance of learning efficiency for superhuman RL agents and revealed the low learning efficiency problem for current SOTA algorithms \citep{agent57,muzero}.
    \item \textbf{Perfection of the world records human baseline.} We provide complete human world records overall the 57 Atari games, rather than part of them \citep{dreamerv2,atarihuman}. We further extended the SABER \citep{atarihuman} to a more comprehensive evaluation system with several new evaluation metrics based on human world records.
    \item \textbf{The proposal, description, and justification of a superhuman benchmark for ALE.} We argue that the human world records are more representative for the human level instead of human baselines used in most of the previous works. 
    \item \textbf{A novel benchmark results of current state-of-the-art reinforcement learning algorithms.} We review several milestones in the Atari benchmarks from DQN to GDI \citep{fan2021gdi}, and then show the benchmark results of them. From these new benchmark results, we see the ALE is still challenging even for so-called SOTA algorithms. 
    \item \textbf{Introduction of current challenges and promising solutions for superhuman agents on ALE. } From the new benchmark results, we see massive existing problems hindering RL agents from achieving superhuman performance. We conclude those problems and provide promising solutions for those problems. 
\end{enumerate}

\section{Background}
Similar to deep learning \citep{cl,actionr,6d}, reinforcement learning is also a branch of machine learning.  Most of the previous work has introduced the background knowledge of RL in detail. In this section, we only summarize and recall the background knowledge used in this paper. If you are interested in the relevant content, we recommend you read the relevant material \citep{sutton,trrg,eerr,serl}. 

\subsection{Reinforcement Learning}
 The RL problem can be formulated as a Markov Decision Process \citep[MDP]{howard1960dynamic} defined by $\left(\mathcal{S}, \mathcal{A}, p, r, \gamma, \rho_{0}\right)$. 
 Considering a discounted episodic MDP, the initial state $s_0$ is sampled from the initial distribution $\rho_0(s): \mathcal{S} \rightarrow \Delta(\mathcal{S})$, where we use $\Delta$ to represent the probability simplex.
 At each time $t$, the agent chooses an action $a_t \in \mathcal{A}$ according to the policy $\pi(a_t|s_t): \mathcal{S} \rightarrow \Delta(\mathcal{A})$ at state $s_t \in \mathcal{S}$. 
 The environment receives $a_t$, produces the reward $r_t \sim r(s,a): \mathcal{S} \times \mathcal{A} \rightarrow \mathbf{R}$ and transfers to the next state $s_{t+1}$  according to the transition distribution $p\left(s^{\prime} \mid s, a\right): \mathcal{S} \times \mathcal{A} \rightarrow \Delta(\mathcal{S})$. 
 The process continues until the agent reaches a terminal state or a maximum time step. 
 Define the discounted state visitation distribution as 
 $d_{\rho_0}^{\pi} (s) = (1 - \gamma) \textbf{E}_{s_0 \sim \rho_0} 
 \left[ \sum_{t=0}^{\infty} \gamma^t \textbf{P} (s_t = s | s_0) \right]$.
 The goal of reinforcement learning is to find the optimal policy $\pi^*$ that maximizes the expected sum of discounted rewards, denoted by $\mathcal{J}$ \citep{sutton}:
\begin{equation}
\label{eq_accmulate_reward}
\begin{aligned}
    \pi^{*}
&=\underset{\pi}{\operatorname{argmax}} \mathcal{J}_{\pi} \\
&= \underset{\pi}{\operatorname{argmax}} \textbf{E}_{s_t \sim d_{\rho_0}^{\pi}} \textbf{E}_{\pi}\left[G_t | s_t \right] \\
&= \underset{\pi}{\operatorname{argmax}} \textbf{E}_{s_t \sim d_{\rho_0}^{\pi}} \textbf{E}_{\pi} \left[\sum_{k=0}^{\infty} \gamma^{k} r_{t+k} | s_t \right]
\end{aligned}
\end{equation}
where $\gamma \in(0,1)$ is the discount factor.

RL algorithms can be divided into off-policy manners \citep{dqn,a3c,sac,impala} and on-policy manners \citep{ppo}. 
Off-policy algorithms select actions according to a behavior policy $\mu$ that differs from the learning policy $\pi$.
On-policy algorithms evaluate and improve the learning policy through data sampled from the same policy.
RL algorithms can also be divided into value-based methods \citep{dqn,rainbow,apex} and policy-based methods \citep{impala,laser}. 
In the value-based methods, agents learn the policy indirectly, where the policy is defined by consulting the learned value function, like $\epsilon$-greedy, and a typical GPI learns the value function.
In the policy-based methods, agents learn the policy directly, where the correctness of the gradient direction is guaranteed by the policy gradient theorem \citep{sutton}, and the convergence of the policy gradient methods is also guaranteed \citep{pgtheory}.

\section{Evaluation Metrics for ALE}

In this section, we will mainly introduce the evaluation metrics in ALE, including those that have been commonly used by previous works like the raw score and the normalized score over all the Atari games based on human average score baseline, and some novel evaluation criteria for the superhuman Atari benchmark such as the normalized score based on human world records, learning efficiency, and human world record breakthrough.

\subsection{Raw Score}

Raw score refers to using tables (e.g., Table of Scores) or figures (e.g., Training Curve) to show the total scores of RL algorithms on all Atari games, which can be calculated by the sum of the undiscounted reward of the \textbf{g}th game of Atari using algorithm \textbf{i} as follows:

\begin{equation}
\label{equ: raw score}
G_{g,i} =  \textbf{E}_{s_t \sim d_{\rho_0}^{\pi}} \textbf{E}_{\pi} \left[\sum_{k=0}^{\infty} r_{t+k} | s_t \right], g \in [1,57]
\end{equation}

As \citet{ale} firstly put it, raw score over the whole 57 Atari games can reflect the performance and generality of RL agents to a certain extent. However, this evaluation metric has many limitations:

\begin{enumerate}
    \item It is difficult to compare the performance of the two algorithms directly.
    \item Its value is easily affected by the score scale. For example, the score scale of Pong is [-21,21], but that of chopper command is [0,999900], so the chopper command will dominate the mean score of those games.
\end{enumerate}

In recent RL advances, this metric is used to avoid any issues that aggregated metrics may have \citep{agent57,fan2021gdi,fan2020critic,CASA-B,casa_bridge,casa_entropy}. Furthermore, this paper used these metrics to prove whether the RL agents have surpassed the human world records, which will be introduced in detail later.

\subsection{Normalized Scores}
To handle the drawbacks of the raw score, some researchers \citep{ale,dqn} proposed the normalized score. The normalized score of the \textbf{g}th game of Atari using algorithm \textbf{i} can be calculated as follows:
\begin{equation}
\label{equ: Normalized Score}
    Z_{g,i} = \frac{G_{g,i}-G_{g,base}}{G_{g,reference}-G_{g,base}}
\end{equation}
As \citet{ale} put it, we can compare games with different scoring scales by  normalizing scores, which makes the numerical values become comparable. In practice, we can make $G_{g,base} = r_{g,min}$ and $G_{g,reference} = r_{g,max}$, where $[r_{g,min},r_{g,max}]$ is the score scale of the \textbf{g}th game. Then Equ. \eqref{equ: Normalized Score} becomes $Z_{g,i} = \frac{G_{g,i}-r_{g,min}}{r_{i,max}-r_{g,min}}$, which is a \textbf{Min-Max Scaling} and thus $Z_{g,i} \in [0,1]$ become comparable across the 57 games. It seems this metric can be served to compare the performance between two different algorithms. However, the Min-Max normalized score fail to intuitively reflect the gap between the algorithm and the average level of humans. Thus, we need a human baseline normalized score.

\subsubsection{Human Average Score Baseline}

As we mentioned above, recent reinforcement learning advances \citep{agent57,ngu,r2d2,goexplore,muzero,muesli,rainbow} are seeking agents that can achieve superhuman performance. Thus, we need a metric to intuitively reflect the level of the algorithms compared to human performance. Since being proposed by \citep{ale}, the Human Normalized Score (HNS) is widely used in the RL research\citep{ale2}. HNS can be calculated as follows:

\begin{equation}
     \text{HNS}_{g,i} = \frac{G_{g,i}-G_{g,\text{random}}}{G_{g,\text{human average}}-G_{g,\text{random}}}
\end{equation}
wherein g donates the gth game of Atari, i represents the algorithm i, $G_{g,\text{human average}}$ represents the human average score baseline \citep{atarihuman}, and $G_{g,\text{random}}$ represents the performance of a random policy. Adopting HNS as an evaluation metric has the following advantages:

\begin{enumerate}
    \item \textbf{Intuitive comparison with human performance.} $\text{HNS}_{g,i} \geq 100\%$ means algorithm i have surpassed the human average performance in game g. Therefore, we can directly use HNS to reflect which games the RL agents have surpassed the average human performance. 
    \item \textbf{Performance across algorithms become comparable.} Like Max-Min Scaling, the human normalized score can also make two different algorithms comparable. The value of $\text{HNS}_{g,i}$ represents the degree to which algorithm i surpasses the average level of humans in game g.
\end{enumerate}

\textbf{Mean HNS} represents the mean performance of the algorithms across the 57 Atari games  based on the human average score. However, it is susceptible to interference from individual high-scoring games like the hard-exploration problems in Atari \citep{goexplore}. While taking it as the only evaluation metric, Go-Explore\citep{goexplore} has achieved SOTA compared to Agent57\citep{agent57}, NGU\citep{ngu}, R2D2\citep{r2d2}. However, Go-Explore fails to handle many other games in Atari like demon attack, breakout, boxing, phoenix \citep{fan2021gdi}. Additionally, Go-Explore fails to balance the trade-off between exploration and exploitation, which makes it suffer from the low sample efficiency problem, which will be discussed later.

\textbf{Median HNS} represents the median performance of the algorithms across the 57 Atari games based on the human average score. Massive researchers \citep{muzero,muesli} have adopted it as a more reasonable metric for comparing performance between different algorithms. The median HNS has overcome the interference from individual high-scoring games. However, As far as we can see, there are at least two problems while only referring to it as the evaluation metrics. First of all, the median HNS only represents the mediocre performance of an algorithm. How about the top performance? One algorithm \citep{muesli} can easily achieve high median HNS, but at the same time obtain a poor mean HNS by adjusting the hyperparameters of algorithms for games near the median score. It shows that these metrics can show the generality of the algorithms but fail to reflect the algorithm's potential. Moreover, adopting these metrics will urge us to pursue rather mediocre methods.

In practice, we often use \textbf{mean HNS} or \textbf{median HNS} to show the final performance or generality of an algorithm. Dispute upon whether the mean value or the median value is more representative to show the generality and performance of the algorithms lasts for several years \citep{dqn,rainbow,dreamerv2,muesli,ale,ale2}. To avoid any issues that aggregated metrics may have, \textbf{we advocate calculating both of them in the final results} because they serve different purposes, and we could not evaluate any algorithm via a single one of them.

\subsubsection{Capped Normalized Score}
Capped Normalized Score is also widely used in many reinforcement learning advances \citep{atarihuman,agent57}. Among them, Agent57 \citep{agent57} adopts the capped human normalized score (CHNS) as a better descriptor for evaluating general performance, which can be calculated as $\mathrm{CHNS}=\max \{\min \{\mathrm{HNS}, 1\}, 0\}$. Agent57 claimed CHNS emphasizes the games that are below the average human performance benchmark and used CHNS to judge whether an algorithm has surpassed the human performance via $\mathrm{CHNS} \geq 100\%$.  The mean/median CHNS represents the mean/median completeness of surpassing human performance. However, there are several problems while adopting 
these metrics:
\begin{enumerate}
    \item CHNS fails to reflect the real performance in specific games. For example, $\mathrm{CHNS} \geq 100\%$ represents the algorithms surpassed the human performance but failed to reveal how good the algorithm is in this game. From the view of CHNS, Agent57 \citep{agent57} has achieved SOTA performance across 57 Atari games, but while referring to the mean HNS or median HNS, Agent57 lost to MuZero \citep{fan2021gdi}.
    \item It is still controversial that using $\mathrm{CHNS} \geq 100\%$ to represent the superhuman performance because it underestimates the human performance \citep{atarihuman}.
    \item CHNS ignores the low sample efficiency problem as other metrics using normalized scores.
\end{enumerate}

In practice, CHNS can serve as an indicator to reflect whether RL agents can surpass the average human performance. The mean/median CHNS can be used to reflect the generality of the algorithms.

\subsubsection{Human World Records Baseline}
As \citep{atarihuman} put it, the Human Average Score Baseline potentially underestimates human performance relative to what is possible. To better reflect the performance of the algorithm compared to the human world record, we introduced a complete human world record baseline extended from \citep{dreamerv2,atarihuman} to normalize the raw score, which is called the Human World Records Normalized Score (HWRNS), which can be calculated as follows:

\begin{equation}
     \text{HWRNS}_{g,i} = \frac{G_{g,i}-G_{g,\text{random}}}{G_{g,\text{human world records}}-G_{g,\text{random}}}
\end{equation}
wherein g donates the gth game of Atari, i represents the RL algorithm, $G_{i,human}$ represents the human world records, and $G_{g,random}$ represents means the performance of a random policy. Adopting HWRNS as an evaluation metric of algorithm performance has the following advantages:

\begin{enumerate}
    \item \textbf{Intuitive comparison with human world records.} As $\text{HNS}_{g,i} \geq 100\%$ means algorithm i have surpassed the human world records performance in game g. We can directly use HWRNS to reflect which games the RL agents have surpassed the human world records, which can be used to calculate the human world records breakthrough in Atari benchmarks.
    \item \textbf{Performance across algorithms become comparable.} Like the Max-Min Scaling, the HWRNS can also make two different algorithms comparable. The value of $\text{HWRNS}_{g,i}$ represents the degree to which algorithm i has surpassed the human world records in game g.
\end{enumerate}

\textbf{Mean HWRNS} represents the mean performance of the algorithms across the 57 Atari games. Compared to mean HNS, mean HWRNS put forward higher requirements on the algorithm. Poor performance algorithms like SimPLe \citep{modelbasedatari} will can be directly distinguished from other algorithms. It requires the algorithms to pursue a better performance across all the games rather than concentrate on one or two of them because breaking through any human world record is a huge milestone, which puts forward significant challenges to the performance and generality of the algorithm. For example, current model-free SOTA algorithms on HNS is Agent57 \citep{agent57}, which only acquires 125.92\% mean HWRNS, while GDI-H$^3$ obtained 154.27\% mean HWRNS and thus became the new state-of-the-art.

\textbf{Median HWRNS} represents the median performance of the algorithms across the 57 Atari games. Compared to Median HNS, median HWRNS also puts forward higher requirements for the algorithm. For example, current SOTA RL algorithms like Muzero \citep{muzero} obtain much higher median HNS over GDI-H$^3$ \citep{fan2021gdi} but relatively lower median HWRNS.

\textbf{Capped HWRNS} Capped HWRNS (also called SABER) is firstly proposed and used by \citep{atarihuman}, which is calculated by $\mathrm{SABER}=\max \{\min \{\mathrm{HWRNS}, 2\}, 0\}$. SABER also has the same problems as CHNS, and we will not repeat them here. For more details on SABER, we recommend referring \citep{atarihuman}.

\subsection{Learning Efficiency}
As we mentioned above, traditional SOTA algorithms typically ignore the low learning efficiency problem, which makes the data used for training continuously increasing (e.g., from 10B \citep{r2d2} to 100B \citep{agent57}). Increasing the training volume hinders the application of reinforcement learning algorithms into the real world. In this paper, we advocate not to improve the final performance via improving the learning efficiency instead of increasing the training volume. We advocate achieving SOTA within 200M training frames for Atari. To evaluate the learning efficiency of an algorithm, we introduce three promising metrics.

\subsubsection{Training Scale}
As one of the commonly used metrics to reveal the learning efficiency for machine learning algorithms, training scale can also serve the purpose in RL problems. In ALE, the training scale means the scale of video frames used for training. Training frames for world modeling or planning via real-world models also need to be counted in model-based settings.

\subsubsection{Game Time}
Game time is a unique metric of Atari, which means the real-time gameplay \citep{ale2,fan2021gdi}. We can use the following formula to calculate this metric:
\begin{equation}
    \text{Game Time (day)} = \frac{\text{Num.Frames}}{\text{108000*2*24}}
\end{equation}
For example, 200M training frames equal to 38.5 days real-time gameplay \citep{fan2021gdi}, and 100B training frames equal to 19250 days (52.7 years) real-time gameplay \citep{agent57}. As far as we know, no Atari human world record was achieved by playing a game continuously for more than 52.7 years because it is less than 52.7 years since the birth of the Atari games.

\subsubsection{Learning Efficiency}
As we mentioned several times while discussing the drawbacks of the normalized score, learning efficiency has been ignored in massive SOTA algorithms. Many SOTA algorithms achieved SOTA through training with vast amounts of data, which may equal 52.7 years continuously playing for a human. In this paper, we argue it is unreasonable to rely on the increase of data to improve the algorithm's performance. Thus, we proposed the following metric to evaluate the learning efficiency of an algorithm: 
\begin{equation}
    \text{Learning Efficiency}=\frac{\text{Related Evaluation Metric}}{\text{Num.Frames}}
\end{equation}
For example, the learning efficiency of an algorithm over means HNS is $\frac{\text{mean HNS}}{\text{Num.Frames}}$, which means the algorithms obtaining higher mean HNS via lower training frames are better than those acquiring more training data methods.
\subsection{Human World Record Breakthrough}
As we mentioned above, we need higher requirements to prove RL agents achieve real superhuman performance. Therefore, like the CHNS \citep{agent57}, the Human World Record Breakthrough (HWRB) can serve as the metric to reveal whether the algorithm has achieved the real superhuman performance, which can be calculated by $\text{HWRB} =\sum_{i=1}^{57}(\text{HWRNS} \geq 1) $.

\section{Human World Records Benchmark for Reinforcement Learning on Atari}
Since we have thoroughly discussed the evaluation metrics in ALE, in this section, we mainly introduce the Human World Records Benchmark for Reinforcement Learning on Atari. Firstly, we will discuss some methodological differences in ALE benchmarks found in the literature \citep{ale,ale2,agent57,rainbow}. Then, we will introduce the training and evaluation procedures. In the next section, we will report the benchmark results among representative reinforcement learning algorithms.
\subsection{Methodological Differences in ALE Benchmarks}
\subsubsection{Episode Termination}
In the origin benchmark of ALE \citep{ale}, episodes terminate when all the lives of the player are lost. Nevertheless, some articles \citep{dqn,rainbow} will end a training episode after every loss of life for training while ending an episode after losing all lives for testing. It helps the agent to value their lives more and learn to avoid death. We argue that is a kind of game-specific knowledge, which should not be concluded in the benchmark for ALE. As \citep{ale2} put it, \textit{we also advocate to use the game over signal be used for termination}.

\subsubsection{Maximum Episode Length}
Several related work \citep{atarihuman} also noticed that this setting of ALE would affect the results of the algorithms. Maximum Episode Length means the maximum number of frames
allowed per episode. This parameter ends the episode after a fixed number of time steps even if the game is not over. In most advance in RL \citep{agent57,fan2021gdi,r2d2,ngu}, this parameter has been set to 30min (equal to 1E+5 frames), while that in \citep{ale2}  is set to 5min. To put forward higher requirements on learning efficiency of methods, \textit{we advocate to use 30min as the maximum episode length}, which not only require the agents to find an optimal solution of the game but also require it to acquire the optimal score as soon as possible. We argue that the proposal of no maximum episode length \citep{atarihuman} is unreasonable because some games like Kangaroo will never stop being tracked in a circle.

\subsubsection{Action Set}
In the ALE, there are two sets of actions for each game, namely the useful set and the full set. Instead of using the useful set consisting of 4 actions that have been used in massive works \citep{dqn,rainbow}, \textit{we advocate to use the full set of actions which consists of 18 actions.}

\subsection{Training and Evaluation Procedures}
As recommended by \citep{ale2,atarihuman}, we adopt the same settings in both training and evaluations, which is more realistic.

\subsubsection{Training Procedures}

As we mentioned above, in the training phase, \textit{we advocate to use at most 200M frames and end an episode when all the lives are lost or the episode exceeds 30min}. Inside an episode, the agent should select a proper action from the full action set.

\subsubsection{Evaluation Procedures}
Except for the same setting as training, in evaluation procedures, \textit{we advocate to record the training score by averaging k consecutive episodes across the whole training.}

\subsubsection{Reporting performance}
As recommended by \citep{ale2}, we also advocate reporting the training score calculated by averaging k consecutive episodes across the whole training. It gives more information about the stability of the training and removes the statistical bias induced when reporting the score of the best policy, which is today a common practice \citep{rainbow,dqn,agent57,ngu}. \textit{Except for the HNS, we advocate that use evaluation metrics based on human world records like HWRB, HWRNS, SABER should be included in the final performance, and at the same time, the learning efficiency should also be considered while evaluating whether the RL agents have achieved superhuman performance.}

\begin{figure*}[!t]
    \centering
	\subfigure{
		\includegraphics[width=0.45\textwidth]{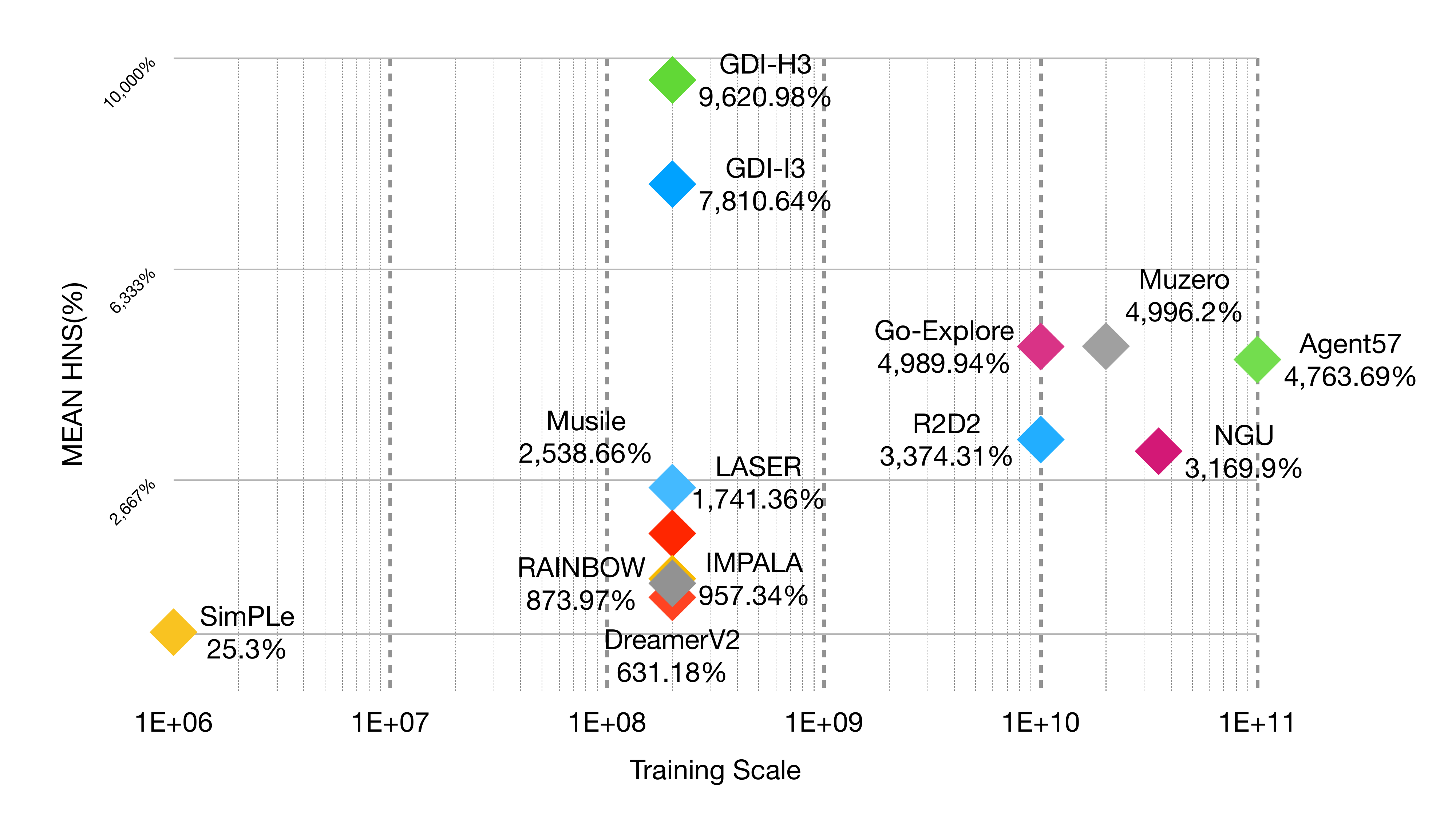}
	}
	\subfigure{
		\includegraphics[width=0.45\textwidth]{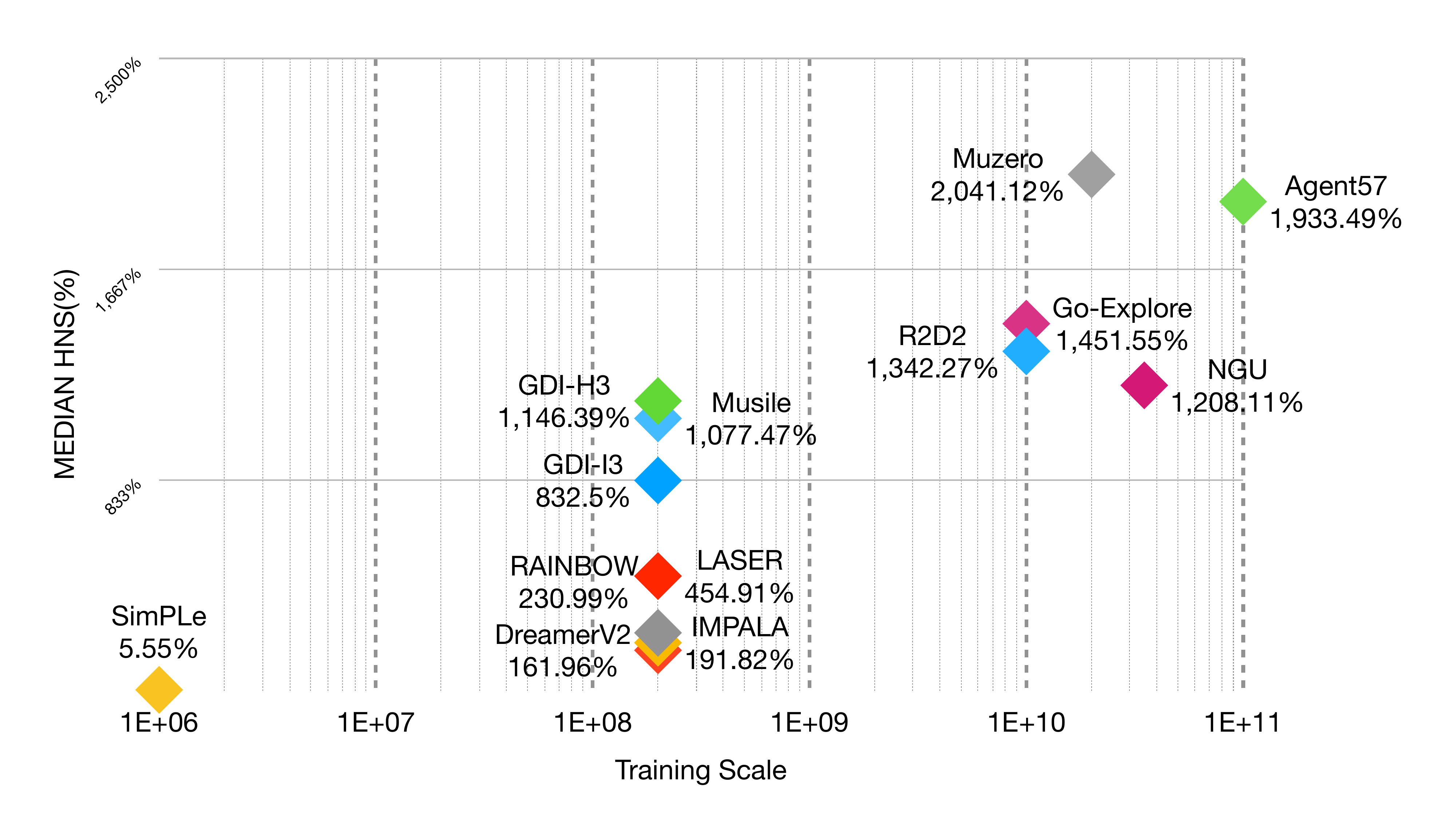}
	}
	\centering
	\caption{SOTA algorithms of Atari 57 games on mean and median HNS (\%) and corresponding training scale.}
	\label{fig: scale mean HNS time}
\end{figure*}

\begin{figure*}[!t]
    \centering
	\subfigure{
		\includegraphics[width=0.45\textwidth]{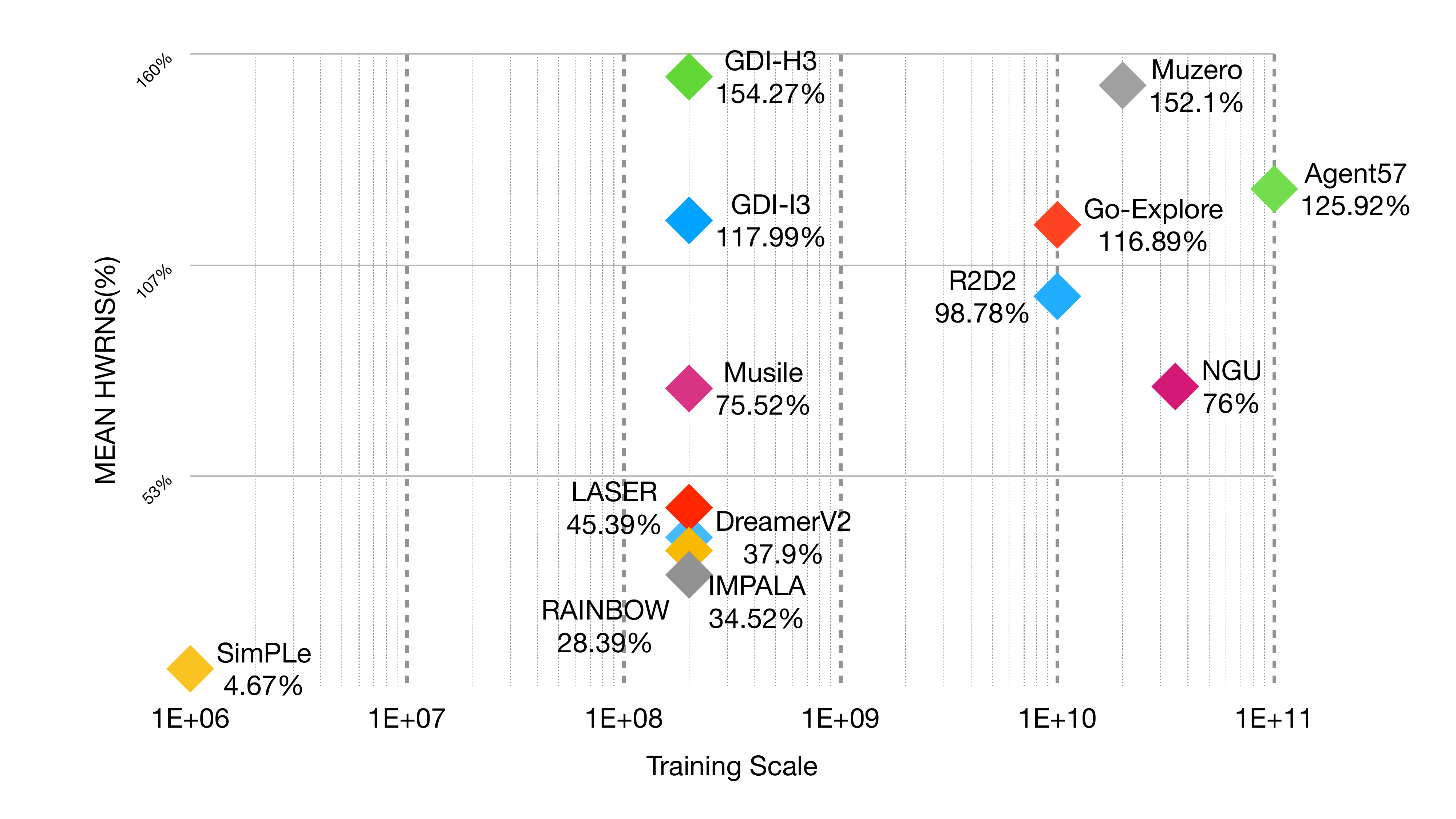}
	}
	\subfigure{
		\includegraphics[width=0.45\textwidth]{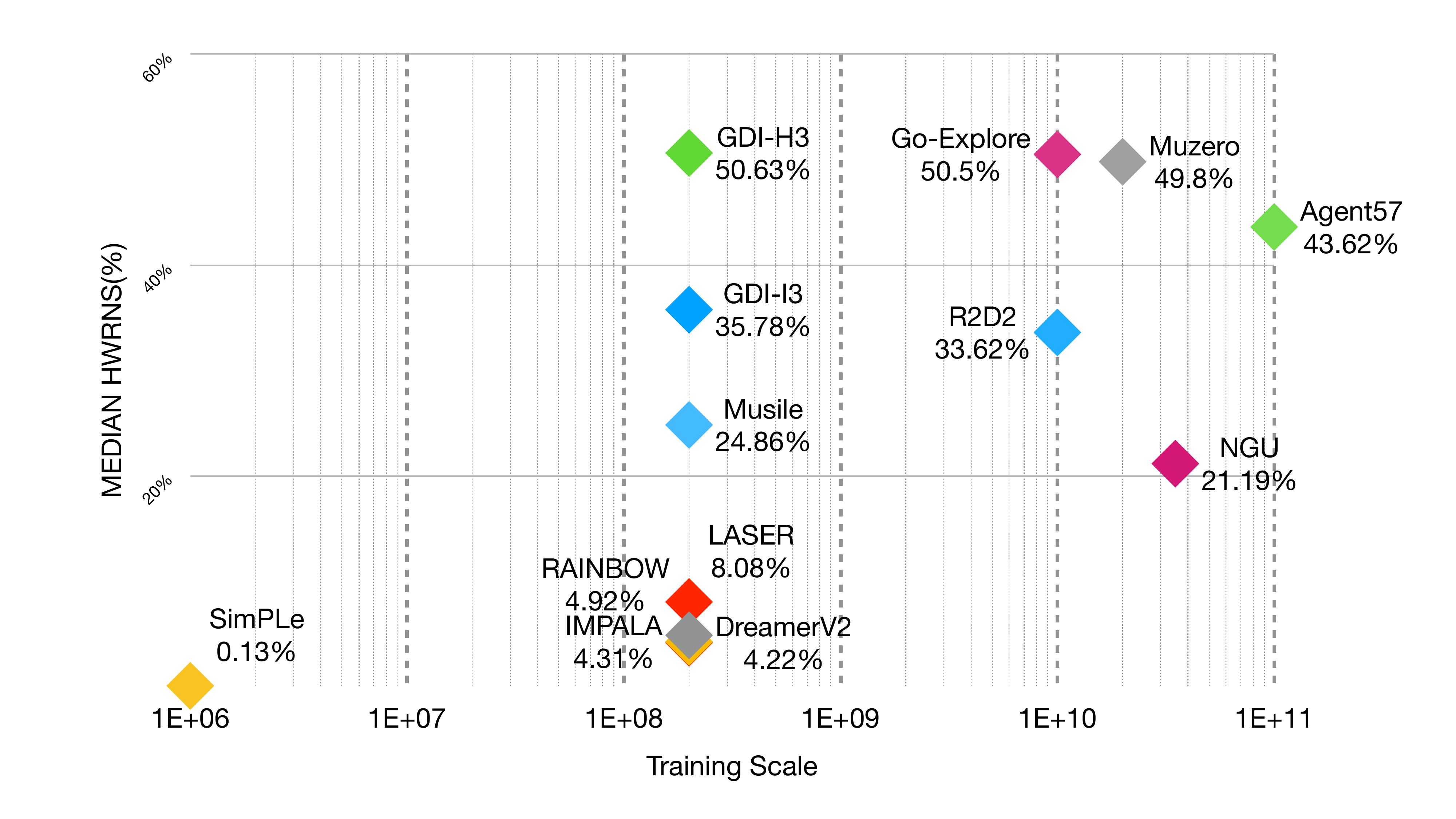}
	}
	\centering
	\caption{SOTA algorithms of Atari 57 games on mean and median HWRNS (\%) and corresponding training scale.}
	\label{fig: scale mean HWRNS time}
\end{figure*}

\section{Benchmark Results}
Since most of the previous work does not experiment on the standard human world records benchmark for reinforcement learning in ALE, we will report the final performance of each algorithm and provide the specific benchmarks settings upon the methodological differences in the Appendix for a fair comparison.

\subsection{Model-Free Reinforcement Learning}
\subsubsection{Rainbow}
Rainbow \citep{rainbow} is a classic value-based RL algorithm among the  DQN algorithm family, which has fruitfully combined six extensions of the DQN algorithm family. It is recognized to achieve state-of-the-art performance on the ALE benchmark. Thus, we select it as one of the representative algorithms of the SOTA DQN algorithms.
\subsubsection{IMPALA}
IMPALA, namely the Importance Weighted Actor Learner Architecture \citep{impala}, is a classic distributed off-policy actor-critic framework, which decouples acting from learning and learning from experience trajectories using V-trace. IMPALA actors communicate trajectories of experience (sequences of states, actions, and rewards) to a centralized learner, which boosts distributed large-scale training. Thus, we select it as one of the representative algorithms of the traditional distributed RL algorithm.
\subsubsection{LASER}
LASER \citep{laser} is a classic Actor-Critic algorithm, which investigated the combination of Actor-Critic algorithms with a uniform large-scale experience replay. It trained populations of actors with shared experiences and claimed to achieve SOTA in Atari. Thus, we select it as one of the SOTA RL algorithms within 200M training frames.

\subsubsection{R2D2}
\citep{r2d2}
Like IMPALA, R2D2 \citep{r2d2} is also a classic distributed RL algorithms. It trained  RNN-based RL agents from distributed prioritized experience replay, which achieved SOTA in Atari. Thus, we select it as one of the representative value-based distributed RL algorithms.

\subsubsection{NGU}
One of the classical problems in ALE for RL agents is the hard exploration problems \citep{goexplore,ale,agent57} like \textit{Private Eye, Montezuma’s Revenge, Pitfall!}. NGU \citep{ngu}, or Never Give Up, try to ease this problem by augmenting the reward signal with an internally generated intrinsic reward that is sensitive to novelty at two levels: short-term novelty within an episode and long-term novelty across episodes. It then learns a family
of policies for exploring and exploiting (sharing the same parameters) to obtain the highest score under the exploitative policy. NGU has achieved SOTA in Atari, and thus we selected it as one of the representative population-based model-free RL algorithms.

\subsubsection{Agent57}
Agent57 \citep{agent57} is the SOTA model-free RL algorithms on CHNS or Median HNS of Atari Benchmark. Built on the NGU agents, Agent57 proposed a novel state-action value function parameterization method and adopted an adaptive exploration over a family of policies, which overcome the drawback of NGU \citep{agent57}. We select it as one of the SOTA model-free RL algorithms.

\subsubsection{GDI}
GDI \citep{fan2021gdi}, or Generalized Data Distribution Iteration, claimed to have achieved  SOTA on mean/median HWRNS, mean HNS, HWRB, median SABER of Atari Benchmark. GDI is one of the novel Reinforcement Learning paradigms, which combined a data distribution optimization operator into the traditional generalized policy iteration (GPI) \citep{sutton} and thus achieved human-level learning efficiency.  Thus, we select them as one of the SOTA model-free RL algorithms.

\subsection{Model-Based Reinforcement Learning}
\subsubsection{SimPLe}
As one of the classic model-based RL algorithms on Atari, SimPLe, or Simulated Policy Learning \citep{modelbasedatari}, adopted a video prediction model to enable RL agents to solve Atari problems with higher sample efficiency. It claimed to outperform the SOTA model-free algorithms in most games, so we selected it as representative model-based RL algorithms.
\subsubsection{Dreamer-V2}
Dreamer-V2 \citep{dreamerv2} built world models to facilitate generalization across the experience and allow learning behaviors from imagined outcomes in the compact latent space of the world model to increase sample efficiency. Dreamer-V2 is claimed to achieve SOTA in Atari, and thus we select it as one of the SOTA model-based RL algorithms within the 200M training scale.
\subsubsection{Muzero}
Muzero \citep{muzero} combined a tree-based search with a learned model and has achieved superhuman performance on Atari. We thus selected it as one of the SOTA model-based RL algorithms.
\subsection{Other SOTA algorithms}
\subsubsection{Go-Explore}
As mentioned in NGU, a grand challenge in reinforcement learning is intelligent exploration, which is called the hard-exploration problem \citep{ale2}. Go-Explore \citep{goexplore} adopted three principles to solve this problem. Firstly, agents remember previously visited states. Secondly, agents first return to a promising state and then explore it. Finally, solve simulated environment through any available means, and then robustify via imitation learning. Go-Explore has achieved SOTA in Atari, so we select it as one of the SOTA algorithms of the hard exploration problem.

\subsubsection{Musile}
Musile \citep{muesli} proposed a novel policy update that combines regularized policy optimization with model learning as an auxiliary loss. It acts directly with a policy network and has a computation speed comparable to model-free baselines. As it claimed to achieve SOTA in Atari within 200M training frames, we select it as one of the SOTA  RL algorithms within 200M training frames.

\subsection{Summary of Benchmark Results}
\begin{figure*}[!t]
    \centering
	\subfigure{
		\includegraphics[width=0.45\textwidth]{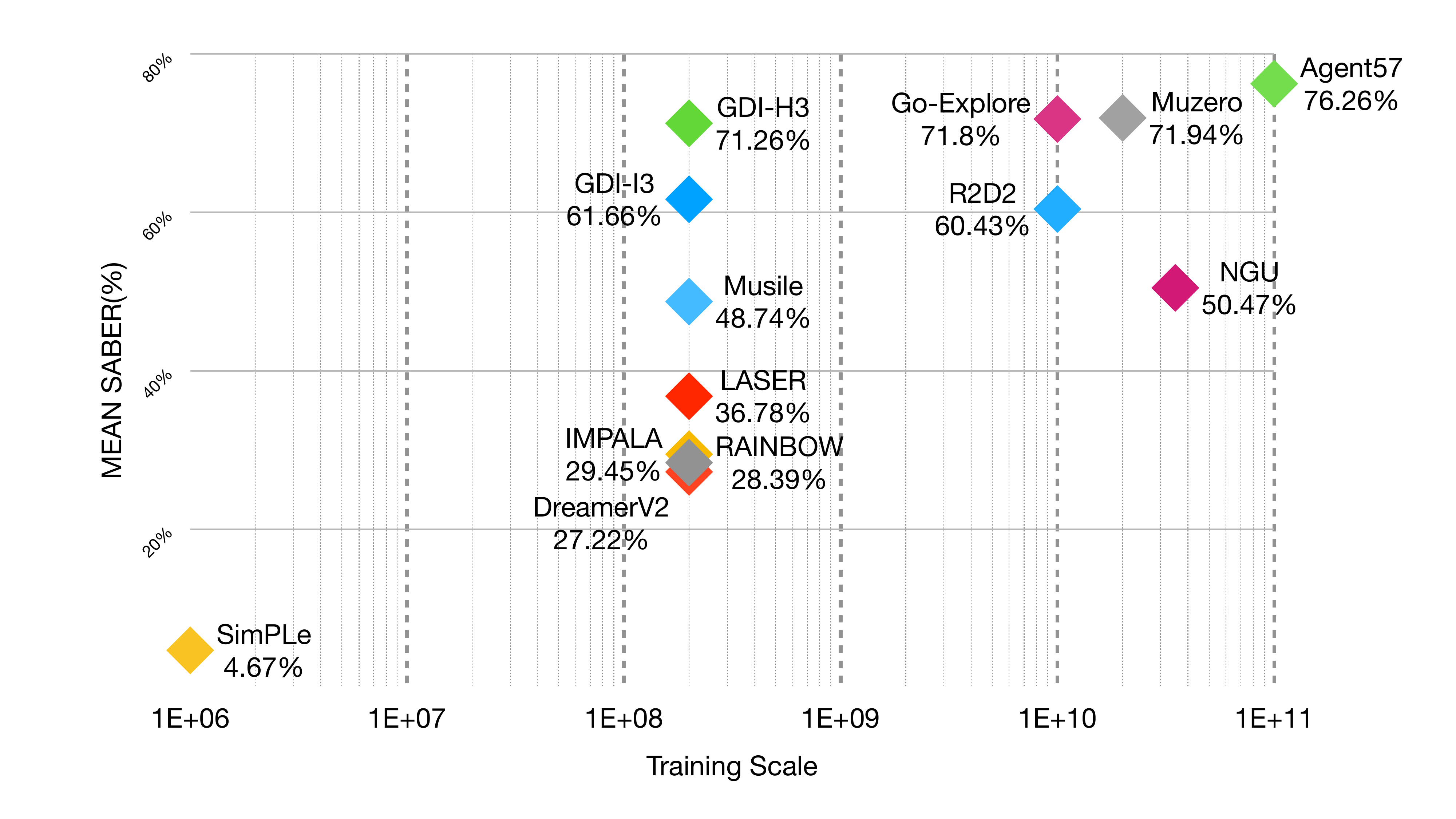}
	}
	\subfigure{
		\includegraphics[width=0.45\textwidth]{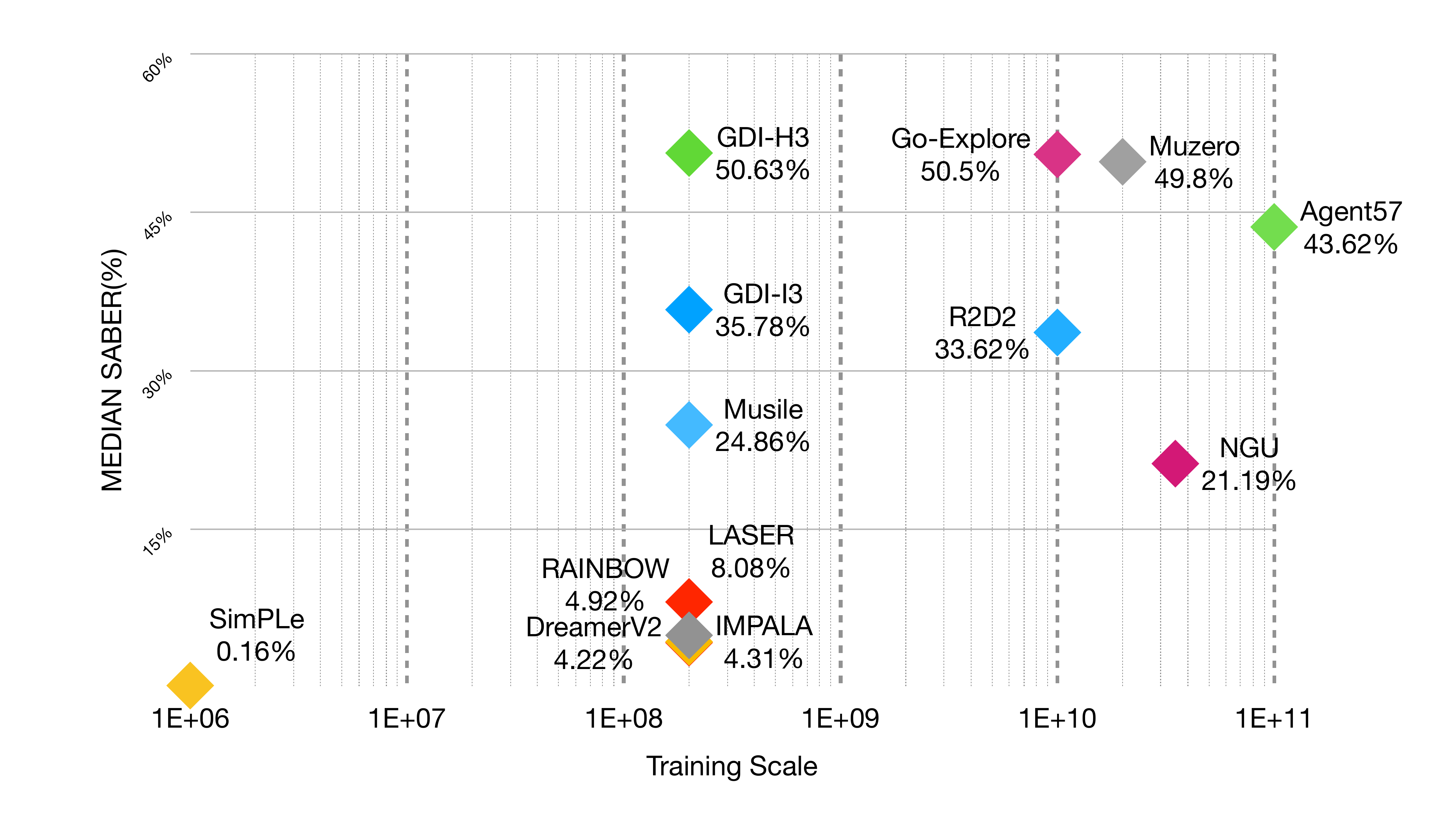}
	}
	\centering
	\caption{SOTA algorithms of Atari 57 games on mean and median SABER (\%) and corresponding training scale.}
	\label{fig: scale mean SABER time}
\end{figure*}

\begin{figure*}[!t]
    \centering
	\subfigure{
		\includegraphics[width=0.45\textwidth]{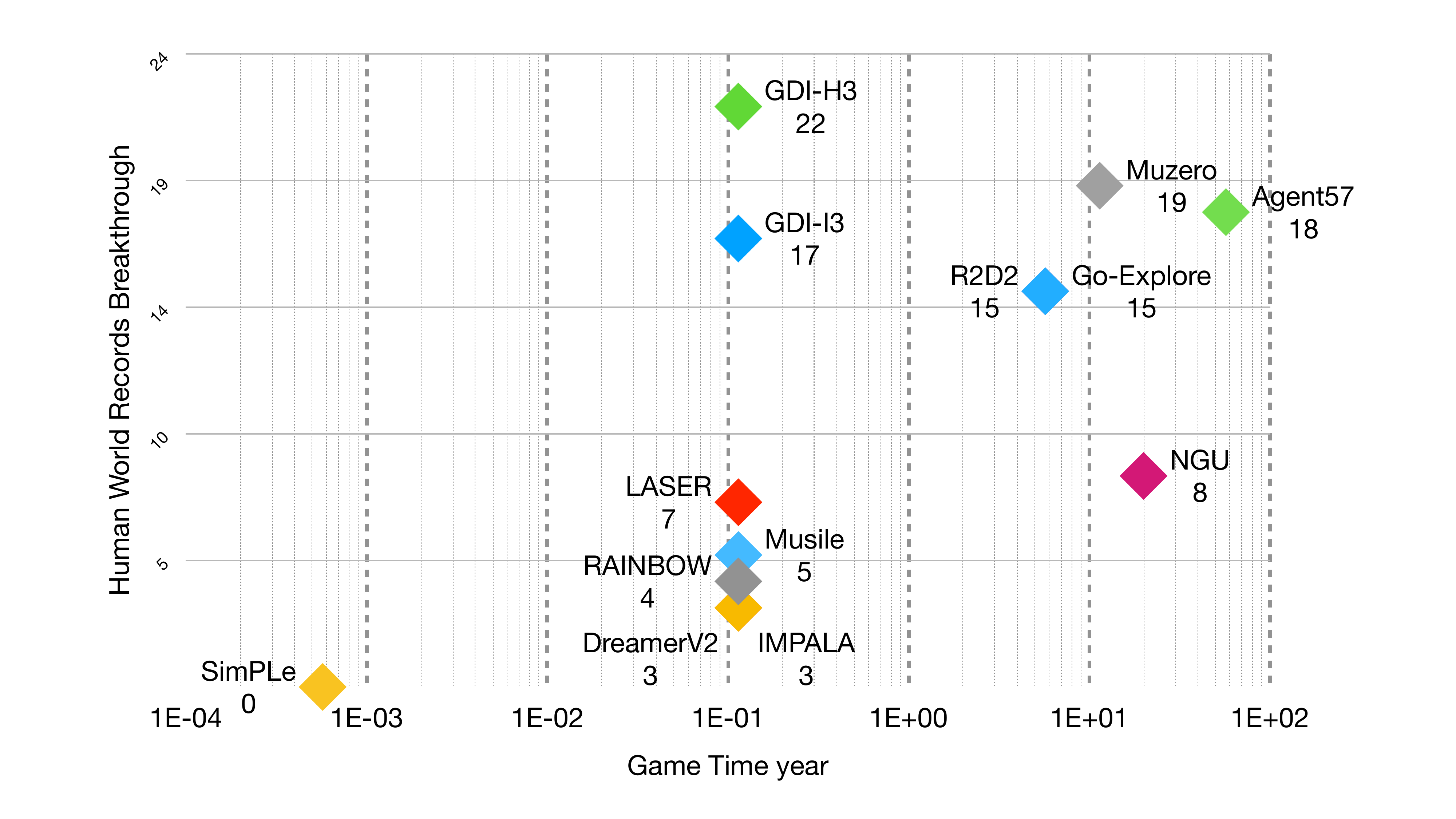}
	}
	\subfigure{
		\includegraphics[width=0.45\textwidth]{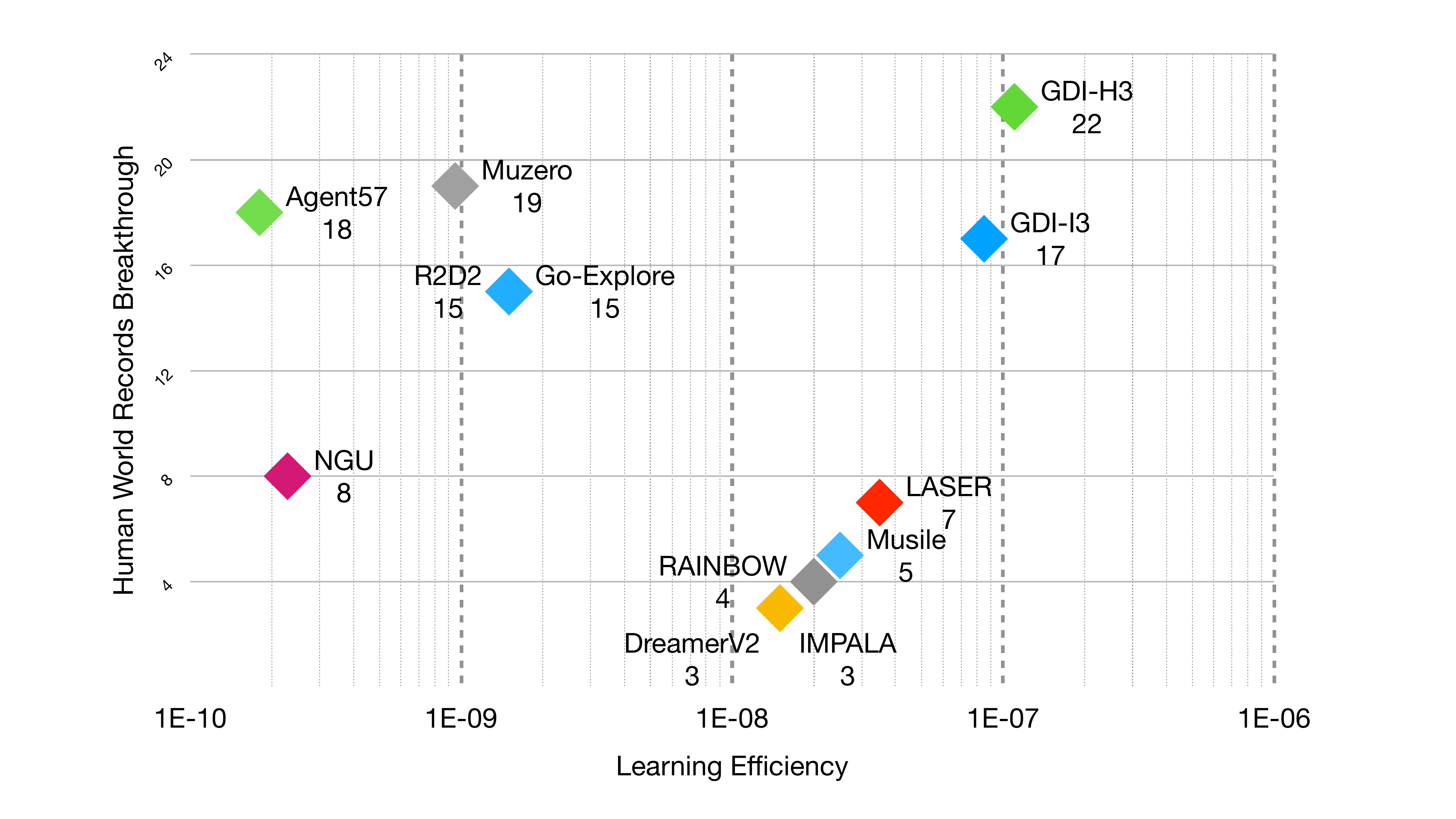}
	}
	\centering
	\caption{SOTA algorithms of Atari 57 games on HWRB.
	HWRB of SimPLe is 0, so it's not shown in the up-right figure.}
	\label{fig: HWRB time}
\end{figure*}

This part summarizes the results among all the algorithms we mentioned above on the human world record benchmark for Atari. In Figs, we illustrated the benchmark results on HNS, HWRNS, SABER, and the corresponding training scale. \ref{fig: scale mean HNS time}, \ref{fig: scale mean HWRNS time} and \ref{fig: scale mean SABER time}, HWRB and corresponding game time (year) and learning efficiency in Fig. \ref{fig: HWRB time}. From those results, we see GDI \citep{fan2021gdi} has achieved SOTA in learning efficiency, HWRB, HWRNS, mean HNS, and median SABER within 200M training frames. Agent57 has achieved SOTA in mean SABER, and Muzero \citep{muzero} has achieved SOTA in median HNS. To avoid any aggregated metrics issues, we provide all the raw scores of those algorithms in the Appendix.

\section{Challenges and Solutions}
Although RL has achieved fantastic achievements in Atari benchmarks, we could not claim that we have made superhuman agents in Atari. There are still many challenges in the Atari Benchmarks, revealing the drawback of current RL algorithms. We believe discussing those challenges and solutions may promote the development of RL research. Therefore, in this section, we discuss the current challenges and promising solutions.
\subsection{Current Challenges}
\subsubsection{Human World Record} From Figs. \ref{fig: HWRB time}, we see there are at least 35 human world records that have not been broken through by current SOTA RL algorithms. Therefore, it is  too early to say we have achieved superhuman performance. 
\subsubsection{Hard Exploration Problem} From Tabs \ref{tab: 200 model-free Atari Games Table of Scores Based on Human World Records}, \ref{tab: 10B model-free Atari Games Table of Scores Based on Human World Records}, and \ref{tab: other Atari Games Table of Scores Based on Human World Records} in the Appendix, we see current SOTA algorithms are fragile facing the hard exploration problems, and others like Agent57 or Go-Explore that have overcome those problems failed to balance the trade-off between exploration and exploitation leading to lower learning efficiency.

\subsubsection{Planning and Modeling} Learning from sparse rewards is extremely difficult for model-free RL algorithms, especially those without intrinsic rewards that struggle to learn from weak gradient signals. Model-based methods can ease those problems by adopting a world model to planning \citep{muzero} or replay \citep{dreamerv2}, which both enhanced the gradient signals. However, being utterly dependent on planning is unrealistic and will lose generality in some Atari Games like Tennis.

\subsubsection{Learning Efficiency} From Figs. \ref{fig: HWRB time}, current SOTA algorithms like Agent57 may require more than 52.7 years of game-play to achieve SOTA performance, which revealed its low learning efficiency. As recommended in \citep{dreamerv2}, we also argue for high learning efficiency algorithms, and we advocate that 200M training frames (equal to 38 days) are enough for achieving a superhuman agent.

\subsection{Promising Solutions}

\subsubsection{Adaptive Exploration-Exploitation Balance} The trade-off between exploration-exploitation is a classic difficult problem in RL algorithms \cite{sutton}. Algorithms designed for hard exploration problems may fail to trade-off the balance leading to low sample efficiency. Others may suffer from hard exploration problems. Thus how to trade off balance becomes more important. NGU, Agent57 tried to ease this problem by training a family of policies from extremely explorative to highly exploitative. Based on that, GDI \citep{fan2021gdi} proposed the data distribution iterator to formulate this procedure and revealed its superiority to the origin process without the data distribution iterator. It may be a promising way to solve this problem.

\subsubsection{Long Term Planning} Planning algorithms like Muzero \citep{muzero} fail when the outcome signals of the planning algorithms become misleading or indistinguishable. The former may come from the accumulation of model approximation errors, and the latter may come from a relatively sparse rewards environment like Montezuma Revenge. The former problem may need more assistance from more advanced deep learning methods, but the latter can be solved by a long time of planning. More precisely, we need a big picture that guides agents towards a better decision. GDI \citep{fan2021gdi} showed a promising way to combine techniques from Go-Explore and Muzero into the data distribution iteration operator and guide the policy inside an episode, which may solve the low learning efficiency and hard exploration problems. Musile \citep{muesli} also offers another interesting combination of policy-based methods with model-based RL.

\subsubsection{Compatible Reinforcement Learning Frameworks}
As \citet{rainbow} put it, the DRL community has made several independent improvements those years, but it is unclear whether there are unified frameworks that can fruitfully combine those improvements and make each component compatible. CASA \citep{CASA-B} provided promising frameworks, and based on CASA, GDI further proposed a more general paradigm. We believe a unified framework may help to obtain the superhuman agents.

\section{Conclusion}
In this paper, we reviewed the current evaluation metrics for Atari Benchmarks and discussed their advantages and disadvantages. To further the progress in the field, we proposed the Human World Records Benchmark for Reinforcement Learning on Atari, which we suggest for testing a real superhuman agent. Besides, we also provide benchmark results in the human world record benchmark, which may serve as a point of comparison for future work in the ALE. In the final part of this paper, we concluded the challenges and promising solutions that we took from revisiting those milestones in Atari. We also highlighted the current open challenges, including planning and modeling, hard exploration, human world records, and low learning efficiency.

\bibliography{aaai22}

\onecolumn
\begin{appendix}
\begin{center}
\Huge
    \textbf{Appendix} 
\normalsize
\end{center}

\section{Atari Benchmark Settings}
In this part, we will provide the benchmark settings of each algorithm.

\begin{table}[!hb]
\scriptsize
\begin{center}
\setlength{\tabcolsep}{1.0pt}
\begin{tabular}{ c c c c c c c c}
\toprule
                        & Max episode length & Num. Action Repeats & Num. Frame Stacks & Image Size  & Grayscaled/RGB & Live Information & Action Space Dimension\\
\midrule
RainBow                 & 30min & 4 & 4 & (84, 84) & Grayscaled & Yes  & 4  \\
IMPALA                  & 30min & 4 & 4 & (84, 84) & Grayscaled & Yes  & 18 (Full)\\
LASER                   & 30min & \textbf{4}  & \textbf{4}  & (84, 84) & Grayscaled & \textbf{No}   & \textbf{18 (Full)} \\
R2D2                    & 30min & \textbf{4} & 4 & (84, 84) & Grayscaled & No  & 18 (Full) \\
NGU                     & 30min & 4 & 1 & (84, 84) & Grayscaled & No  & 18 (Full) \\
Agent57                 & 30min & 4 & 1 & (84, 84) & Grayscaled & No  & 18 (Full) \\
GDI                     & 30min & 4 & 4 & (84, 84) & Grayscaled & No  & 18 (Full) \\ 
SimPLe                  & \textbf{30min}  & 4 & 4 & (210, 160) & RGB & No  & \textbf{4}  \\
Dreamer-V2              & 30min & 4 & 1 & (84, 84) & Grayscaled & No  & 18 (Full) \\
Muzero                  & 30min & \textbf{4} & 4 & (84, 84) & Grayscaled & No  & 18 (Full) \\
Go-Explore              & 30min & 4 & \textbf{1} & \textbf{(84, 84)}  & Grayscaled & Yes  & \textbf{18 (Full)} \\
Musile                  & 30min & 4 & 4 & (96, 96) & Grayscaled & No  & 18 (Full) \\
\bottomrule
\end{tabular}
\caption{Atari hyperparameters for training. The values in bold have not been mentioned  in the original articles, so we consider them as default values.}
\label{tab:ale_process}
\end{center}
\end{table}

\begin{table}[!hb]
\scriptsize
\begin{center}
\setlength{\tabcolsep}{1.0pt}
\begin{tabular}{ c c c c c c c c c}
\toprule
                        & Max episode length & Num. Action Repeats & Num. Frame Stacks in & Image Size  & Grayscaled/RGB & Episode Termination  & Action Space Dimension & Num. Averaging Episodes k\\
\midrule
RainBow                 & 30min & 4 & 4 & (84, 84) & Grayscaled & All lives lost  & 4                                              & 200 \\
IMPALA                  & 30min & 4 & 4 & (84, 84) & Grayscaled & All lives lost  & 18 (Full)                                      & 200\\
LASER                   & 30min & \textbf{4}  & \textbf{4}  & (84, 84) & Grayscaled & All lives lost   & \textbf{18 (Full)}& 100 \\
R2D2                    & 30min & \textbf{4} & 4 & (84, 84) & Grayscaled & All lives lost  & 18 (Full)                              & 10\\
NGU                     & 30min & 4 & 1 & (84, 84) & Grayscaled & All lives lost  & 18 (Full)                                       & 32\\
Agent57                 & 30min & 4 & 1 & (84, 84) & Grayscaled & All lives lost  & 18 (Full)                                       & 50\\
GDI                     & 30min & 4 & 4 & (84, 84) & Grayscaled & All lives lost  & 18 (Full)                                       & 32\\ 
SimPLe                  & \textbf{30min}  & 4 & 4 & (210, 160) & RGB & All lives lost & \textbf{4}                                 & 5\\
Dreamer-V2              & 30min & 4 & 1 & (84, 84) & Grayscaled & All lives lost  & 18 (Full)                                       & 10\\
Muzero                  & 30min & \textbf{4} & 4 & (84, 84) & Grayscaled & All lives lost  & 18 (Full)                              & 1000\\
Go-Explore              & 30min & 4 & \textbf{1} & \textbf{(84, 84)}  & Grayscaled & All lives lost  & \textbf{18 (Full)}          & 50\\
Musile                  & 30min & 4 & 4 & (96, 96) & Grayscaled & All lives lost  & 18 (Full)                                       & 100\\
\bottomrule
\end{tabular}
\caption{Atari hyperparameters for evaluation. The values in bold have not been mentioned  in the original articles, so we consider them as default values.}
\label{tab:ale_process}
\end{center}
\end{table}

\clearpage

\section{Atari Games Table of Scores Based on Human Average Records}
\label{app: Atari Games Table of Scores Based on Human Average Records}
In this part, we detail the raw score of several representative SOTA algorithms, including the SOTA 200M model-free algorithms, SOTA 10B+ model-free algorithms, SOTA model-based algorithms, and other SOTA algorithms.\footnote{200M and 10B+ represent the training scale.} Additionally, we calculate the Human Normalized Score (HNS) of each game with each algorithm. First of all, we demonstrate the sources of the scores that we used.
Random scores and average human's scores are from \citep{agent57}.
Rainbow's scores are from \citep{rainbow}.
IMPALA's scores are from \citep{impala}.
LASER's scores are from \citep{laser}, no sweep at 200M.
As there are many versions of R2D2 and NGU, we use original papers'.
R2D2's scores are from \citep{r2d2}.
NGU's scores are from \citep{ngu}.
Agent57's scores are from \citep{agent57}.
MuZero's scores are from \citep{muzero}.
DreamerV2's scores are from \citep{dreamerv2}.
SimPLe's scores are form \citep{modelbasedatari}.
Go-Explore's scores are form \citep{goexplore}.
Muesli's scores are form \citep{muesli}.
In the following we detail the raw scores and HNS of each algorithm on 57 Atari games.
\clearpage
\begin{table}[!hb]
\scriptsize
\begin{center}
\setlength{\tabcolsep}{1.0pt}
\begin{tabular}{ c c c c c c c c c c c c c}
\toprule
Games & RND & HUMAN & RAINBOW & HNS(\%) & IMPALA & HNS(\%) & LASER & HNS(\%) & GDI-I$^3$ & HNS(\%) & GDI-H$^3$ & HNS(\%)\\
\midrule
Scale  &     &       & 200M   &       &  200M    &        & 200M   &         &  200M   &  &  200M   &\\
\midrule
 alien  & 227.8 & 7127.8            & 9491.7 & 134.26 & 15962.1  & 228.03 & 35565.9 & 512.15                        &43384             &625.45         &\textbf{48735}             &\textbf{703.00}      \\
 amidar & 5.8   & 1719.5            & \textbf{5131.2} & \textbf{299.08} & 1554.79  & 90.39  & 1829.2  & 106.4       &1442              &83.81          &1065                       &61.81       \\
 assault & 222.4 & 742              & 14198.5 & 2689.78 & 19148.47 & 3642.43  & 21560.4 & 4106.62                   &63876             &12250.50       &\textbf{97155}             &\textbf{18655.23}    \\
 asterix & 210   & 8503.3           & 428200 & 5160.67 & 300732   & 3623.67  & 240090  & 2892.46                    &759910            &9160.41        &\textbf{999999}            &\textbf{12055.38}    \\
 asteroids & 719 & 47388.7          & 2712.8 & 4.27   & 108590.05 & 231.14  & 213025  &  454.91                     &751970            &1609.72        &\textbf{760005 }           &\textbf{1626.94}     \\
 atlantis & 12850 & 29028.1         & 826660 & 5030.32 & 849967.5 & 5174.39 & 841200 & 5120.19                      &3803000           &23427.66       &\textbf{3837300}           &\textbf{23639.67}     \\
 bank heist & 14.2 & 753.1          & 1358   & 181.86  & 1223.15  & 163.61  & 569.4  & 75.14                        &\best{1401}       &\best{187.68}  &1380                       &184.84       \\
 battle zone & 236 & 37187.5        & 62010 & 167.18  & 20885    & 55.88  & 64953.3 & 175.14                        &478830            &1295.20        &\textbf{824360}            &\textbf{2230.29}      \\
 beam rider & 363.9 & 16926.5       & 16850.2 & 99.54 & 32463.47 & 193.81 & 90881.6 & 546.52                        &162100            &976.51         &\textbf{422890}            &\textbf{2551.09}      \\
 berzerk & 123.7 & 2630.4           & 2545.6   & 96.62  & 1852.7   & 68.98  & \textbf{25579.5} & \textbf{1015.51}   &7607              &298.53         &14649             &579.46       \\
 bowling & 23.1 & 160.7             & 30   & 5.01        & 59.92    & 26.76  & 48.3    & 18.31                      &201.9             &129.94         &\textbf{205.2}             &\textbf{132.34}       \\
 boxing  & 0.1  & 12.1              & 99.6 & 829.17      & 99.96    & 832.17 & \textbf{100} & \textbf{832.5}        &\best{100}        &\best{832.50}  &\textbf{100}               &\textbf{832.50}       \\
 breakout & 1.7 & 30.5              & 417.5 & 1443.75    & 787.34   & 2727.92 & 747.9 & 2590.97                     &\best{864}        &\best{2994.10} &\textbf{864}               &\textbf{2994.10}      \\
 centipede & 2090.9 & 12017         & 8167.3 & 61.22   & 11049.75 & 90.26   & \textbf{292792} & \textbf{2928.65}    &155830            &1548.84        &195630                     &1949.80      \\
 chopper command & 811 & 7387.8     & 16654 & 240.89 & 28255  & 417.29  & 761699 & 11569.27                         &\best{999999}     &\best{15192.62}&\textbf{999999}            &\textbf{15192.62}     \\
 crazy climber & 10780.5 & 36829.4  & 168788.5 & 630.80 & 136950 & 503.69 & 167820  & 626.93                        &201000            &759.39         &\textbf{241170}            &\textbf{919.76}  \\
 defender & 2874.5 & 18688.9        & 55105 & 330.27 & 185203 & 1152.93 & 336953  & 2112.50                         &893110            &5629.27        &\textbf{970540}            &\textbf{6118.89}     \\
 demon attack & 152.1 & 1971        & 111185 & 6104.40 & 132826.98 & 7294.24 & 133530 & 7332.89                     &675530                 &37131.12   &\textbf{787985}                     &\textbf{43313.70}     \\
 double dunk & -18.6 & -16.4        & -0.3   & 831.82  & -0.33     & 830.45  & 14     & 1481.82                     &\best{24}         &\best{1936.36} &\textbf{24}                &\textbf{1936.36}     \\
 enduro      & 0   & 860.5          & 2125.9 & 247.05  & 0       & 0.00     & 0    & 0.00                           &\best{14330}      &\best{1665.31 }&14300                      &1661.82      \\
 fishing derby & -91.7 & -38.8      & 31.3 & 232.51  & 44.85   & 258.13    & 45.2   & 258.79                        &59         &285.71                &\textbf{65}                &\textbf{296.22}   \\
 freeway       & 0     & 29.6       & \textbf{34} & \textbf{114.86}  & 0     & 0.00       & 0    & 0.00             &\best{34}          &\best{114.86   }&\textbf{34}               &\textbf{114.86}   \\
 frostbite     & 65.2  & 4334.7     & 9590.5 & 223.10 & 317.75 & 5.92     & 5083.5 & 117.54                         &10485              &244.05          &\textbf{11330}            &\textbf{263.84}   \\
 gopher  & 257.6 & 2412.5           & 70354.6 & 3252.91    & 66782.3 & 3087.14 & 114820.7 & 5316.40                 &\best{488830}     &\best{22672.63} &473560           &21964.01    \\
 gravitar & 173 & 3351.4            & 1419.3  & 39.21   & 359.5      & 5.87    & 1106.2   & 29.36                   &5905       &180.34                 &\textbf{5915}             &\textbf{180.66}   \\
 hero   & 1027 & 30826.4            & \textbf{55887.4} & \textbf{184.10}   & 33730.55  & 109.75  & 31628.7 & 102.69 &38330             &125.18          &38225            &124.83    \\
 ice hockey & -11.2 & 0.9           & 1.1    & 101.65   & 3.48      & 121.32   & 17.4    & 236.36                   &44.94      &463.97&\textbf{47.11}           &\textbf{481.90}   \\
 jamesbond  & 29    & 302.8         & 19809 & 72.24   & 601.5     & 209.09   & 37999.8 & 13868.08                   &594500     &217118.70 &\textbf{620780}          &\textbf{226716.95}    \\
 kangaroo   & 52    & 3035          & \textbf{14637.5} & \textbf{488.05} & 1632    & 52.97    & 14308   & 477.91    &14500             &484.34           &14636           &488.00    \\
 krull     & 1598   & 2665.5        & 8741.5  & 669.18 & 8147.4  & 613.53   & 9387.5  &  729.70                     &97575      &8990.82   &\textbf{594540}          &\textbf{55544.92}     \\
 kung fu master & 258.5 & 22736.3   & 52181 & 230.99 & 43375.5 & 191.82 & 607443 & 2701.26        &140440            &623.64           &\textbf{1666665}          &\textbf{7413.57}         \\
 montezuma revenge&0&\textbf{4753.3}& 384   & 8.08   & 0       & 0.00   & 0.3    & 0.01                             &3000              &63.11            &2500            &52.60   \\
 ms pacman  & 307.3 & 6951.6        & 5380.4  & 76.35   & 7342.32 & 105.88 & 6565.5   & 94.19                       &11536      &169.00    &\textbf{11573}           &\textbf{169.55}    \\
 name this game & 2292.3 & 8049     & 13136 & 188.37   & 21537.2 & 334.30 & 26219.5 & 415.64                        &34434      &558.34    &\textbf{36296}           &\textbf{590.68}    \\
 phoenix & 761.5 & 7242.6  & 108529 & 1662.80   & 210996.45  & 3243.82 & 519304 & 8000.84                           &894460     &13789.30  &\textbf{959580}          &\textbf{14794.07}  \\
 pitfall & -229.4 & \textbf{6463.7} & 0      & 3.43      & -1.66      & 3.40    & -0.6   & 3.42                     &0                 &3.43             &-4.345            &3.36 \\
 pong    & -20.7  & 14.6   & 20.9   & 117.85    & 20.98      & 118.07  & \textbf{21}     &  \textbf{118.13}         &\best{21   }      &\best{118.13}    &\textbf{21}              &\textbf{118.13}    \\
 private eye & 24.9&\textbf{69571.3}& 4234 & 6.05     & 98.5       & 0.11    & 96.3   & 0.10                        &15100             &21.68            &15100           &21.68       \\
 qbert  & 163.9 & 13455.0 & 33817.5 & 253.20   & \textbf{351200.12}  & \textbf{2641.14} & 21449.6 & 160.15          &27800             &207.93           &28657           &214.38    \\
 riverraid & 1338.5 & 17118.0       & 22920.8 & 136.77 & 29608.05  & 179.15  & \textbf{40362.7} & \textbf{247.31}   &28075             &169.44           &28349           &171.17    \\
 road runner & 11.5 & 7845          & 62041   & 791.85 & 57121     & 729.04  & 45289   & 578.00                     &878600            &11215.78         &\textbf{999999} &\textbf{12765.53} \\
 robotank   & 2.2   & 11.9          & 61.4   & 610.31    & 12.96     & 110.93  & 62.1    & 617.53                   &108.2             &1092.78          &\textbf{113.4}           &\textbf{1146.39}  \\
 seaquest  & 68.4 & 42054.7         & 15898.9 & 37.70    & 1753.2    & 4.01    & 2890.3  & 6.72                     &943910	           &2247.98  &\textbf{1000000}          &\textbf{2381.57}\\
 skiing & -17098  & \textbf{-4336.9}& -12957.8 & 32.44  & -10180.38 & 54.21   & -29968.4 & -100.86                  &-6774             &80.90            &-6025	          &86.77   \\
 solaris & 1236.3 & \textbf{12326.7}& 3560.3  & 20.96  & 2365      & 10.18   & 2273.5   & 9.35                      &11074             &88.70            &9105            &70.95   \\
 space invaders & 148 & 1668.7      & 18789 & 1225.82 & 43595.78 & 2857.09 & 51037.4 & 3346.45                      &140460            &9226.80          &\textbf{154380} &\textbf{10142.17}   \\
 star gunner & 664 & 10250          & 127029    & 1318.22 & 200625   & 2085.97 & 321528  & 3347.21                  &465750            &4851.72          &\textbf{677590} &\textbf{7061.61}    \\
 surround    & -10 & 6.5            & \textbf{9.7}       & \textbf{119.39}  & 7.56     & 106.42  & 8.4     & 111.52 &-7.8              &13.33            &2.606           &76.40    \\
 tennis  & -23.8   & -8.3           & 0        & 153.55    & 0.55     & 157.10  & 12.2    & 232.26                  &\best{24       }  &\best{308.39   } &\textbf{24}              &\textbf{308.39}  \\
 time pilot & 3568 & 5229.2         & 12926 & 563.36     & 48481.5  & 2703.84 & 105316  & 6125.34                   &216770     &12834.99         &\textbf{450810}          &\textbf{26924.45}   \\
 tutankham  & 11.4 & 167.6          & 241   & 146.99     & 292.11   & 179.71  & 278.9   & 171.25                    &\best{423.9 }     &\best{264.08   } &418.2           &260.44  \\
 up n down  & 533.4 & 11693.2       & 125755 & 1122.08 & 332546.75 & 2975.08 & 345727 & 3093.19                     &\best{986440}     &\best{8834.45 }  &966590          &8656.58    \\
 venture    & 0     & 1187.5        & 5.5    & 0.46    & 0         & 0.00    & 0      & 0.00                        &\best{2035     }  &\best{171.37   } &2000            &168.42   \\
 video pinball & 0 & 17667.9        & 533936.5 & 3022.07 & 572898.27 & 3242.59 & 511835 & 2896.98                   &925830     &5240.18                 &\textbf{978190} &\textbf{5536.54}     \\
 wizard of wor & 563.5 & 4756.5     & 17862.5 & 412.57 & 9157.5    & 204.96  & 29059.3 & 679.60                     &\best{64239 }     &\best{1519.90 }  &63735           &1506.59     \\
 yars revenge & 3092.9 & 54576.9    & 102557 & 193.19 & 84231.14  & 157.60 & 166292.3  & 316.99                     &\textbf{972000}     &\textbf{1881.96}   &968090          &1874.36     \\
 zaxxon       & 32.5   & 9173.3     & 22209.5 & 242.62 & 32935.5   & 359.96 & 41118    & 449.47                     &109140     &1193.63   &\textbf{216020} &\textbf{2362.89}     \\
\hline
MEAN HNS(\%)        &     0.00 & 100.00   &         & 873.97 &         & 957.34  &        & 1741.36 &      & \GDIImeanhns       &      & \textbf{\GDIHmeanhns} \\
Learning Efficiency &     0.00 & N/A   &         & 4.37E-08 &         & 4.79E-08  &        & 8.71E-08 &      & 3.91E-07       &      & \textbf{4.70E-07} \\
\hline
MEDIAN HNS(\%)      & 0.00   & 100.00   &         & 230.99 &         & 191.82  &        & 454.91  &      & \GDIImedianhns        &      & \textbf{\GDIHmedianhns} \\
Learning Efficiency & 0.00   & N/A   &         & 1.15E-08 &         & 9.59E-09  &        & 2.27E-08 &      & 4.16E-08       &      & \textbf{5.73E-08} \\
\bottomrule
\end{tabular}
\caption{Score table of SOTA 200M model-free algorithms on HNS.}
\end{center}
\end{table}
\clearpage
\begin{table}[!hb]
\scriptsize
\begin{center}
\setlength{\tabcolsep}{1.0pt}
\begin{tabular}{ c c c c c c c c c c c}
\toprule
 Games & R2D2 & HNS(\%) & NGU & HNS(\%) & AGENT57 & HNS(\%) & GDI-I$^3$ & HNS(\%) & GDI-H$^3$ & HNS(\%) \\
\midrule
Scale  & 10B   &        & 35B &         & 100B     &        & 200M &     &  200M   &\\
\midrule
 alien  & 109038.4 & 1576.97 & 248100 & 3592.35 & \textbf{297638.17} & \textbf{4310.30}             &43384             &625.45                &48735             &703.00       \\
 amidar & 27751.24 & 1619.04 & 17800  & 1038.35 & \textbf{29660.08}  & \textbf{1730.42}             &1442              &83.81                 &1065              &61.81        \\
 assault &  90526.44 & 17379.53 & 34800 & 6654.66 & 67212.67 & 12892.66            &63876           &12250.50          &\textbf{97155}        &\textbf{18655.23}     \\
 asterix &  999080   & 12044.30 & 950700 & 11460.94 & 991384.42 & 11951.51         &759910          &9160.41           &\textbf{999999}       &\textbf{12055.38}     \\
 asteroids & 265861.2 & 568.12 & 230500 & 492.36   & 150854.61 & 321.70                             &751970            &1609.72               &\textbf{760005}            &\textbf{1626.94}      \\
 atlantis & 1576068   & 9662.56 & 1653600 & 10141.80 & 1528841.76 & 9370.64                         &3803000           &23427.66              &\textbf{3837300}           &\textbf{23639.67}     \\
 bank heist & \textbf{46285.6} & \textbf{6262.20} & 17400   & 2352.93  & 23071.5& 3120.49           &1401              &187.68                &1380              &184.84       \\
 battle zone & 513360 & 1388.64 & 691700  & 1871.27  & \textbf{934134.88}& \textbf{2527.36}         &478830            &1295.20               &824360            &2230.29      \\
 beam rider & 128236.08 & 772.05 & 63600  & 381.80   & 300509.8 & 1812.19         &162100            &976.51                &\textbf{422390}            &\textbf{2548.07}      \\
 berzerk & 34134.8      & 1356.81 & 36200 & 1439.19  & \textbf{61507.83} & \textbf{2448.80}         &7607              &298.53                &14649             &579.46       \\
 bowling & 196.36       & 125.92  & 211.9 & 137.21   & \textbf{251.18}   & \textbf{165.76}          &201.9             &129.94                &205.2             &132.34       \\
 boxing  & 99.16        & 825.50  & 99.7  & 830.00   & \textbf{100}      & \textbf{832.50}          &\best{100}        &\best{832.50}         &\textbf{100}               &\textbf{832.50}       \\
 breakout & 795.36      & 2755.76 & 559.2 & 1935.76  & 790.4 & 2738.54                  &\best{864}        &\best{2994.10}                    &\textbf{864}             &\textbf{2994.10}      \\
 centipede & 532921.84  & 5347.83 & \textbf{577800} & \textbf{5799.95} & 412847.86& 4138.15         &155830            &1548.84               &195630            &1949.80\\
 chopper command&960648&14594.29&999900&15191.11&999900&15191.11                                    &\best{999999}&\best{15192.62}            &\textbf{999999}            &\textbf{15192.62}\\
 crazy climber & 312768   & 1205.59  & 313400 & 1208.11&\textbf{565909.85}&\textbf{2216.18}         &201000            &759.39                &241170            &919.76\\
 defender & 562106        & 3536.22  & 664100 & 4181.16  & 677642.78 & 4266.80                      &893110     &5629.27                      &\textbf{970540}            &\textbf{6118.89}\\
 demon attack & 143664.6  & 7890.07  & 143500 & 7881.02  & 143161.44 & 7862.41                      &675530     &37131.12       &\textbf{787985}                     &\textbf{43313.70}\\
 double dunk & 23.12      & 1896.36  & -14.1  & 204.55   & 23.93& 1933.18         &\textbf{24}      &\textbf{1936.36}                         &\textbf{24}                &\textbf{1936.36}\\
 enduro      & 2376.68    & 276.20   & 2000   & 232.42   & 2367.71   & 275.16                       &\best{14330}      &\best{1665.31}        &14300             &1661.82\\
 fishing derby & 81.96    & 328.28   & 32     & 233.84   & \textbf{86.97}& \textbf{337.75}          &59                &285.71                &65               &296.22\\
 freeway       & \textbf{34}       & \textbf{114.86}   & 28.5   & 96.28    & 32.59& 110.10          &\best{34}         &\best{114.86}         &\textbf{34}               &\textbf{114.86}\\
 frostbite    & 11238.4  & 261.70   & 206400 & 4832.76&\textbf{541280.88}&\textbf{12676.32}         &10485             &244.05                &11330	           &263.84\\
 gopher  & 122196        & 5658.66  & 113400 & 5250.47  & 117777.08 & 5453.59                       &\best{488830}     &\best{22672.63}       &473560           &21964.01\\
 gravitar & 6750         & 206.93   & 14200  & 441/32  &\textbf{19213.96}&\textbf{599.07}           &5905              &180.34                &5915             &180.66\\
 hero   & 37030.4        & 120.82   & 69400  & 229.44&\textbf{114736.26}&\textbf{381.58}            &38330             &125.18                &38225	            &124.83\\
 ice hockey & \textbf{71.56}      & \textbf{683.97}   &-4.1   & 58.68    & 63.64& 618.51            &44.94             &463.97                &47.11           &481.90\\
 jamesbond  & 23266      & 8486.85  & 26600  & 9704.53  & 135784.96 & 49582.16                      &594500            &217118.70             &\textbf{620780	}          &\textbf{226716.95}\\
 kangaroo   & 14112      & 471.34   & \textbf{35100}  & \textbf{1174.92}&24034.16& 803.96           &14500             &484.34                &14636           &488.90\\
 krull     & 145284.8    & 13460.12 & 127400&11784.73& 251997.31&23456.61         &97575             &8990.82                                 &\textbf{594540}          &\textbf{55544.92}\\
 kung fu master & 200176 & 889.40   & 212100 & 942.45   & 206845.82 & 919.07      &140440            &623.64                &\textbf{1666665}	         &\textbf{7413.57}\\
 montezuma revenge & 2504 & 52.68   & \textbf{10400}  & \textbf{218.80} &9352.01& 196.75            &3000              &63.11                 &2500            &52.60\\
 ms pacman  & 29928.2     & 445.81  & 40800  & 609.44& \textbf{63994.44}&\textbf{958.52}            &11536             &169.00                &11573           &169.55\\
 name this game & 45214.8 & 745.61  & 23900  & 375.35&\textbf{54386.77}&\textbf{904.94}             &34434             &558.34                &36296           &590.68\\
 phoenix & 811621.6       & 125.11  & 959100 &14786.66 &908264.15&14002.29         &894460            &13789.30              &\textbf{959580}	          &\textbf{14794.07}\\
 pitfall & 0              & 3.43    & 7800   & 119.97&\textbf{18756.01}&\textbf{283.66}             &0                 &3.43                  &-4.3            &3.36\\
 pong    & \textbf{21}             & \textbf{118.13}  & 19.6   & 114.16   & 20.67& 117.20           &\best{21}         &\best{118.13}         &\textbf{21}     &\textbf{118.13}\\
 private eye & 300        & 0.40    & \textbf{100000} & \textbf{143.75}& 79716.46&114.59            &15100             &21.68                 &15100           &21.68\\
 qbert  & 161000          & 1210.10 & 451900 & 3398.79&\textbf{580328.14}&\textbf{4365.06}          &27800             &207.93                &28657           &214.38\\
 riverraid & 34076.4      & 207.47  & 36700  & 224.10 & \textbf{63318.67}&\textbf{392.79}           &28075             &169.44                &28349           &171.17\\
 road runner & 498660     & 6365.59 & 128600 & 1641.52  & 243025.8&3102.24                          &878600            &11215.78       &\textbf{999999} &\textbf{12765.53}\\
 robotank   & \textbf{132.4}       & \textbf{1342.27} & 9.1    & 71.13 &127.32 &1289.90             &108.2             &1092.78               &113.4           &1146.39\\
 seaquest  & 999991.84    & 2381.55 & \textbf{1000000} & \textbf{2381.57}&999997.63&2381.56         &943910	           &2247.98        &\textbf{1000000}          &\textbf{2381.57}\\
 skiing & -29970.32       & -100.87 & -22977.9 & -46.08 & \textbf{-4202.6}  &\textbf{101.05}        &-6774             &80.90                 &-6025	       &86.77\\
 solaris & 4198.4         & 26.71   & 4700     & 31.23  & \textbf{44199.93}& \textbf{387.39}        &11074             &88.70                 &9105            &70.95\\
 space invaders & 55889   & 3665.48 & 43400    & 2844.22 & 48680.86 & 3191.48                       &140460            &9226.80               &\textbf{154380} &\textbf{10142.17}\\
 star gunner & 521728     & 5435.68 & 414600   &4318.13&\textbf{839573.53}&\textbf{8751.40}         &465750            &4851.72               &677590	         &7061.61\\
 surround    & \textbf{9.96}       & \textbf{120.97}  & -9.6     & 2.42    & 9.5&118.18             &-7.8              &13.33                 &2.606           &76.40\\
 tennis  & \textbf{24}             & \textbf{308.39}  & 10.2     & 219.35 & 23.84& 307.35           &\best{24}         &\best{308.39}         &\textbf{24}     &\textbf{308.39}             \\
 time pilot & 348932    & 20791.28 & 344700 & 20536.51&405425.31&24192.24         &216770            &12834.99              &\textbf{450810}	          &\textbf{26924.45}\\
 tutankham  & 393.64    & 244.71   & 191.1   & 115.04   & \textbf{2354.91}&\textbf{1500.33}         &423.9             &264.08                &418.2           &260.44\\
 up n down  & 542918.8  & 4860.17  & 620100  & 5551.77  & 623805.73 & 5584.98                       &\best{986440}     &\best{8834.45}        &966590          &8656.58\\
 venture    & 1992      & 167.75   & 1700    & 143.16   &\textbf{2623.71}  &\textbf{220.94}         &2035              &171.37                &2000	            &168.42\\
 video pinball & 483569.72 & 2737.00 & 965300 & 5463.58 &\textbf{992340.74}&\textbf{5616.63}        &925830            &5240.18               &978190          &5536.54\\
 wizard of wor & 133264 & 3164.81  & 106200  & 2519.35  &\textbf{157306.41}&\textbf{3738.20}        &64293             &1519.90               &63735           &1506.59\\
 yars revenge & 918854.32 & 1778.73 & 986000 & 1909.15  &\textbf{998532.37}&\textbf{1933.49}        &972000            &1881.96               &968090          &1874.36\\
 zaxxon & 181372        & 1983.85  & 111100  & 1215.07  &\textbf{249808.9} &\textbf{2732.54}        &109140            &1193.63               &216020	         &2362.89\\
\hline
MEAN HNS(\%)            &               & 3374.31  &         &  3169.90 &           & 4763.69  &     & \GDIImeanhns &      & \textbf{\GDIHmeanhns} \\
Learning Efficiency     &               &3.37E-09  &         & 9.06E-10  &        &  4.76E-10 &      & 3.91E-07       &      & \textbf{\GDIHmeanhnsle} \\
\hline
MEDIAN HNS(\%) &            & 1342.27   &         & 1208.11  &           & \textbf{1933.49}  &     & \GDIImedianhns &      & \GDIHmedianhns \\
Learning Efficiency     &    & 1.34E-09 &         & 3.45E-10  &        & 1.93E-10&      & 4.16E-08       &      & \textbf{\GDIHmedianhnsle} \\
\bottomrule
\end{tabular}
\caption{Score table of SOTA  model-free algorithms on HNS.}
\end{center}
\end{table}
\clearpage

\begin{table}[!hb]
\scriptsize
\begin{center}
\setlength{\tabcolsep}{1.0pt}
\begin{tabular}{c c c c c c c c c c c}
\toprule
 Games              & MuZero         & HNS(\%)      & DreamerV2 & HNS(\%)    & SimPLe             & HNS(\%)          & GDI-I$^3$     & HNS(\%) & GDI-H$^3$ & HNS(\%) \\
\midrule
Scale               & 20B            &              & 200M      &            & 1M               &                  & 200M     &      &  200M   &\\
\midrule    
 alien              & \textbf{741812.63}      & \textbf{10747.61}     &3483       & 47.18      &616.9     & 5.64    & 43384       & 625.45                    &48735	             &703.00      \\
 amidar             & \textbf{28634.39 }      & \textbf{1670.57    }  &2028       & 118.00     &74.3      & 4.00    & 1442        & 83.81                     &1065              &61.81 \\
 assault            & \textbf{143972.03}      & \textbf{27665.44}     &7679       & 1435.07    &527.2     & 58.66   & 63876       & 12250.50                  &97155	             &18655.23\\
 asterix            & 998425                  & 12036.40     &25669      & 306.98     &1128.3    & 11.07   & 759910      & 9160.41                   &\textbf{999999}   &\textbf{12055.38} \\
 asteroids          & 678558.64             & 1452.42      &3064       & 5.02       &793.6     & 0.16               &751970& 1609.72           &\textbf{760005}            &\textbf{1626.94}  \\
 atlantis           & 1674767.2             & 10272.64     &989207     & 6035.05    &20992.5   & 50.33              &3803000&23427.66          &\textbf{3837300}           &\textbf{23639.67}   \\
 bank heist         & 1278.98               & 171.17       &1043       & 139.23     &34.2      & 2.71               &\best{1401}  & \best{187.68  }           &1380              &184.84 \\
 battle zone        & \textbf{848623         }& \textbf{2295.95}      &31225      & 83.86      &4031.2    & 10.27   & 478830      & 1295.20                   &824360            &2230.29\\
 beam rider         & \textbf{454993.53}      & \textbf{2744.92}      &12413      & 72.75      &621.6     & 1.56    & 162100      & 976.51                    &422390            &2548.07\\
 berzerk            & \textbf{85932.6        }& \textbf{3423.18}      &751        & 25.02      &N/A       & N/A     & 7607        & 298.53                    &14649             &579.46\\
 bowling            & \textbf{260.13         }& \textbf{172.26 }      &48         & 18.10      &30        & 5.01    & 202         & 129.94                    &205.2             &132.34\\
 boxing             & \textbf{100}                   & \textbf{832.50}       &87         & 724.17     &7.8       & 64.17   & \best{100}  & \best{832.50  }    &\textbf{100}      &\textbf{832.50}  \\
 breakout           & \textbf{864}                   & \textbf{2994.10}      &350        & 1209.38    &16.4      & 51.04   & \best{864}  & \best{2994.10    } &\textbf{864}      &\textbf{2994.10}\\
 centipede          & \textbf{1159049.27}     & \textbf{11655.72}     &6601       & 45.44      &N/A       & N/A     & 155830      & 1548.84                   &195630            &1949.80\\
 chopper command    & 991039.7              & 15056.39     &2833       & 30.74      & 979.4    & 2.56    & \best{999999}& \best{15192.62}                     &\textbf{999999}   &\textbf{15192.62}\\
 crazy climber      & \textbf{458315.4}       & \textbf{1786.64    }  &141424     & 521.55     & 62583.6  & 206.81  & 201000      & 759.39                    &241170	            &919.76\\
 defender           & 839642.95             & 5291.18      & N/A       & N/A        & N/A      & N/A     & 893110      & 5629.27               &\textbf{970540}   &\textbf{6118.89}\\
 demon attack       & 143964.26             & 7906.55      & 2775     &144.20      & 208.1    & 3.08    & 675530      & 37131.12                &\textbf{787985}                     &\textbf{43313.70}\\
 double dunk        & 23.94          & 1933.64      & 22        &1845.45     & N/A      & N/A     & \textbf{24}          & \textbf{1936.36}                   &\textbf{24 }      &\textbf{1936.36}\\
 enduro             & 2382.44               & 276.87       & 2112     &245.44      & N/A      & N/A     & \best{14330}       & \best{1665.31}                 &14300             &1661.82\\
 fishing derby      & \textbf{91.16}          & \textbf{345.67     }  & 60        &286.77      &-90.7     & 1.89    & 59          & 285.71                    &65               &296.22\\
 freeway            & 33.03                 & 111.59       & \textbf{34}        &\textbf{114.86}      &16.7      & 56.42   & \best{34}      & \best{114.86 }  &\textbf{34}      &\textbf{114.86}\\
 frostbite          & \textbf{631378.53}      & \textbf{14786.59}     & 15622    &364.37      &236.9     & 4.02    & 10485       & 244.05                     &11330	            &263.84\\
 gopher             & 130345.58             & 6036.85      & 53853    &2487.14     &596.8     & 15.74   & \best{488830}      & \best{22672.6}                 &473560           &21964.01\\
 gravitar           & \textbf{6682.7     }    & \textbf{204.81     }  & 3554     &106.37      &173.4     & 0.01    & 5905        & 180.34                     &5915             &180.66\\
 hero               & \textbf{49244.11}       & \textbf{161.81     }  & 30287    &98.19       &2656.6    & 5.47    & 38330       & 125.18                     &38225	            &124.83\\
 ice hockey         & \textbf{67.04      }    & \textbf{646.61     }  & 29        &332.23      &-11.6     & -3.31   & 44.94       & 463.97                    &47.11           &481.90\\
 jamesbond          & 41063.25              & 14986.94     &  9269     &3374.73     &100.5     & 26.11   & 594500      & 217118.70              &\textbf{620780	}  &\textbf{226716.95}\\
 kangaroo           & \textbf{16763.6        }& \textbf{560.23     }  & 11819     &394.47      &51.2      & -0.03   & 14500       & 484.34                    &14636           &488.90\\
 krull              & 269358.27      & 25082.93     & 9687     &757.75      &2204.8    & 56.84   & 97575       & 8990.82                    &\textbf{594540}          &\textbf{55544.92}\\
 kung fu master     & 204824         & 910.08       & 66410    &294.30      &14862.5   & 64.97   & 140440      & 623.64                     &\textbf{1666665}	          &\textbf{7413.57}\\
 montezuma revenge  & 0                     & 0.00         & 1932     &40.65       &N/A       & N/A     & \best{3000}        & \best{63.11  }                 &2500            &52.60\\
 ms pacman          & \textbf{243401.1 }      & \textbf{3658.68    }  & 5651     &80.43       &1480      & 17.65   & 11536       & 169.00                     &11573           &169.55\\
 name this game     & \textbf{157177.85}      & \textbf{2690.53    }  & 14472    &211.57      &2420.7    & 2.23    & 34434       & 558.34                     &36296           &590.68\\
 phoenix            & 955137.84      & 14725.53     & 13342     &194.11      &N/A       & N/A     & 894460      & 13789.30                  &\textbf{959580 }         &	\textbf{14794.07}\\
 pitfall            &\textbf{ 0}                     & \textbf{3.43}         & -1        &3.41        &N/A       & N/A     & \best{0}       & \best{3.43  }   &-4.3            &3.36\\
 pong               & \textbf{21}                    & \textbf{118.13}       & 19        &112.46      & 12.8     & 94.90   & \best{21}      & \best{118.13}   &\textbf{21}     &\textbf{118.13}   \\
 private eye        & \textbf{15299.98 }      & \textbf{21.96  }      & 158       &0.19        & 35       & 0.01    & 15100       & 21.68                     &15100           &21.68\\
 qbert              & \textbf{72276          }& \textbf{542.56 }      & 162023    &1217.80     & 1288.8   & 8.46    & 27800       & 207.93                    &28657           &214.38\\
 riverraid          & \textbf{323417.18}      & \textbf{2041.12}      & 16249    &94.49       & 1957.8   & 3.92    & 28075       & 169.44                     &28349           &171.17\\
 road runner        & 613411.8              & 7830.48      & 88772    &1133.09     & 5640.6   & 71.86   & 878600      & 11215.78                              &\textbf{999999	} &\textbf{12765.53}\\
 robotank           & \textbf{131.13}         & \textbf{1329.18}      & 65        &647.42      & N/A      & N/A     & 108         & 1092.78                   &113.4           &1146.39\\
 seaquest           & 999976.52             & 2381.51      & 45898    &109.15      & 683.3    & 1.46    &943910	           &2247.98             &\textbf{1000000}          &\textbf{2381.57}\\
 skiing             & -29968.36      & -100.86     & -8187    &69.83       & N/A      & N/A     & -6774       & 80.90                       &\textbf{-6025}  &\textbf{86.77}\\
 solaris            & 56.62                 & -10.64       & 883       &-3.19       & N/A      & N/A     & \best{11074}       & \best{88.70   }               &9105            &70.95\\
 space invaders     & 74335.3               & 4878.50      & 2611      &161.96      & N/A      & N/A     & 140460     & 9226.80                 &\textbf{154380}          &\textbf{10142.17}\\
 star gunner        & 549271.7       & 5723.01      & 29219    &297.88      & N/A      & N/A     & 465750      & 4851.72                    &\textbf{677590}          &\textbf{7061.61}\\
 surround           & \textbf{9.99       }    & \textbf{121.15     }  & N/A       &N/A         & N/A      & N/A     & -7.8          & 13.33                   &2.606           &76.40 \\
 tennis             & 0       & 153.55  & 23        &301.94      & N/A      & N/A     & \textbf{24}          & \textbf{308.39}                                &\textbf{24}              &\textbf{308.39}  \\
 time pilot         & \textbf{476763.9}       & \textbf{28486.90}     & 32404    &1735.96     & N/A      & N/A     & 216770      & 12834.99                   &450810	          &26924.45 \\
 tutankham          & \textbf{491.48     }    & \textbf{307.35     }  & 238       &145.07      & N/A      & N/A     & 424         & 264.08                    &418.2           &260.44 \\
 up n down          & 715545.61             & 6407.03      & 648363   &5805.03     & 3350.3   & 25.24   & \best{986440}      & \best{8834.45}                 &966590          &8656.58\\
 venture            & 0.4                   & 0.03         & 0         &0.00        & N/A      & N/A     & \best{2035}        & \best{171.37 }                &2000	            &168.42\\
 video pinball      & \textbf{981791.88}      & \textbf{5556.92}      & 22218    &125.75      & N/A      & N/A     & 925830      & 5240.18                    &978190          &5536.54\\
 wizard of wor      & \textbf{197126         }& \textbf{4687.87}      & 14531    &333.11      & N/A      & N/A     & 64439       & 1523.38                    &63735           &1506.59\\
 yars revenge       & 553311.46             & 1068.72      & 20089    &33.01       & 5664.3   & 4.99    & \best{972000}      & \best{1881.96}                 &968090          &1874.36\\
 zaxxon             & \textbf{725853.9}       & \textbf{7940.46}      & 18295    &199.79      & N/A      & N/A     & 109140      & 1193.63                    &216020	          &2362.89\\
\hline    
MEAN HNS(\%)        &                & 4996.20      &           &  631.17   &           & 25.3    &             & \GDIImeanhns&      & \textbf{\GDIHmeanhns} \\
Learning Efficiency     &     &2.50E-09  &         & 3.16E-08  &        & 2.53E-07 &      & 3.91E-07       &      & \textbf{\GDIHmeanhnsle} \\
\hline
MEDIAN HNS(\%)      &                & \textbf{2041.12}      &           & 161.96    &           & 5.55    &             & \GDIImedianhns &      & \GDIHmedianhns \\
Learning Efficiency     &     & 1.02E-09 &         & 8.10E-09  &        & 5.55E-08&      & 4.16E-08       &      & \textbf{\GDIHmedianhnsle} \\
\bottomrule
\end{tabular}
\caption{Score table of SOTA model-based algorithms on HNS.}
\end{center}
\end{table}
\clearpage

\begin{table}[!hb]
\scriptsize
\begin{center}
\setlength{\tabcolsep}{1.0pt}
\begin{tabular}{c c c c c c c c c}            
\toprule
 Games        & Muesli & HNS(\%)      & Go-Explore              & HNS(\%)                     & GDI-I$^3$ & HNS(\%)               & GDI-H$^3$ & HNS(\%) \\
\midrule
Scale         &  200M   &            & 10B                     &                             & 200M              &                &  200M   &\\
\midrule    
 alien        &139409          &2017.12                   &\textbf{959312}       &\textbf{13899.77}              & 43384             &625.45            &48735	             &703.00              \\
 amidar       &\textbf{21653}  &\textbf{1263.18}          &19083                 &1113.22                        & 1442              &83.81             &1065              &61.81           \\
 assault      &36963           &7070.94                   &30773                 &5879.64                        & 63876      &12250.50   &\textbf{97155	}    &\textbf{18655.23}       \\
 asterix      &316210          &3810.30                   &999500       &12049.37              & 759910            &9160.41           &\textbf{999999}    &\textbf{12055.38}\\
 asteroids    &484609          &1036.84                   &112952                &240.48                         & 751970     &1609.72    &\textbf{760005}            &\textbf{1626.94}       \\
 atlantis     &1363427         &8348.18                   &286460                &1691.24                        & 3803000    &23427.66   &\textbf{3837300}           &\textbf{23639.67}       \\
 bank heist   &1213            &162.24                    &\textbf{3668}         &\textbf{494.49}                & 1401              &187.68            &1380              &184.84\\
 battle zone  &414107          &1120.04                   &\textbf{998800}       &\textbf{2702.36}               & 478830            &1295.20           &824360            &2230.29\\
 beam rider   &288870          &1741.91                   &371723       &2242.15               & 162100            &976.51            &\textbf{422390}   &\textbf{2548.07}\\
 berzerk      &44478           &1769.43                   &\textbf{131417}       &\textbf{5237.69}               & 7607              &298.53            &14649             &579.46\\
 bowling      &191             &122.02                    &\textbf{247}           &\textbf{162.72}                & 202               &129.94           &205.2             &132.34\\
 boxing       &99              &824.17                    &91                     &757.50                         & \best{100}        &\best{832.50}    &\textbf{100}      &\textbf{832.50}        \\
 breakout     &791             &2740.63                   &774                    &2681.60                        & \best{864}        &\best{2994.10}   &\textbf{864}      &\textbf{2994.10}        \\
 centipede    &\textbf{869751} &\textbf{8741.20}          &613815                &6162.78               & 155830            &1548.84                    &195630            &1949.80\\
 chopper command &101289       &1527.76            &996220                &15135.16                       & \best{999999}     &\best{15192.62}          &\textbf{999999}   &\textbf{15192.62}\\
 crazy climber   &175322       &656.88             &235600       &897.52                & 201000            &759.39                   &\textbf{241170}	            &\textbf{919.76}\\
 defender        &629482       &3962.26            &N/A                    &N/A                            & 893110     &5629.27                        &\textbf{970540}   &\textbf{6118.89}\\
 demon attack    &129544       &7113.74            &239895                 &13180.65                       & 675530     &37131.12         &\textbf{787985}                     &\textbf{43313.70}\\
 double dunk     &-3           &709.09             &\textbf{24}                     &\textbf{1936.36}                        & \best{24}         &\best{1936.36}          &\textbf{24}       &\textbf{1936.36}\\
 enduro          &2362         &274.49             &1031                   &119.81                         & \best{14330}      &\best{1665.31}          &14300             &1661.82\\
 fishing derby   &51           &269.75             &\textbf{67}            &\textbf{300.00}                & 59                &285.71                  &65               &296.22\\
 freeway         &33           &111.49             &\textbf{34}            &\textbf{114.86}                & \best{34}         &\best{114.86}           &\textbf{34}        &\textbf{114.86}\\
 frostbite       &301694       &7064.73            &\textbf{999990}       &\textbf{23420.19}              & 10485             &244.05                   &11330	            &263.84\\
 gopher          &104441       &4834.72            &134244                &6217.75                        & \best{488830}     &\best{22672.63}          &473560           &21964.01\\
 gravitar        &11660        &361.41             &\textbf{13385}        &\textbf{415.68}                & 5905              &180.34                   &5915             &180.66\\
 hero            &37161        &121.26            &37783                  &123.34                         & \textbf{38330}      &\textbf{125.18}            &38225	   &124.83\\
 ice hockey      &25           &299.17             &33                     &365.29                         & 44.94         &463.97        &\textbf{47.11}           &\textbf{481.90}    \\
 jamesbond       &19319        &7045.29            &200810                &73331.26                       & 594500     &217118.70         &\textbf{620780	}          &\textbf{226716.95}\\
 kangaroo        &14096        &470.80             &\textbf{24300}        &\textbf{812.87}                & 14500             &484.34                   &14636           &488.90\\
 krull           &34221        &3056.02            &63149                 &5765.90                        & 97575      &8990.82           &\textbf{594540}          &\textbf{55544.92}\\
 kung fu master  &134689       &598.06             &24320                 &107.05                         & 140440     &623.64            &\textbf{1666665	}          &\textbf{7413.57}\\
 montezuma revenge  &2359      &49.63               &\textbf{24758}        &\textbf{520.86}                & 3000              &63.11                   &2500            &52.60\\
 ms pacman          &65278     &977.84              &\textbf{456123}       &\textbf{6860.25}               & 11536             &169.00                  &11573           &169.55\\
 name this game     &105043    &1784.89              &\textbf{212824}       &\textbf{3657.16}               & 34434             &558.34                 &36296           &590.68\\
 phoenix        &805305        &12413.69                    &19200                 &284.50                  & 894460     &13789.30 &\textbf{959580	}          &\textbf{14794.07}   \\
 pitfall        &0             &3.43                    &\textbf{7875}          &\textbf{121.09}                & 0                 &3.43               &-4.3            &3.36\\
 pong           &20            &115.30                  &\textbf{21}            &\textbf{118.13}                & \best{21}         &\best{118.13}      &\textbf{21}              &\textbf{118.13}      \\
 private eye    &10323         &14.81                   &\textbf{69976}        &\textbf{100.58}                & 15100             &21.68               &15100           &21.68\\
 qbert          &157353        &1182.66                 &\textbf{999975}       &\textbf{7522.41}               & 27800             &207.93              &28657           &214.38\\
 riverraid      &\textbf{47323}&\textbf{291.42}         &35588                 &217.05                & 28075             &169.44                       &28349           &171.17\\
 road runner    &327025        &4174.55                 &999900        &12764.26              & 878600            &11215.78           &\textbf{999999}	          &\textbf{12765.53}\\
 robotank       &59            &585.57                  &\textbf{143}           &\textbf{1451.55}               & 108               &1092.78            &113.4           &1146.39\\
 seaquest       &815970        &1943.26                 &539456                &1284.68               &943910	           &2247.98               &\textbf{1000000}          &\textbf{2381.57}\\
 skiing         &-18407        &-10.26                  &\textbf{-4185}        &\textbf{101.19}                & -6774             &80.90               &-6025	         &86.77\\
 solaris        &3031          &16.18                   &\textbf{20306}        &\textbf{171.95}                & 11074             &88.70               &9105            &70.95\\
 space invaders &59602         &3909.65                &93147                 &6115.54                        & 140460     &9226.80       &\textbf{154380}          &\textbf{10142.17}\\
 star gunner    &214383        &2229.49                &609580       &6352.14               & 465750     &4851.72                     &\textbf{677590}	          &\textbf{7061.61}\\
 surround       &\textbf{9}    &\textbf{115.15}                 &N/A                    &N/A                  & -8        &13.33                        &2.606           &76.40\\
 tennis         &12            &230.97                 &\best{24}              &\best{308.39}                  & \best{24}         &\best{308.39}       &\textbf{24}    &\textbf{308.39}     \\
 time pilot     &\textbf{359105} &\textbf{21403.71}                   &183620                &10839.32       & 216770     &12834.99                     &450810	          &26924.45\\
 tutankham      &252           &154.03                 &\textbf{528}           &\textbf{330.73}                & 424               &264.08              &418.2           &260.44\\
 up n down      &649190        &5812.44                &553718                &4956.94                        & \best{986440}     &\best{8834.45}       &966590          &8656.58    \\
 venture        &2104          &177.18                 &\textbf{3074}         &\textbf{258.86}                & 2035              &171.37               &2000	            &168.42\\
 video pinball  &685436        &3879.56                &\textbf{999999}       &\textbf{5659.98}               & 925830            &5240.18              &978190          &5536.54\\
 wizard of wor  &93291         &2211.48                &\textbf{199900}       &\textbf{4754.03}               & 64293             &1519.90              &63735           &1506.59\\
 yars revenge   &557818        &1077.47                &\textbf{999998}       &\textbf{1936.34}               & 972000            &1881.96              &968090          &1874.36\\
 zaxxon         &65325         &714.30                 &18340                 &200.28                         & 109140     &1193.63       &\textbf{216020	}          &\textbf{2362.89}    \\
\hline    
MEAN HNS(\%)        & & 2538.66           &                       & 4989.94                       &            &  \GDIImeanhns &      & \textbf{\GDIHmeanhns} \\
Learning Efficiency & & 1.27E-07          &                       & 4.99E-09                      &      & 3.91E-07       &      & \textbf{\GDIHmeanhnsle} \\
\hline
MEDIAN HNS(\%)      & & 1077.47           &                       & \textbf{1451.55}              &            & \GDIImedianhns   &      & \GDIHmedianhns \\
Learning Efficiency & & 5.39E-08          &                       & 1.45E-09                      &      & 4.16E-08       &      & \textbf{\GDIHmedianhnsle} \\
\bottomrule
\end{tabular}
\caption{Score table of other SOTA algorithms on HNS.}
\end{center}
\end{table}

\clearpage

\section{Atari Games Table of Scores Based on Human World Records}
\label{app: Atari Games Table of Scores Based on Human World Records}
In this part, we detail the raw score of several representative SOTA algorithms, including the SOTA 200M model-free algorithms, SOTA 10B+ model-free algorithms, SOTA model-based algorithms, and other SOTA algorithms.\footnote{200M and 10B+ represent the training scale.} Additionally, we calculate the human world records normalized world score (HWRNS) of each game with each algorithm. First of all, we demonstrate the sources of the scores that we used.
Random scores  are from \citep{agent57}.
Human world records (HWR) are form \citep{dreamerv2,atarihuman}.
Rainbow's scores are from \citep{rainbow}.
IMPALA's scores are from \citep{impala}.
LASER's scores are from \citep{laser}, no sweep at 200M.
As there are many versions of R2D2 and NGU, we use original papers'.
R2D2's scores are from \citep{r2d2}.
NGU's scores are from \citep{ngu}.
Agent57's scores are from \citep{agent57}.
MuZero's scores are from \citep{muzero}.
DreamerV2's scores are from \citep{dreamerv2}.
SimPLe's scores are form \citep{modelbasedatari}.
Go-Explore's scores are form \citep{goexplore}.
Muesli's scores are form \citep{muesli}.
In the following we detail the raw scores and HWRNS of each algorithm on 57 Atari games.
\clearpage
\begin{table}[!hb]
\scriptsize
\begin{center}
\setlength{\tabcolsep}{1.0pt}
\begin{tabular}{c c c c c c c c c c c c c}
\toprule
Games               & RND       & HWR       & RAINBOW  & HWRNS(\%) & IMPALA  & HWRNS(\%)  & LASER  & HWRNS(\%) & GDI-I$^3$ & HWRNS(\%)  & GDI-H$^3$ & HWRNS(\%) \\
\midrule
Scale               &           &           & 200M     &           &  200M   &            & 200M    &           &  200M    &            &    200M   &\\
\midrule
 alien              & 227.8     & \textbf{251916}    & 9491.7   &3.68    & 15962.1    & 6.25       & 976.51  & 14.04     &43384             &17.15  &48735             &19.27   \\
 amidar             & 5.8       & \textbf{104159}    & 5131.2   &4.92    & 1554.79    & 1.49       & 1829.2  & 1.75      &1442              &1.38   &1065              &1.02          \\
 assault            & 222.4     & 8647             & 14198.5  &165.90  & 19148.47   & 224.65     & 21560.4 & 253.28    &63876        &755.57  &\textbf{97155}             &\textbf{1150.59} \\
 asterix            & 210       & \textbf{1000000}   & 428200   &42.81   & 300732     & 30.06      & 240090  & 23.99     &759910            &75.99  &999999            &100.00 \\
 asteroids          & 719       & \textbf{10506650}  & 2712.8   &0.02    & 108590.05  & 1.03       & 213025  & 2.02      &751970            &7.15   &760005            &7.23\\
 atlantis           & 12850     & \textbf{10604840}  & 826660   &7.68    & 849967.5   & 7.90       & 841200  & 7.82      &3803000           &35.78  &3837300           &36.11\\
 bank heist         & 14.2      & \textbf{82058}     & 1358     &1.64    & 1223.15    & 1.47       & 569.4   & 0.68      &1401              &1.69   &1380              &1.66  \\
 battle zone        & 236       & 801000    & 62010    &7.71    & 20885      & 2.58       & 64953.3 & 8.08      &478830            &59.77  &\textbf{824360}            &102.92 \\
 beam rider         & 363.9     & \textbf{999999}    & 16850.2  &1.65    & 32463.47   & 3.21       & 90881.6 & 9.06      &162100            &16.18  &422390            &42.22   \\
 berzerk            & 123.7     & \textbf{1057940}   & 2545.6   &0.23    & 1852.7     & 0.16       & 25579.5 & 2.41      &7607                       &0.71   &14649             &1.37          \\
 bowling            & 23.1      & \textbf{300}       & 30       &2.49    & 59.92      & 13.30      & 48.3    & 9.10      &201.9             &64.57  &205.2             &65.76   \\
 boxing             & 0.1       & \textbf{100}       & 99.6     &99.60   & 99.96      & 99.96      & \textbf{100}     & \textbf{100.00}    &\best{100}        &\best{100.00 } &\textbf{100}               &\textbf{100.00}    \\
 breakout           & 1.7       & \textbf{864}       & 417.5    &48.22   & 787.34     & 91.11      & 747.9   & 86.54     &\best{864}        &\best{100.00 } &\textbf{864}             &\textbf{100.00 }  \\
 centipede          & 2090.9    & \textbf{1301709}   & 8167.3   &0.47    & 11049.75   & 0.69       & 292792  & 22.37     &155830            &11.83          &195630            &14.89 \\
 chopper command    & 811       & \textbf{999999}    & 16654    &1.59    & 28255      & 2.75       & 761699  & 76.15     &\best{999999}     &\best{100.00 } &\textbf{999999}            &\textbf{100.00}\\
 crazy climber      & 10780.5   & 219900    & 168788.5 &75.56   & 136950     & 60.33      & 167820  & 75.10     &201000            &90.96          &\textbf{241170}            &\textbf{110.17}\\
 defender           & 2874.5    & \textbf{6010500}   & 55105    &0.87    & 185203     & 3.03       & 336953  & 5.56      &893110            &14.82          &970540            &16.11\\
 demon attack       & 152.1     & \textbf{1556345}   & 111185   &7.13    & 132826.98  & 8.53       & 133530  & 8.57      &675530            &43.40          &\textbf{787985}   &\textbf{50.63}\\
 double dunk        & -18.6     & 21        & -0.3     &46.21   & -0.33      & 46.14      & 14      & 82.32     &\best{24}                  &\best{107.58 } &\textbf{24}                &\textbf{107.58}  \\
 enduro             & 0         & 9500      & 2125.9   &22.38   & 0          & 0.00       & 0       & 0.00      &\best{14330}               &\best{150.84  }&14300             &150.53  \\
 fishing derby      & -91.7     & \textbf{71}        & 31.3     &75.60   & 44.85      & 83.93      & 45.2    & 84.14     &59                &92.89          &65               &96.31\\
 freeway            & 0         & \textbf{38}        & 34       &89.47   & 0          & 0.00       & 0       & 0.00      &34                &89.47          &34               &89.47\\
 frostbite          & 65.2      & \textbf{454830}    & 9590.5   &2.09    & 317.75     & 0.06       & 5083.5  & 1.10      &10485             &2.29           &11330            &2.48 \\
 gopher             & 257.6     & 355040    & 70354.6  &19.76   & 66782.3    & 18.75      & 114820.7& 32.29     &\best{488830}              &\best{137.71 } &473560           &133.41 \\
 gravitar           & 173       & \textbf{162850}    & 1419.3   &0.77    & 359.5      & 0.11       & 1106.2  & 0.57      &5905              &3.52           &5915             &3.53\\
 hero               & 1027      & \textbf{1000000}   & 55887.4  &5.49    & 33730.55   & 3.27       & 31628.7 & 3.06      &38330                      &3.73           &38225            &3.72  \\
 ice hockey         & -11.2     & 36        & 1.1      &26.06   & 3.48       & 31.10      & 17.4    & 60.59     &44.92              &118.94      &\textbf{47.11}           &\textbf{123.54} \\
 jamesbond          & 29        & 45550     & 19809    &43.45   & 601.5      & 1.26       & 37999.8 & 83.41     &594500              &1305.93 &\textbf{620780}          &\textbf{1363.66}\\
 kangaroo           & 52        & \textbf{1424600}   & 14637.5  &1.02    & 1632       & 0.11       & 14308   & 1.00      &14500                      &1.01           &14636           &1.02  \\
 krull              & 1598      & 104100    & 8741.5   &6.97    & 8147.4     & 6.39       & 9387.5  & 7.60      &97575             &93.63          &\textbf{594540}          &\textbf{578.47}\\
 kung fu master     & 258.5     & 1000000   & 52181    &5.19    & 43375.5    & 4.31       & 607443  & 60.73     &140440            &14.02          &\textbf{1666665}          &\textbf{166.68}\\
 montezuma revenge  &0          & \textbf{1219200}   & 384      &0.03    & 0          & 0.00       & 0.3     & 0.00      &3000              &0.25           &2500            &0.21\\
 ms pacman          & 307.3     & \textbf{290090}    & 5380.4   &1.75    & 7342.32    & 2.43       & 6565.5  & 2.16      &11536             &3.87           &11573           &3.89\\
 name this game     & 2292.3    & 25220     & 13136    &47.30   & 21537.2    & 83.94      & 26219.5 & 104.36    &34434               &140.19  &\textbf{36296}  &\textbf{148.31}    \\
 phoenix            & 761.5     & \textbf{4014440}   & 108529   &2.69    & 210996.45  & 5.24       & 519304  & 12.92     &894460            &22.27          &959580          &23.89\\
 pitfall            & -229.4    & \textbf{114000}    & 0        &0.20    & -1.66      & 0.20       & -0.6    & 0.20      &\best{0    }      &0.20           &-4.3            &0.20\\
 pong               & -20.7     & \textbf{21}        & 20.9     &99.76   & 20.98      & 99.95      & \textbf{21}      & \textbf{100.00}    &\best{21   }      &\best{100.00 } &\textbf{21}     &\textbf{100.00}    \\
 private eye        & 24.9      & \textbf{101800}    & 4234     &4.14    & 98.5       & 0.07       & 96.3    & 0.07      &15100             &14.81          &15100           &14.81\\
 qbert              & 163.9     & \textbf{2400000}   & 33817.5  &1.40    & 351200.12  & 14.63      & 21449.6 & 0.89      &27800             &1.15           &28657           &1.19\\
 riverraid          & 1338.5    & \textbf{1000000}   & 22920.8  &2.16    & 29608.05   & 2.83       & 40362.7 & 3.91      &28075             &2.68           &28349           &2.70\\
 road runner        & 11.5      & \textbf{2038100}   & 62041    &3.04    & 57121      & 2.80       & 45289   & 2.22      &878600            &43.11          &999999          &49.06\\
 robotank           & 2.2       & 76        & 61.4     &80.22   & 12.96      & 14.58      & 62.1    & 81.17     &108.2               &143.63  &\textbf{113.4}           &\textbf{150.68}\\
 seaquest           & 68.4      & 999999    & 15898.9  &1.58    & 1753.2     & 0.17       & 2890.3  & 0.28      &943910	             &94.39&\textbf{1000000}          &\textbf{100.00}\\
 skiing             & -17098    & \textbf{-3272}     & -12957.8 &29.95   & -10180.38  & 50.03      & -29968.4& -93.09    &-6774             &74.67          &-6025	          &86.77\\
 solaris            & 1236.3    & \textbf{111420}    & 3560.3   &2.11    & 2365       & 1.02       & 2273.5  & 0.94      &11074             &8.93           &9105            &7.14\\
 space invaders     & 148       & \textbf{621535 }   & 18789    &3.00    & 43595.78   & 6.99       & 51037.4 & 8.19      &140460            &22.58          &154380          &24.82\\
 star gunner        & 664       & 77400     & 127029   &164.67  & 200625     & 260.58     & 321528  & 418.14    &465750              &606.09  &\textbf{677590} &\textbf{882.15}\\
 surround           & -10       & 9.6       & \textbf{9.7}      &\textbf{100.51}  & 7.56       & 89.59      & 8.4     & 93.88  &-7.8        &11.22          &2.606           &64.32\\
 tennis             & -23.8     & 21        & 0        &53.13   & 0.55       & 54.35      & 12.2    & 80.36     &\best{24       }           &\best{106.70    } &\textbf{24}  &\textbf{106.70}\\
 time pilot         & 3568      & 65300     & 12926    &15.16   & 48481.5    & 72.76      & 105316  & 164.82    &216770              &345.37         &\textbf{450810} &\textbf{724.49}\\
 tutankham          & 11.4      & \textbf{5384}      & 241      &4.27    & 292.11     & 5.22       & 278.9   & 4.98      &423.9             &7.68           &418.2           &7.57\\
 up n down          & 533.4     & 82840     & 125755   &152.14  & 332546.75  & 403.39     & 345727  & 419.40    &\best{986440}              &\best{1197.85  } &966590          &1173.73  \\
 venture            & 0         & \textbf{38900}     & 5.5      &0.01    & 0          & 0.00       & 0       & 0.00      &2000              &5.23           &2000            &5.14\\
 video pinball      & 0         & \textbf{89218328}  & 533936.5 &0.60    & 572898.27  & 0.64       & 511835  & 0.57      &925830            &1.04           &978190          &1.10\\
 wizard of wor      & 563.5     & \textbf{395300}    & 17862.5  &4.38    & 9157.5     & 2.18       & 29059.3 & 7.22      &64439             &16.14          &63735           &16.00\\
 yars revenge       & 3092.9    & \textbf{15000105}  & 102557   &0.66    & 84231.14   & 0.54       & 166292.3& 1.09      &972000            &6.46           &968090          &6.43\\
 zaxxon             & 32.5      & 83700     & 22209.5  &26.51   & 32935.5    & 39.33      & 41118   & 49.11     &109140              &130.41  &\textbf{216020}          &\textbf{258.15}\\
\hline
MEAN HWRNS(\%)      & 0.00      & 100.00    &          & 28.39  &            & 34.52  &        & 45.39 &      & \GDIImeanHWRNS  &      & \textbf{\GDIHmeanHWRNS} \\
Learning Efficiency &     0.00 & N/A  &         & 1.42E-09 &         & 1.73E-09  &        & 2.27E-09 &      & 5.90E-09       &      & \textbf{\GDIHmeanHWRNSle} \\
\hline   
MEDIAN HWRNS(\%)    & 0.00      & \textbf{100.00}    &          & 4.92   &            & 4.31  &        & 8.08  &      &  \GDIImedianHWRNS  &      & \GDIHmedianHWRNS  \\
Learning Efficiency & 0.00   & N/A   &         & 2.46E-10 &         & 2.16E-10  &        & 4.04E-10 &      & 1.79E-09       &      & \textbf{\GDIHmedianHWRNSle} \\
\hline   
HWRB    & 0      &\textbf{57}    &          & 4   &            & 3  &        & 7  &      &  \GDIIHWRB  &      & \GDIHHWRB\\
\bottomrule
\end{tabular}
\caption{Score table of SOTA 200M model-free algorithms on HWRNS}
\label{tab: 200 model-free Atari Games Table of Scores Based on Human World Records}
\end{center}
\end{table}
\clearpage
\begin{table}[!hb]
\scriptsize
\begin{center}
\setlength{\tabcolsep}{1.0pt}
\begin{tabular}{c c c c c c c c c c c }
\toprule
 Games & R2D2 & HWRNS(\%) & NGU & HWRNS(\%) & AGENT57 & HWRNS(\%) & GDI-I$^3$ & HWRNS(\%) & GDI-H$^3$ & HWRNS(\%) \\
\midrule
Scale  & 10B   &        & 35B &         & 100B     &        & 200M &  &    200M   &\\
\midrule
 alien              & 109038.4          & 43.23       & 248100          & 98.48          & \textbf{297638.17}   &\textbf{118.17}         &43384             &17.15   &48735             &19.27    \\
 amidar             & 27751.24          & 26.64       & 17800           & 17.08          & \textbf{29660.08}    &\textbf{28.47}          &1442              &1.38    &1065              &1.02    \\
 assault            & 90526.44          & 1071.91     & 34800           & 410.44         & 67212.67             &795.17         &63876             &755.57  &\textbf{97155}             &\textbf{1150.59}    \\
 asterix            & 999080   & 99.91       & 950700          & 95.07          & 991384.42            &99.14          &759910            &75.99   &\textbf{999999}            &\textbf{100.00}    \\
 asteroids          & 265861.2          & 2.52        & 230500          & 2.19           & 150854.61            &1.43           &751970     &7.15    &\textbf{760005}            &\textbf{7.23}    \\
 atlantis           & 1576068           & 14.76       & 1653600         & 15.49          & 1528841.76           &14.31          &3803000    &35.78   &\textbf{3837300}           &\textbf{36.11}    \\
 bank heist         & \textbf{46285.6}  & \textbf{56.40}       & 17400           & 21.19          & 23071.5              &28.10          &1401              &1.69    &1380              &1.66    \\
 battle zone        & 513360            & 64.08       & 691700          & 86.35          & \textbf{934134.88}   &\textbf{116.63}         &478830            &59.77   &824360            &102.92    \\
 beam rider         & 128236.08         & 12.79       & 63600           & 6.33           & 300509.8    &30.03          &162100            &16.18   &\textbf{422390}            &\textbf{42.22}    \\
 berzerk            & 34134.8           & 3.22        & 36200           & 3.41           & \textbf{61507.83}    &\textbf{5.80 }          &7607              &0.71    &14649             &1.37    \\
 bowling            & 196.36            & 62.57       & 211.9           & 68.18          & \textbf{251.18}      &\textbf{82.37}          &201.9             &64.57   &205.2             &65.76    \\
 boxing             & 99.16             & 99.16       & 99.7            & 99.70          & \textbf{100}         &\textbf{100.00}         &\best{100}        &\textbf{100.00} &\textbf{100}       &\textbf{100.00}     \\
 breakout           & 795.36            & 92.04       & 559.2           & 64.65          & 790.4                &91.46          &\best{864}        &\textbf{100.00}  &\textbf{864}             &\textbf{100.00}    \\
 centipede          & 532921.84         & 40.85       & \textbf{577800} & \textbf{44.30}          & 412847.86            &31.61          &155830            &11.83   &195630            &14.89    \\
 chopper command    &960648             & 96.06       &999900           & 99.99          &999900                &99.99          &\best{999999}     &\textbf{100.00}  &\textbf{999999}            &\textbf{100.00}    \\
 crazy climber      & 312768            & 144.41      & 313400          & 144.71         &\textbf{565909.85}    &\textbf{265.46}         &201000            &90.96   &241170            &110.17    \\
 defender           & 562106            & 9.31        & 664100          & 11.01          & 677642.78            &11.23          &893110     &14.82   &\textbf{970540}            &\textbf{16.11}    \\
 demon attack       & 143664.6          & 9.22        & 143500          & 9.21           & 143161.44            &9.19           &675530     &43.40    &\textbf{787985}   &\textbf{50.63}   \\
 double dunk        & 23.12             & 105.35      & -14.1           & 11.36          & 23.93       &107.40         &\textbf{24}                &\textbf{107.58}  &\textbf{24}                &\textbf{107.58}    \\
 enduro             & 2376.68           & 25.02       & 2000            & 21.05          & 2367.71              &24.92          &\best{14330}      &\textbf{150.84}  &14300             &150.53    \\
 fishing derby      & 81.96             & 106.74      & 32              & 76.03          & \textbf{86.97}       &\textbf{109.82}         &59                &92.89   &65                &96.31    \\
 freeway            & \textbf{34}       & \textbf{89.47}       & 28.5            & 75.00          & 32.59                &85.76          &\best{34}         &\textbf{89.47} &\textbf{34}         &\textbf{89.47}       \\
 frostbite          & 11238.4           & 2.46        & 206400          & 45.37          &\textbf{541280.88}    &\textbf{119.01}         &10485             &2.29    &11330            &2.48    \\
 gopher             & 122196            & 34.37       & 113400          & 31.89          & 117777.08            &33.12          &\best{488830}     &\textbf{137.71}  &473560           &133.41    \\
 gravitar           & 6750              & 4.04        & 14200           & 8.62           &\textbf{19213.96}     &\textbf{11.70}          &5905              &3.52    &5915             &3.53    \\
 hero               & 37030.4           & 3.60        & 69400           & 6.84           &\textbf{114736.26}    &\textbf{11.38}          &38330             &3.73    &38225            &3.72    \\
 ice hockey         & \textbf{71.56}    & \textbf{175.34}      &-4.1             & 15.04          & 63.64                &158.56         &37.89             &118.94  &47.11           &123.54    \\
 jamesbond          & 23266             & 51.05       & 26600           & 58.37          & 135784.96            &298.23         &594500              &1305.93 &\textbf{620780}          &\textbf{1363.66}    \\
 kangaroo           & 14112             & 0.99        & \textbf{35100}  & \textbf{2.46}           &24034.16              &1.68           &14500             &1.01    &14636           &1.02    \\
 krull              & 145284.8          & 140.18      & 127400          & 122.73         & 251997.31   &244.29         &97575             &93.63   &\textbf{594540}          &\textbf{578.47}    \\
 kung fu master     & 200176            & 20.00       & 212100          & 21.19          & 206845.82            &20.66          &140440            &14.02            &\textbf{1666665}	          &\textbf{166.68}\\
 montezuma revenge  & 2504              & 0.21        & \textbf{10400}  & \textbf{0.85}           &9352.01               &0.77           &3000              &0.25    &2500            &0.21    \\
 ms pacman          & 29928.2           & 10.22       & 40800           & 13.97          & \textbf{63994.44}    &\textbf{21.98}          &11536             &3.87    &11573           &3.89    \\
 name this game     & 45214.8           & 187.21      & 23900           & 94.24          &\textbf{54386.77}     &\textbf{227.21}         &34434             &140.19  &36296           &148.31    \\
 phoenix            & 811621.6          & 20.20       & \textbf{959100} & \textbf{23.88}          &908264.15             &22.61          &894460            &22.27   &959580          &23.89    \\
 pitfall            & 0                 & 0.20        & 7800            & 7.03           &\textbf{18756.01}     &\textbf{16.62}          &0                 &0.20    &-4.3            &0.20    \\
 pong               & \textbf{21}       & \textbf{100.00}      & 19.6            & 96.64          & 20.67                &99.21          &\best{21}         &\textbf{100.00} &\textbf{21}      &\textbf{100.00}      \\
 private eye        & 300               & 0.27        & \textbf{100000} & \textbf{98.23}          & 79716.46             &78.30          &15100             &14.81   &15100           &14.81    \\
 qbert              & 161000            & 6.70        & 451900          & 18.82          &\textbf{580328.14}    &\textbf{24.18}          &27800             &1.15    &28657           &1.19   \\
 riverraid          & 34076.4           & 3.28        & 36700           & 3.54           & \textbf{63318.67}    &\textbf{6.21}           &28075             &2.68    &28349           &2.70    \\
 road runner        & 498660            & 24.47       & 128600          & 6.31           & 243025.8             &11.92          &878600     &43.11   &\textbf{999999}          &\textbf{49.06}    \\
 robotank           & \textbf{132.4}    & \textbf{176.42}      & 9.1             & 9.35           &127.32                &169.54         &108               &143.63  &113.4           &150.68    \\
 seaquest           & 999991.84         & 100.00      & \textbf{1000000}& \textbf{100.00}         &999997.63             &100.00         &943910	             &94.39 &\textbf{1000000}  &\textbf{100.00}    \\
 skiing             & -29970.32         & -93.10      & -22977.9        & -42.53         & \textbf{-4202.6}     &\textbf{93.27}          &-6774             &74.67           &-6025	          &86.77\\
 solaris            & 4198.4            & 2.69        & 4700            & 3.14           & \textbf{44199.93}    &\textbf{38.99}          &11074             &8.93            &9105            &7.14\\
 space invaders     & 55889             & 8.97        & 43400           & 6.96           & 48680.86             &7.81           &140460     &22.58           &\textbf{154380} &\textbf{24.82}\\
 star gunner        & 521728            & 679.03      & 414600          & 539.43         &\textbf{839573.53}    &\textbf{1093.24}        &465750            &606.09          &677590          &882.15\\
 surround           & \textbf{9.96}     & \textbf{101.84}      & -9.6            & 2.04           & 9.5                  &99.49          &-7.8              &11.22           &2.606           &64.32\\
 tennis             & \textbf{24}       & \textbf{106.70}      & 10.2            & 75.89          & 23.84                &106.34         &\textbf{24}                &\textbf{106.70}          &\textbf{24}              &\textbf{106.70}\\
 time pilot         & 348932            & 559.46      & 344700          & 552.60         &\textbf{405425.31}    &\textbf{650.97}         &216770            &345.37          &450810          &724.49\\
 tutankham          & 393.64            & 7.11        & 191.1           & 3.34           & \textbf{2354.91}     &\textbf{43.62}          &423.9             &7.68            &418.2           &7.57\\
 up n down          & 542918.8          & 658.98      & 620100          & 752.75         & 623805.73            &757.26         &\best{986440}     &\textbf{1197.85}         &966590          &1173.73\\
 venture            & 1992              & 5.12        & 1700            & 4.37           &\textbf{2623.71}      &\textbf{6.74}           &2000              &5.23            &2000            &5.14\\
 video pinball      & 483569.72         & 0.54        & 965300          & 1.08           &\textbf{992340.74}    &\textbf{1.11}           &925830            &1.04            &978190          &1.10\\
 wizard of wor      & 133264            & 33.62       & 106200          & 26.76          &\textbf{157306.41}    &\textbf{39.71}          &64439             &16.14           &63735           &16.00\\
 yars revenge       & 918854.32         & 6.11        & 986000          & 6.55           &\textbf{998532.37}    &\textbf{6.64}           &972000            &6.46            &968090          &6.43\\
 zaxxon             & 181372            & 216.74      & 111100          & 132.75         &\textbf{249808.9}     &\textbf{298.53}         &109140            &130.41          &216020          &258.15\\
\hline
MEAN HWRNS(\%)        &                   & 98.78       &                 &  76.00         &                      &125.92           &                  & \GDIImeanHWRNS        &         & \textbf{\GDIHmeanHWRNS} \\
Learning Efficiency &     & 9.88E-11 &         & 2.17E-11  &        & 1.26E-11 &      & 5.90E-09       &      & \textbf{\GDIHmeanHWRNSle} \\
\hline
MEDIAN HWRNS(\%)      &                   & 33.62       &                 &  21.19         &                      &43.62   &                  & \GDIImedianHWRNS          &         & \textbf{\GDIHmedianHWRNS} \\
Learning Efficiency & & 3.36E-11 &         & 6.05E-12  &        & 4.36E-12 &      & 1.79E-09       &      & \textbf{\GDIHmedianHWRNSle} \\
\hline
HWRB                  &                   & 15          &                 &  8             &                      &18               &                  & \GDIIHWRB             &       & \textbf{\GDIHHWRB} \\
\bottomrule
\end{tabular}
\caption{Score table of SOTA 10B+ model-free algorithms on HWRNS.}
\label{tab: 10B model-free Atari Games Table of Scores Based on Human World Records}
\end{center}
\end{table}
\clearpage
\begin{table}[!hb]
\scriptsize
\begin{center}
\setlength{\tabcolsep}{1.0pt}
\begin{tabular}{ c c c c c c c c c c c }
\toprule
 Games              & MuZero         & HWRNS(\%)      & DreamerV2 & HWRNS(\%)    & SimPLe             & HWRNS(\%)          & GDI-I$^3$     & HWRNS(\%) & GDI-H$^3$ & HWRNS(\%) \\
\midrule
Scale               & 20B            &              & 200M      &            & 1M               &                  & 200M     & &    200M   &\\
\midrule
 alien              & \textbf{741812.63}      & \textbf{294.64  }     &3483       & 1.29     &616.9     & 0.15    & 43384       &  17.15            &48735             &19.27       \\
 amidar             & \textbf{28634.39 }      & \textbf{27.49   }  &2028       & 1.94     &74.3      & 0.07    & 1442        &  1.38                &1065              &1.02     \\
 assault            & \textbf{143972.03}      & \textbf{1706.31 }     &7679       & 88.51       &527.2     & 3.62       & 63876       &  755.57     &97155             &1150.59    \\
 asterix            & 998425         & 99.84        &25669      & 2.55     &1128.3    & 0.09       & 759910      &  75.99         &\textbf{999999}            &\textbf{100.00} \\
 asteroids          & 678558.64               & 6.45        &3064                & 0.02    &793.6              & 0.00    &751970         & 7.15   &\textbf{760005}            &\textbf{7.23}  \\
 atlantis           & 1674767.2               & 15.69           &989207              & 9.22       &20992.5            & 0.08       &3803000        & 35.78  &\textbf{3837300}           &\textbf{36.11}  \\
 bank heist         & 1278.98                 & 1.54        &1043                & 1.25    &34.2               & 0.02    &\best{1401}           & \best{1.69  }  &1380              &1.66 \\
 battle zone        & \textbf{848623         }& \textbf{105.95}      &31225      & 3.87     &4031.2    & 0.47       & 478830      &  59.77      &824360            &102.92     \\
 beam rider         & \textbf{454993.53}      & \textbf{45.48}      &12413      & 1.21     &621.6     & 0.03    & 162100      &  16.18          &422390            &42.22 \\
 berzerk            & \textbf{85932.6        }& \textbf{8.11}      &751        & 0.06     &N/A       & N/A     & 7607        &  0.71            &14649             &1.37 \\
 bowling            & \textbf{260.13         }& \textbf{85.60}      &48         & 8.99     &30        & 2.49    & 202         &  64.57          &205.2             &65.76   \\
 boxing             & \textbf{100}                     & \textbf{100.00}       &87                  & 86.99       &7.8         & 7.71       & \best{100}  & \best{100.00 } &\textbf{100}               &\textbf{100.00} \\
 breakout           & \textbf{864}                     & \textbf{100.00}          &350                 & 40.39       &16.4     & 1.70       & \best{864}  & \best{100.00}  &\textbf{864}             &100.00 \\
 centipede          & \textbf{1159049.27}     & \textbf{89.02}     &6601       & 0.35     &N/A       & N/A     & 155830      & 11.83                     &195630            &14.89\\
 chopper command    & 991039.7                & 99.10     &2833                & 0.20     & 979.4             & 0.02    & \best{999999}        & \best{100.00} &\textbf{999999}            &\textbf{100.00}  \\
 crazy climber      & \textbf{458315.4}       & \textbf{214.01    }  &141424     & 62.47       & 62583.6  & 24.77   & 201000      & 90.96                      &241170            &110.17 \\
 defender           & 839642.95               & 13.93      & N/A                & N/A        & N/A               & N/A      & 893110 & 14.82 &\textbf{970540}            &\textbf{16.11}   \\
 demon attack       & 143964.26               & 9.24      & 2775              &0.17      & 208.1             & 0.00    & 675530      & 43.40  &\textbf{787985}   &\textbf{50.63} \\
 double dunk        & 23.94          & 107.42  & 22        &102.53        & N/A      & N/A     & \textbf{24}          & \textbf{107.58}                    &\textbf{24}                &\textbf{107.58}    \\
 enduro             & 2382.44                 & 25.08       & 2112              &22.23         & N/A               & N/A     & \best{14330} & \best{150.84}&14300             &150.53   \\
 fishing derby      & \textbf{91.16}          & \textbf{112.39     }              &93.24          &286.77      &-90.7     & 0.61    & 59          & 92.89  &65               &96.31        \\
 freeway            & 33.03                   & 86.92       & \textbf{34}                 &\textbf{89.47}          &16.7               & 43.95   & \best{34}      & \best{89.47} &\textbf{34}               &\textbf{89.47}   \\
 frostbite          & \textbf{631378.53}      & \textbf{138.82}     & 15622    &3.42       &236.9     & 0.04    & 10485       & 2.29                           &11330            &2.48\\
 gopher             & 130345.58               & 36.67      & 53853             &15.11         &596.8              & 0.10       & \best{488830} & \best{137.71} &473560           &133.41  \\
 gravitar           & \textbf{6682.7     }    & \textbf{4.00    }  & 3554     &2.08      &173.4     & 0.00    & 5905        & 3.52     &5915             &3.53       \\
 hero               & \textbf{49244.11}       & \textbf{4.83    }  & 30287    &2.93      &2656.6    & 0.16        & 38330       & 3.73 &38225            &3.72           \\
 ice hockey         & \textbf{67.04      }    & \textbf{165.76  }  & 29        &85.17         &-11.6     & -0.85       & 38          & 118.94 &47.11           &123.54         \\
 jamesbond          & 41063.25                & 90.14     & 9269              &20.30         &100.5     & 0.16    &594500              &1305.93 &\textbf{620780}          &\textbf{ 1363.66}  \\
 kangaroo           & \textbf{16763.6        }& \textbf{1.17}  & 11819     &0.83      &51.2      & 0.00    & 14500       & 1.01          &14636           &1.02   \\
 krull              & 269358.27               & 261.22     & 9687     &7.89       &2204.8    & 0.59    & 97575       & 93.63    &\textbf{594540}          &\textbf{578.47}       \\
 kung fu master     & 204824         & 20.46  & 66410    &6.62       &14862.5   & 1.46    & 140440      & 14.02        &\textbf{1666665}          &\textbf{166.68 } \\
 montezuma revenge  & 0                       & 0.00         & 1932              &0.16       &N/A       & N/A     & \best{3000}          & \best{0.25  } &2500            &0.21  \\
 ms pacman          & \textbf{243401.1 }      & \textbf{83.89   }  & 5651     &1.84       &1480      & 0.40    & 11536       & 3.87       &11573           &3.89     \\
 name this game     & \textbf{157177.85}      & \textbf{675.54  }  & 14472    &53.12          &2420.7    & 0.56    & 34434       & 140.19 &36296           &148.31        \\
 phoenix            & \textbf{955137.84}      & \textbf{23.78   }  & 13342     &0.31       &N/A       & N/A     & 894460      & 22.27     &959580          &23.89     \\
 pitfall            & \textbf{0}                       & \textbf{0.20}               & -1                 &0.20       &N/A       & N/A     & \best{0} & \best{0.20} &-4.3            &0.20     \\
 pong               & \textbf{21}                      & 100.00             & 19                 &95.20          & 12.8     & 80.34   & \best{21} & \best{100.00} &\textbf{21}              &\textbf{100.00}   \\
 private eye        & \textbf{15299.98 }      & \textbf{15.01}     & 158       &0.13       & 35       & 0.01    & 15100       & 14.81             &15100           &14.81     \\
 qbert              & \textbf{72276          }& \textbf{3.00}      & 162023    &6.74       & 1288.8   & 0.05    & 27800       & 1.15              &28657           &1.19 \\
 riverraid          & \textbf{323417.18}      & \textbf{32.25}     & 16249    &1.49       & 1957.8   & 0.06    & 28075       & 2.68               &28349           &2.70\\
 road runner        & 613411.8                & 30.10              & 88772             &4.36       & 5640.6   & 0.28       & 878600 & 43.11   &\textbf{999999}          &\textbf{49.06}   \\
 robotank           & \textbf{131.13}         & \textbf{174.70}    & 65        &85.09          & N/A      & N/A     & 108         & 143.63        &113.4           &150.68 \\
 seaquest           & 999976.52               & 100.00             & 45898             &4.58       & 683.3             & 0.06    &943910	             &94.39 &\textbf{1000000}          &\textbf{100.00}\\
 skiing             & -29968.36      & -93.09      & -8187    &64.45          & N/A      & N/A     & -6774       & 74.67        &\textbf{-6025}	         &\textbf{86.77} \\
 solaris            & 56.62                   & -1.07              & 883                &-0.32          & N/A               & N/A     & \best{11074}         & \best{8.93 } &9105            &7.14\\
 space invaders     & 74335.3                 & 11.94                 & 2611               &0.40       & N/A               & N/A     & 140460        & 22.58  &\textbf{154380}          &\textbf{24.82}   \\
 star gunner        & 549271.7       & 714.93     & 29219    &37.21          & N/A      & N/A     & 465750      & 606.09      &\textbf{677590}          &\textbf{882.15}\\
 surround           & \textbf{9.99       }    & \textbf{101.99}     & N/A       &N/A         & N/A      & N/A     & -8          & 11.22         &2.606           &64.32 \\
 tennis             & 0                       & 53.13      & 23        &104.46        & N/A      & N/A     & \textbf{24}          & \textbf{106.70} &\textbf{24}              &\textbf{106.70}      \\
 time pilot         & \textbf{476763.9}       & \textbf{766.53}         & 32404    &46.71         & N/A      & N/A     & 216770      & 345.37       &450810          &724.49   \\
 tutankham          & \textbf{491.48     }    & \textbf{8.94}   & 238       &4.22         & N/A      & N/A     & 424         & 7.68                 &418.2           &7.57 \\
 up n down          & 715545.61               & 868.72            & 648363            &787.09        & 3350.3            & 3.42   & \best{986440}   & \best{1197.85} &966590          &1173.73    \\
 venture            & 0.4                     & 0.00        & 0                  &0.00       & N/A               & N/A     & \best{2030}          & \best{5.23}      &2000            &5.14  \\
 video pinball      & \textbf{981791.88}      & \textbf{1.10}      & 22218    &0.02     & N/A      & N/A     & 925830      & 1.04                                    &978190          &1.10\\
 wizard of wor      & \textbf{197126         }& \textbf{49.80}      & 14531    &3.54     & N/A      & N/A     & 64439       & 16.14                                  &63735           &16.00\\
 yars revenge       & 553311.46               & 3.67      & 20089             &0.11     & 5664.3            & 0.02    & \best{972000}        & \best{6.46}           &968090          &6.43\\
 zaxxon             & \textbf{725853.9}       & \textbf{867.51}      & 18295    &21.83          & N/A      & N/A     & 109140      & 130.41                          &216020          &258.15\\
\hline
MEAN HWRNS(\%) &               &152.1  &         &  37.9 &                                          & 4.67  &     & \GDIImeanHWRNS &                  & \textbf{\GDIHmeanHWRNS} \\
Learning Efficiency &      &7.61E-11 &         & 1.89E-09 &        & \textbf{4.67E-08} &      & 5.90E-09       &      & \GDIHmeanHWRNSle \\
\hline
MEDIAN HWRNS(\%) &            & 49.8  &         & 4.22  &                                          & 0.13  &     & \textbf{\GDIImedianHWRNS} &                  & \GDIHmedianHWRNS \\
Learning Efficiency &  & 2.49E-11 &         & 2.11E-10  &        & 1.25E-09 &      & 1.79E-09       &      & \textbf{\GDIHmedianHWRNSle} \\
\hline
HWRB &            & 19   &         & 3    &                                          & 0  &     & \GDIIHWRB & & \textbf{\GDIHHWRB} \\
\bottomrule

\end{tabular}
\caption{Score table of SOTA model-based algorithms on HWRNS.}
\label{tab: model-based Atari Games Table of Scores Based on Human World Records}
\end{center}
\end{table}
\clearpage
\begin{table}[!hb] 
\scriptsize
\begin{center}
\setlength{\tabcolsep}{1.0pt}
\begin{tabular}{ c c c c c c c c c }
\toprule
 Games      & Muesli              & HWRNS(\%)        & Go-Explore              & HWRNS(\%)                     & GDI-I$^3$ & HWRNS(\%) & GDI-H$^3$ & HWRNS(\%) \\
\midrule
Scale       &                     & 200M        & 10B                     &                             & 200M      &    &    200M   &\\
\midrule    
 alien        &139409          &55.30               &\textbf{959312}       &\textbf{381.06}                & 43384             &17.15     &48735             &19.27            \\
 amidar       &\textbf{21653}  &\textbf{20.78}      &19083        &18.32                & 1442              &1.38                         &1065              &1.02          \\
 assault      &36963           &436.11           &30773                 &362.64                         & 63876      &755.57&\textbf{97155}    &\textbf{1150.59}           \\
 asterix      &316210          &31.61            &999500       &99.95                 & 759910            &75.99        &\textbf{999999}            &\textbf{100.00}    \\
 asteroids    &484609          &4.61             &112952                &1.07                           & 751970     &7.15  &\textbf{760005}            &\textbf{7.23}        \\
 atlantis     &1363427         &12.75            &286460                &2.58                           & 3803000    &35.78 &\textbf{3837300}           &\textbf{36.11}         \\
 bank heist   &1213            &1.46          &\textbf{3668}         &\textbf{4.45  }                & 1401              &1.69            &1380              &1.66    \\
 battle zone  &414107          &51.68            &\textbf{998800}       &\textbf{124.70}                & 478830            &59.77        &824360            &102.92     \\
 beam rider   &288870          &28.86         &371723       &37.15                & 162100            &16.18           &\textbf{422390}            &\textbf{42.22}     \\
 berzerk      &44478           &4.19             &\textbf{131417}      &\textbf{12.41 }                & 7607              &0.71          &14649             &1.37       \\
 bowling      &191             &60.64        &\textbf{247}           &\textbf{80.86 }                & 202               &64.57            &205.2             &65.76 \\
 boxing       &99              &99.00        &91                     &90.99                          & \best{100}        &\best{100.00}    &\textbf{100}      &\textbf{100.00}        \\
 breakout     &791             &91.53          &774                    &89.56                          & \best{864}        &\best{100.00}  &\textbf{864}              &\textbf{100.00}       \\
 centipede    &\textbf{869751} &\textbf{66.76}          &613815      &47.07                 & 155830            &11.83                    &195630            &14.89  \\
 chopper command &101289       &10.06       &996220                &99.62                          & \best{999999}     &\best{100.00}     &\textbf{999999}            &\textbf{100.00}    \\
 crazy climber   &175322       &78.68     &235600       &107.51                & 201000            &90.96               &\textbf{241170}            &\textbf{110.17}\\
 defender        &629482       &10.43        &N/A                    &N/A                            & 893110     &14.82    &\textbf{970540}                &\textbf{16.11}      \\
 demon attack    &129544       &8.31        &239895                 &15.41                          & 675530     &43.40     &\textbf{787985}   &\textbf{50.63}    \\
 double dunk     &-3           &39.39       &\textbf{24}                     &\textbf{107.58}       & \best{24}       &\best{107.58}      &\textbf{24}      &\textbf{107.58}      \\
 enduro          &2362         &24.86                   &1031                   &10.85              & \best{14330}      &\best{150.84}    &14300             &150.53        \\
 fishing derby   &51           &87.71       &\textbf{67}            &\textbf{97.54 }                & 59                &92.89            &65               &96.31\\
 freeway         &33           &86.84                   &\textbf{34}            &\textbf{89.47 }    & \best{34}         &\best{89.47}     &\textbf{34}      &\textbf{89.47}      \\
 frostbite       &301694       &66.33            &\textbf{999990}       &\textbf{219.88}                & 10485             &2.29         &11330            &2.48   \\
 gopher          &104441       &29.37            &134244                &37.77                          & \best{488830}     &\best{137.71} &473560           &133.41       \\
 gravitar        &11660        &7.06           &\textbf{13385}        &\textbf{8.12}                  & 5905              &3.52           &5915             &3.53  \\
 hero            &37161        &3.62                     &37783                  &3.68                           & \best{38330}      &\best{3.73}   &38225           &3.72           \\
 ice hockey      &25           &76.69          &33                     &93.64                          & 45         &118.94           &\textbf{47.11}  &\textbf{123.54} \\
 jamesbond       &19319        &42.38            &200810                &441.07                         &594500              &1305.93         &\textbf{620780}       &\textbf{ 1363.66}\\
 kangaroo        &14096        &0.99                   &\textbf{24300}        &\textbf{1.70}                  & 14500             &1.01             &14636           &1.02\\
 krull           &34221        &31.83                     &63149                 &60.05                          & 97575      &93.63  &\textbf{594540}          &\textbf{578.47}        \\
 kung fu master  &134689       &13.45      &24320                 &2.41                           & 140440     &14.02                 &\textbf{1666665}          &\textbf{166.68}\\
 montezuma revenge  &2359      &0.19             &\textbf{24758}        &\textbf{2.03}                  & 3000              &0.25                   &2500            &0.21\\
 ms pacman          &65278     &22.42     &\textbf{456123}       &\textbf{157.30}                & 11536             &3.87                          &11573           &3.89\\
 name this game     &105043    &448.15     &\textbf{212824}       &\textbf{918.24}               & 34434             &140.19                        &36296           &148.31\\
 phoenix        &805305        &20.05                                &19200                 &0.46                           & 894460     &22.27 &\textbf{959580}          &\textbf{23.89}   \\
 pitfall        &0             &0.20                   &\textbf{7875}          &\textbf{7.09   }               & 0                 &0.2             &-4.3            &0.20 \\
 pong           &20            &97.60                       &\textbf{21}            &\textbf{100.00 }               & \best{21}         &\best{100} &\textbf{21}              &\textbf{100.00}          \\
 private eye    &10323         &10.12                &\textbf{69976}        &\textbf{68.73  }               & 15100             &14.81               &15100           &14.81 \\
 qbert          &157353        &6.55                            &\textbf{999975}       &\textbf{41.66  }               & 27800             &1.15     &28657           &1.19       \\
 riverraid      &\textbf{47323}&\textbf{4.60}               &35588        &3.43               & 28075             &2.68                              &28349           &2.70\\
 road runner    &327025        &16.05                        &999900        &49.06                 & 878600           &43.11       &\textbf{999999}          &\textbf{49.06}    \\
 robotank       &59            &76.96                             &\textbf{143}           &\textbf{190.79 }               & 108         &143.63      &113.4           &150.68          \\
 seaquest       &815970        &81.60                &539456       &53.94                 &943910	             &94.39       &\textbf{1000000}          &\textbf{100.00}    \\
 skiing         &-18407        &-9.47                      &\textbf{-4185}        &\textbf{93.40  }               & -6774             &74.67         &-6025	          &86.77      \\
 solaris        &3031          &1.63                         &\textbf{20306}        &\textbf{17.31  }               & 11074             &8.93        &9105            &7.14        \\
 space invaders &59602         &9.57                 &93147                 &14.97                          & 140460     &22.58        &\textbf{154380}          &\textbf{24.82} \\
 star gunner    &214383        &278.51   &609580                 &793.52                         &465750     &606.09                  &\textbf{677590}          &\textbf{882.15}\\
 surround       &\textbf{9}    &\textbf{96.94}             &N/A                    &N/A                            & -8         &11.22               &2.606           &64.32\\
 tennis         &12            &79.91                           &\best{24}              &\best{106.7}                   & \best{24}         &\best{106.70   }  &\textbf{24}              &\textbf{106.70}          \\
 time pilot     &359105 &575.94    &183620                &291.67                         & 216770     & 345.37                    &\textbf{450810}          &\textbf{724.49}\\
 tutankham      &252           &4.48         &\textbf{528}           &\textbf{9.62}                  & 424               &7.68                       &418.2           &7.57\\
 up n down      &649190        &788.10                &553718                &672.10                         & \best{986440}     &\best{1197.85}     &966590          &1173.73      \\
 venture        &2104          &5.41            &\textbf{3074}         &\textbf{7.90}                & 2035              &5.23                       &2000            &5.14\\
 video pinball  &685436        &0.77                 &\textbf{999999}       &\textbf{1.12}               & 925830            &1.04                   &978190          &1.10\\
 wizard of wor  &93291         &23.49                &\textbf{199900}       &\textbf{50.50}               & 64293             &16.14                 &63735           &16.00\\
 yars revenge   &557818        &3.70               &\textbf{999998}       &\textbf{6.65}               & 972000            &6.46                     &968090          &6.43\\
 zaxxon         &65325         &78.04                 &18340                 &21.88                        & 109140     &130.41        &\textbf{216020} &\textbf{258.15}  \\
\hline    
MEAN HWRNS(\%)  &    & 75.52 &                       & 116.89                       &            &  \GDIImeanHWRNS  &                  & \textbf{\GDIHmeanHWRNS} \\
Learning Efficiency &    & 3.78E-09 &                       & 1.17E-10                      &      & 5.90E-09       &      & \textbf{\GDIHmeanHWRNSle}\\
\hline
MEDIAN HWRNS(\%)&   & 24.68  &                       & 50.5              &            & \GDIImedianHWRNS   &                  & \textbf{\GDIHmedianHWRNS} \\
Learning Efficiency &    & 1.24E-09 &                       & 5.05E-11                       &      & 1.79E-09       &      & \textbf{\GDIHmedianHWRNSle} \\
\hline
HWRB    &       & 5 &               & 15              &            & \GDIIHWRB  & & \textbf{\GDIHHWRB} \\
\bottomrule
\end{tabular}
\caption{Score table of other SOTA  algorithms on HWRNS.}
\label{tab: other Atari Games Table of Scores Based on Human World Records}
\end{center}
\end{table}
\clearpage

\section{Atari Games Table of Scores Based on SABER}
\label{app: Atari Games Table of Scores Based on SABER}
In this part, we detail the raw score of several representative SOTA algorithms including the SOTA 200M model-free algorithms, SOTA 10B+ model-free algorithms, SOTA model-based algorithms, and other SOTA algorithms.\footnote{200M and 10B+ represent the training scale.} Additionally, we calculate the capped human world records normalized world score (CHWRNS) or called SABER \citep{atarihuman} of each game with each algorithm. First of all, we demonstrate the sources of the scores that we used.
Random scores  are from \citep{agent57}.
Human world records (HWR) are form \citep{dreamerv2,atarihuman}.
Rainbow's scores are from \citep{rainbow}.
IMPALA's scores are from \citep{impala}.
LASER's scores are from \citep{laser}, no sweep at 200M.
As there are many versions of R2D2 and NGU, we use original papers'.
R2D2's scores are from \citep{r2d2}.
NGU's scores are from \citep{ngu}.
Agent57's scores are from \citep{agent57}.
MuZero's scores are from \citep{muzero}.
DreamerV2's scores are from \citep{dreamerv2}.
SimPLe's scores are form \citep{modelbasedatari}.
Go-Explore's scores are form \citep{goexplore}.
Muesli's scores are form \citep{muesli}.
In the following we detail the raw scores and SABER of each algorithm on 57 Atari games.
\clearpage

\begin{table}[!hb]
\scriptsize
\begin{center}
\setlength{\tabcolsep}{1.0pt}
\begin{tabular}{ c c c c c c c  c c c c c c }
\toprule
Games & RND & HWR & RAINBOW & SABER(\%) & IMPALA & SABER(\%) & LASER & SABER(\%) & GDI-I$^3$ & SABER(\%) & GDI-H$^3$ & SABER(\%)\\
\midrule
Scale  &     &       & 200M   &       &  200M    &        & 200M   & &  200M   & &  200M   & \\
\midrule
 alien              & 227.8     & \textbf{251916}    & 9491.7   &3.68    & 15962.1    & 6.25       & 976.51  & 14.04     &43384      &17.15   &48735             &19.27  \\
 amidar             & 5.8       & \textbf{104159}    & 5131.2   &4.92    & 1554.79    & 1.49       & 1829.2  & 1.75      &1442              &1.38  &1065              &1.02           \\
 assault            & 222.4     & 8647               & 14198.5  &165.90  & 19148.47   & 200.00     & 21560.4 & 200.00    &63876      &200.00  &\textbf{97155}     &\textbf{200.00} \\
 asterix            & 210       & \textbf{1000000}   & 428200   &42.81   & 300732     & 30.06      & 240090  & 23.99     &759910     &75.99   &999999            &100.00\\
 asteroids          & 719       & \textbf{10506650}  & 2712.8   &0.02    & 108590.05  & 1.03       & 213025  & 2.02      &751970     &7.15    &760005            &7.23\\
 atlantis           & 12850     & \textbf{10604840}  & 826660   &7.68    & 849967.5   & 7.90       & 841200  & 7.82      &3803000    &35.78   &3837300           &36.11\\
 bank heist         & 14.2      & \textbf{82058}     & 1358     &1.64    & 1223.15    & 1.47       & 569.4   & 0.68      &1401       &1.69    &1380              &1.66 \\
 battle zone        & 236       &801000    & 62010    &7.71    & 20885      & 2.58       & 64953.3 & 8.08      &478830     &59.77   &\textbf{824360}            &\textbf{102.92} \\
 beam rider         & 363.9     & \textbf{999999}    & 16850.2  &1.65    & 32463.47   & 3.21       & 90881.6 & 9.06      &162100     &16.18   &422390            &42.22 \\
 berzerk            & 123.7     & \textbf{1057940}            & 2545.6   &0.23    & 1852.7     & 0.16       & 25579.5 & 2.41      &7607              &0.71 &14649             &1.37             \\
 bowling            & 23.1      & \textbf{300}       & 30       &2.49    & 59.92      & 13.30      & 48.3    & 9.10      &201.9      &64.57   &205.2             &65.76\\
 boxing             & 0.1       & \textbf{100}                & 99.6     &99.60   & 99.96      & 99.96      & \textbf{100}     & \textbf{100.00}    &\best{100}        &\best{100.00 } &\textbf{100}       &\textbf{100.00}    \\
 breakout           & 1.7       & \textbf{864}                & 417.5    &48.22   & 787.34     & 91.11      & 747.9   & 86.54     &\best{864}        &\best{100.00 } &\textbf{864}     &\textbf{100.00}   \\
 centipede          & 2090.9    & \textbf{1301709}   & 8167.3   &0.47    & 11049.75   & 0.69       & 292792  & 22.37     &155830            &11.83          &195630    &14.89\\
 chopper command    & 811       & \textbf{999999}             & 16654    &1.59    & 28255      & 2.75       & 761699  & 76.15     &\best{999999}     &\best{100.00 } &\textbf{999999}    &\textbf{100.00}\\
 crazy climber      & 10780.5   & 219900    & 168788.5 &75.56   & 136950     & 60.33      & 167820  & 75.10     &201000     &90.96                 &\textbf{241170}    &\textbf{110.17}\\
 defender           & 2874.5    & \textbf{6010500}   & 55105    &0.87    & 185203     & 3.03       & 336953  & 5.56      &893110     &14.82                 &970540    &16.11\\
 demon attack       & 152.1     & \textbf{1556345}   & 111185   &7.13    & 132826.98  & 8.53       & 133530  & 8.57      &675530     &43.10                  &787985   &50.63\\
 double dunk        & -18.6     & 21                 & -0.3     &46.21   & -0.33      & 46.14      & 14      & 82.32     &\best{24}      &\best{107.58} &\textbf{24}        &\textbf{107.58}\\
 enduro             & 0         & 9500               & 2125.9   &22.38   & 0          & 0.00       & 0       & 0.00      &\best{14330}      &\best{150.84  }&14300     &150.53\\
 fishing derby      & -91.7     & \textbf{71}        & 31.3     &75.60   & 44.85      & 83.93      & 45.2    & 84.14     &59                &95.08          &65        &96.31\\
 freeway            & 0         & \textbf{38}        & 34       &89.47   & 0          & 0.00       & 0       & 0.00      &34                &89.47          &34        &89.47\\
 frostbite          & 65.2      & \textbf{454830}    & 9590.5   &2.09    & 317.75     & 0.06       & 5083.5  & 1.10      &10485      &2.29                  &11330     &2.48\\          
 gopher             & 257.6     & 355040             & 70354.6  &19.76   & 66782.3    & 18.75      & 114820.7& 32.29     &\best{488830}     &\best{137.71 } &473560    &133.41\\
 gravitar           & 173       & \textbf{162850}    & 1419.3   &0.77    & 359.5      & 0.11       & 1106.2  & 0.57      &5905       &3.52                  &5915      &3.53\\
 hero               & 1027      & \textbf{1000000}            & 55887.4  &5.49    & 33730.55   & 3.27       & 31628.7 & 3.06      &38330             &3.73           &38225     &3.72\\
 ice hockey         & -11.2     & 36                 & 1.1      &26.06   & 3.48       & 31.10      & 17.4    & 60.59     &44.92              &118.94       &\textbf{47.11}           &\textbf{123.54}\\
 jamesbond          & 29        & 45550              & 19809    &43.45   & 601.5      & 1.26       & 37999.8 & 83.41     &594500              &200.00  &\textbf{620780}          &\textbf{200.00}\\
 kangaroo           & 52        & \textbf{1424600}   & 14637.5  &1.02    & 1632       & 0.11       & 14308   & 1.00      &14500             &1.01           &14636           &1.02\\
 krull              & 1598      & 104100    & 8741.5   &6.97    & 8147.4     & 6.39       & 9387.5  & 7.60      &97575     &93.63                  &\textbf{594540}          &\textbf{200.00}\\
 kung fu master     & 258.5     & 1000000   & 52181    &5.19    & 43375.5    & 4.31       & 607443  & 60.73     &140440            &14.02          &\textbf{1666665}          &\textbf{166.68}\\
 montezuma revenge  &0          & \textbf{1219200}   & 384      &0.03    & 0          & 0.00       & 0.3     & 0.00      &3000              &0.25           &2500            &0.21\\
 ms pacman          & 307.3     & \textbf{290090}    & 5380.4   &1.75    & 7342.32    & 2.43       & 6565.5  & 2.16      &11536              &3.87          &11573           &3.89\\
 name this game     & 2292.3    & 25220              & 13136    &47.30   & 21537.2    & 83.94      & 26219.5 & 104.36    &34434      &140.19  &\textbf{36296}  &\textbf{148.31}\\
 phoenix            & 761.5     & \textbf{4014440}   & 108529   &2.69    & 210996.45  & 5.24       & 519304  & 12.92     &894460         &22.27             &959580          &23.89\\
 pitfall            & -229.4    & \textbf{114000}    & \textbf{0}        &\textbf{0.20}    & -1.66      & 0.20       & -0.6    & 0.20      &\best{0    }      &0.20           &-4.3            &0.20\\
 pong               & -20.7     & \textbf{21}                 & 20.9     &99.76   & 20.98      & 99.95      & \textbf{21}      & \textbf{100.00}    &\best{21   }      &\best{100.00 } &\textbf{21}     &\textbf{100.00}    \\
 private eye        & 24.9      & \textbf{101800}    & 4234     &4.14    & 98.5       & 0.07       & 96.3    & 0.07      &15100      &14.81                 &15100           &14.81\\
 qbert              & 163.9     & \textbf{2400000}   & 33817.5  &1.40    & 351200.12  & 14.63      & 21449.6 & 0.89      &27800             &1.03           &28657           &1.19  \\
 riverraid          & 1338.5    & \textbf{1000000}   & 22920.8  &2.16    & 29608.05   & 2.83       & 40362.7 & 3.91      &28075             &2.68           &28349           &2.70  \\
 road runner        & 11.5      & \textbf{2038100}   & 62041    &3.04    & 57121      & 2.80       & 45289   & 2.22      &878600    &43.11                  &999999          &49.06\\
 robotank           & 2.2       & 76                 & 61.4     &80.22   & 12.96      & 14.58      & 62.1    & 81.17     &108.2      &143.63  &\textbf{113.4}  &\textbf{150.68}  \\
 seaquest           & 68.4      & 999999             & 15898.9  &1.58    & 1753.2     & 0.17       & 2890.3  & 0.28      &943910	             &94.39&\textbf{1000000}          &\textbf{100.00}  \\
 skiing             & -17098    & \textbf{-3272}     & -12957.8 &29.95   & -10180.38  & 50.03      & -29968.4& -93.09    &-6774             &74.67          &-6025	         &86.77   \\
 solaris            & 1236.3    & \textbf{111420}    & 3560.3   &2.11    & 2365       & 1.02       & 2273.5  & 0.94      &11074             &8.93           &9105            &7.14  \\
 space invaders     & 148       & \textbf{621535 }   & 18789    &3.00    & 43595.78   & 6.99       & 51037.4 & 8.19      &140460    &22.58                  &154380          &24.82\\
 star gunner        & 664       & 77400              & 127029   &164.67  & 200625     & 200.00     & 321528  & 418.14    &465750     &200.00  &\textbf{677590} &\textbf{200.00}  \\
 surround           & -10       & 9.6                & \textbf{9.7}      &\textbf{100.51}  & 7.56       & 89.59      & 8.4     & 93.88     &-7.8    &11.22  &2.606           &64.32\\
 tennis             & -23.8     & 21                 & 0        &53.13   & 0.55       & 54.35      & 12.2    & 80.36     &\best{24       }  &\best{106.70    }  &\textbf{24} &\textbf{106.70}  \\
 time pilot         & 3568      & 65300              & 12926    &15.16   & 48481.5    & 72.76      & 105316  & 164.82    &216770     &200.00   &\textbf{450810}          &\textbf{200.00}\\
 tutankham          & 11.4      & \textbf{5384}      & 241      &4.27    & 292.11     & 5.22       & 278.9   & 4.98      &423.9     &7.68                    &418.2           &7.57\\
 up n down          & 533.4     & 82840              & 125755   &152.14  & 332546.75  & 200.00     & 345727  & 200.00    &\best{986440}     &\best{200.00  } &966590        &200.00\\
 venture            & 0         & \textbf{38900}     & 5.5      &0.01    & 0          & 0.00       & 0       & 0.00      &2000    &5.14                      &2000            &5.14\\
 video pinball      & 0         & \textbf{89218328}  & 533936.5 &0.60    & 572898.27  & 0.64       & 511835  & 0.57      &925830  &1.04                      &978190          &1.10\\\
 wizard of wor      & 563.5     & \textbf{395300}    & 17862.5  &4.38    & 9157.5     & 2.18       & 29059.3 & 7.22      &64439   &16.18                     &63735           &16.00\\
 yars revenge       & 3092.9    & \textbf{15000105}  & 102557   &0.66    & 84231.14   & 0.54       & 166292.3& 1.09      &972000  &6.46                      &968090          &6.43\\
 zaxxon             & 32.5      & 83700              & 22209.5  &26.51   & 32935.5    & 39.33      & 41118   & 49.11     &109140     &130.41   &\textbf{216020}          &\textbf{200.00}\\
\hline
MEAN SABER(\%) &     0.00 & \textbf{100.00}   &         & 28.39 &         & 29.45  &        & 36.78 &      & \GDIImeanSABER &      &\GDIHmeanSABER\\
Learning Efficiency &     0.00 & N/A   &         & 1.42E-09 &         & 1.47E-09  &        & 1.84E-09 &      & 3.08E-09       &      & \textbf{\GDIHmeanSABERle} \\
\hline
MEDIAN SABER(\%) & 0.00   & \textbf{100.00}   &         & 4.92 &         & 4.31  &        & 8.08  &      & \GDIImedianSABER &      & \GDIHmedianSABER  \\
Learning Efficiency &     0.00 & N/A   &         &  2.46E-10 &         & 2.16E-10  &        &  4.04E-10  &      &2.27E-09      &      & \textbf{\GDIHmedianSABERle} \\
\hline
HWRB & 0   & \textbf{57}   &         & 4 &         & 3  &        & 7  &      & \GDIIHWRB &      & \GDIHHWRB  \\
\bottomrule
\end{tabular}
\caption{Score table of SOTA 200M model-free algorithms on SABER.}
\end{center}
\end{table}
\clearpage

\begin{table}[!hb]
\scriptsize
\begin{center}
\setlength{\tabcolsep}{1.0pt}
\begin{tabular}{ c c c c c c c c c c c }
\toprule
 Games & R2D2 & SABER(\%) & NGU & SABER(\%) & AGENT57 & SABER(\%) & GDI-I$^3$ & SABER(\%) & GDI-H$^3$ & SABER(\%)\\
\midrule
Scale  & 10B   &        & 35B &         & 100B     &        & 200M & &  200M   & \\
\midrule
 alien              & 109038.4          & 43.23             & 248100          & 98.48          & \textbf{297638.17}   &\textbf{118.17}           &43384             &17.15              &48735  &19.27        \\
 amidar             & 27751.24          & 26.64             & 17800           & 17.08          & \textbf{29660.08}    &\textbf{28.47}            &1442              &1.38               &1065              &1.02     \\
 assault            & 90526.44          & 200.00            & 34800           & 200.00         & 67212.67             &200.00                    &63876             &200.00             &\textbf{97155}     &\textbf{ 200.00}  \\
 asterix            & 999080            & 99.91             & 950700          & 95.07          & 991384.42            &99.14                     &759910            &75.99              &\textbf{999999}     &\textbf{100.00}\\
 asteroids          & 265861.2          & 2.52              & 230500          & 2.19           & 150854.61            &1.43                      &751970            &7.15               &\textbf{760005}     &\textbf{7.23}    \\
 atlantis           & 1576068           & 14.76             & 1653600         & 15.49          & 1528841.76           &14.31                     &3803000           &35.78              &\textbf{3837300}    &\textbf{36.11}            \\
 bank heist         & \textbf{46285.6}  & \textbf{56.40}    & 17400           & 21.19          & 23071.5              &28.10                     &1401              &1.69               &1380       &1.66\\
 battle zone        & 513360            & 64.08             & 691700          & 86.35          & \textbf{934134.88}   &\textbf{116.63}           &478830            &59.77              &824360     &102.92  \\
 beam rider         & 128236.08         & 12.79             & 63600           & 6.33           & 300509.8             &30.03                     &162100            &16.18              &\textbf{422390}     &\textbf{42.22}            \\
 berzerk            & 34134.8           & 3.22              & 36200           & 3.41           & \textbf{61507.83}    &\textbf{5.80}             &7607              &0.71               &14649      &1.37         \\
 bowling            & 196.36            & 62.57             & 211.9           & 68.18          & \textbf{251.18}      &\textbf{82.37}            &201.9             &64.57              &205.2      &65.76   \\
 boxing             & 99.16             & 99.16             & 99.7            & 99.70          & \textbf{100}         &\textbf{100.00}           &\best{100}        &\textbf{100.00}    &\textbf{100} &\textbf{100.00}      \\
 breakout           & 795.36            & 92.04             & 559.2           & 64.65          & 790.4                &91.46                     &\best{864}        &\textbf{100.00}    &\textbf{864}     &\textbf{100}          \\
 centipede          & 532921.84         & 40.85             & \textbf{577800} & 44.30          & 412847.86            &31.61                     &155830            &11.83              &195630    &14.89\\
 chopper command    &960648             & 96.06             &999900           & 99.99          &999900                &99.99                     &\best{999999}     &\textbf{100.00}    &\textbf{999999}  &\textbf{100.00} \\
 crazy climber      & 312768            & 144.41            & 313400          & 144.71         &\textbf{565909.85}    &\textbf{200.00}           &201000            &90.96              &241170    &110.17      \\
 defender           & 562106            & 9.31              & 664100          & 11.01          & 677642.78            &11.23                     &893110            &14.82              &\textbf{970540}&\textbf{16.11}          \\
 demon attack       & 143664.6          & 9.22              & 143500          & 9.21           & 143161.44            &9.19                      &675530            &43.10              &\textbf{787985}   &\textbf{50.63}           \\
 double dunk        & 23.12             & 105.35            & -14.1           & 11.36          & 23.93                &107.40                    &\textbf{24}       &\textbf{107.58}    &\textbf{24}&\textbf{107.58}     \\
 enduro             & 2376.68           & 25.02             & 2000            & 21.05          & 2367.71              &24.92                     &\best{14330}      &\textbf{150.84}    &14300     &150.53          \\
 fishing derby      & 81.96             & 106.74            & 32              & 76.03          & \textbf{86.97}       &\textbf{109.82}           &59                &95.08   &65        &96.31           \\
 freeway            & \textbf{34}       & \textbf{89.47}    & 28.5            & 75.00          & 32.59                &85.76                     &\best{34}         &\textbf{89.47}     &\textbf{34} &\textbf{89.47}           \\
 frostbite          & 11238.4           & 2.46              & 206400          & 45.37          &\textbf{541280.88}    &\textbf{119.01}           &10485             &2.29               &11330     &2.48            \\
 gopher             & 122196            & 34.37             & 113400          & 31.89          & 117777.08            &33.12                     &\best{488830}     &\textbf{137.71}    &473560    &133.41          \\
 gravitar           & 6750              & 4.04              & 14200           & 8.62           &\textbf{19213.96}     &\textbf{11.70}            &5905              &3.52               &5915      &3.53    \\
 hero               & 37030.4           & 3.60              & 69400           & 6.84           &\textbf{114736.26}    &\textbf{11.38}            &38330             &3.73               &38225     &3.72    \\
 ice hockey         & \textbf{71.56}    & \textbf{175.34}   &-4.1             & 15.04          & 63.64                &158.56                    &44.92             &118.94             &\textbf{47.11}  &\textbf{123.54}\\
 jamesbond          & 23266             & 51.05             & 26600           & 58.37          & 135784.96            &200.00                    &594500            &200.00             &\textbf{620780} &\textbf{200.00}   \\
 kangaroo           & 14112             & 0.99              & \textbf{35100}  & 2.46           &24034.16              &1.68                      &14500             &1.01               &14636           &1.02\\
 krull              & 145284.8          & 140.18            & 127400          & 122.73         & 251997.31            &200.00                    &97575             &93.63              &\textbf{594540} &\textbf{200.00}      \\
 kung fu master     & 200176            & 20.00             & 212100          & 21.19          & 206845.82            &20.66                     &140440            &14.02              &\textbf{1666665}&\textbf{166.68}\\
 montezuma revenge  & 2504              & 0.21              & \textbf{10400}  & 0.85           &9352.01               &0.77                      &3000              &0.25               &2500            &0.21   \\
 ms pacman          & 29928.2           & 10.22             & 40800           & 13.97          & \textbf{63994.44}    &\textbf{21.98}            &11536             &3.87               &11573           &3.89 \\
 name this game     & 45214.8           & 187.21            & 23900           & 94.24          &\textbf{54386.77}     &\textbf{200.00}           &34434             &140.19             &36296           &148.31\\
 phoenix            & 811621.6          & 20.20             & 959100          & 23.88          &908264.15             &22.61                     &894460            &22.27              &\textbf{959580} &\textbf{23.89}\\
 pitfall            & \textbf{0}        & \textbf{0.20}     & 7800            & 7.03           &\textbf{18756.01}     &\textbf{16.62}            &0                 &0.20               &-4.3            &0.20\\
 pong               & \textbf{21}       & \textbf{100.00}   & 19.6            & 96.64          & 20.67                &99.21                     &\best{21}         &\textbf{100.00}    &\textbf{21}     &\textbf{100.00}\\
 private eye        & 300               & 0.27              & \textbf{100000} & 98.23          & 79716.46             &78.30                     &15100             &14.81              &15100           &14.81\\
 qbert              & 161000            & 6.70              & 451900          & 18.82          &\textbf{580328.14}    &\textbf{24.18}            &27800             &1.03               &28657           &1.19\\
 riverraid          & 34076.4           & 3.28              & 36700           & 3.54           & \textbf{63318.67}    &\textbf{6.21}             &28075             &2.68               &28349           &2.70\\
 road runner        & 498660            & 24.47             & 128600          & 6.31           & 243025.8             &11.92                     &\best{878600}     &\textbf{43.11}     &999999          &49.06\\
 robotank           & \textbf{132.4}    & \textbf{176.42}   & 9.1             & 9.35           &127.32                &169.54                    &108               &143.63             &113.4           &150.68\\
 seaquest           & 999991.84         & 100.00            & \textbf{1000000}& 100.00         &999997.63             &100.00                    &943910	        &94.39              &\textbf{1000000}&\textbf{100.00}\\
 skiing             & -29970.32         & -93.10            & -22977.9        & -42.53         & \textbf{-4202.6}     &\textbf{93.27}            &-6774             &74.67              &-6025	         &86.77\\
 solaris            & 4198.4            & 2.69              & 4700            & 3.14           & \textbf{44199.93}    &\textbf{38.99}            &11074             &8.93               &9105            &7.14\\
 space invaders     & 55889             & 8.97              & 43400           & 6.96           & 48680.86             &7.81                      &140460            &22.58              &\textbf{154380} &\textbf{24.82}\\
 star gunner        & 521728            & 200.00            & 414600          & 200.00         &\textbf{839573.53}    &\textbf{200.00}           &465750            &200.00             &677590          &200.00\\
 surround           & \textbf{9.96}     & \textbf{101.84}   & -9.6            & 2.04           & 9.5                  &99.49                     &-7.8              &11.22              &2.606           &64.32\\
 tennis             & \textbf{24}       & \textbf{106.70}   & 10.2            & 75.89          & 23.84                &106.34                    &\textbf{24}       &\textbf{106.70}    &\textbf{24}     &\textbf{106.70}\\
 time pilot         & 348932            & 200.00            & 344700          & 200.00         &405425.31             &200.00                    &216770            &200.00             &\textbf{450810} &\textbf{200.00}\\
 tutankham          & 393.64            & 7.11              & 191.1           & 3.34           & \textbf{2354.91}     &\textbf{43.62}            &423.9             &7.68               &418.2           &7.57\\
 up n down          & 542918.8          & 200.00            & 620100          & 200.00         & 623805.73            &200.00                    &\best{986440}     &\textbf{200.00}    &966590          &200.00\\
 venture            & 1992              & 5.12              & 1700            & 4.37           &\textbf{2623.71}      &\textbf{6.74}             &2000              &5.14               &2000            &5.14\\
 video pinball      & 483569.72         & 0.54              & 965300          & 1.08           &\textbf{992340.74}    &\textbf{1.11}             &925830            &1.04               &978190          &1.10\\
 wizard of wor      & 133264            & 33.62             & 106200          & 26.76          &\textbf{157306.41}    &\textbf{39.71}            &64439             &16.18              &63735           &16.00\\
 yars revenge       & 918854.32         & 6.11              & 986000          & 6.55           &\textbf{998532.37}    &\textbf{6.64}             &972000            &6.46               &968090          &6.43\\
 zaxxon             & 181372            & 200.00            & 111100          & 132.75         &\textbf{249808.9}     &\textbf{200.00}           &109140            &130.41             &216020          &200.00\\
\hline
MEAN SABER(\%)        &                   & 60.43             &                 &  50.47         &                      & \textbf{76.26}  &                  & \GDIImeanSABER &      & \GDIHmeanSABER \\
Learning Efficiency &    & 6.04E-11 &         &  1.44E-11  &        & 7.63E-12 &      & 5.90E-09       &      & \textbf{\GDIHmeanSABERle} \\
\hline
MEDIAN SABER(\%)      &                   & 33.62             &                 & 21.19          &                      & 43.62  &                  & \GDIImedianSABER &      & \textbf{\GDIHmedianSABER}  \\
Learning Efficiency &    & 3.36E-11 &         & 6.05E-12  &        & 4.36E-12 &      &2.27E-09      &      & \textbf{\GDIHmedianSABERle}\\
\hline
HWRB                  &                   & 15       &                 & 9              &                      & 18     &                  & \GDIIHWRB &                  & \textbf{\GDIHHWRB}\\
\bottomrule
\end{tabular}
\caption{Score table of SOTA  10B+ model-free algorithms on SABER.}
\end{center}
\end{table}
\clearpage

\begin{table}[!hb]
\scriptsize
\begin{center}
\setlength{\tabcolsep}{1.0pt}
\begin{tabular}{ c c c c c c c c c c c }
\toprule
 Games              & MuZero         & SABER(\%)      & DreamerV2 & SABER(\%)    & SimPLe             & SABER(\%)          & GDI-I$^3$     & SABER(\%) & GDI-H$^3$ & SABER(\%)\\
\midrule
Scale               & 20B            &              & 200M      &            & 1M               &                  & 200M     & &  200M   & \\
\midrule
 alien              & \textbf{741812.63}      & \textbf{200.00  }     &3483       & 1.29     &616.9     & 0.15    & 43384                       &  17.15      &48735    &19.27   \\
 amidar             & \textbf{28634.39 }      & \textbf{27.49   }  &2028       & 1.94     &74.3      & 0.07    & 1442                           &  1.38       &1065     &1.02          \\
 assault            & \textbf{143972.03}      & \textbf{200.00 }     &7679       & 88.51       &527.2     & 3.62       & 63876                  &  200.00     &97155     &200.00    \\
 asterix            & 998425         & 99.84        &25669      & 2.55     &1128.3    & 0.09       & 759910                     &  75.99      &\textbf{999999}            &\textbf{100.00}    \\
 asteroids          & 678558.64               & 6.45        &3064                & 0.02    &793.6              & 0.00           &751970         & 7.15   &\textbf{760005}            &\textbf{7.23}   \\
 atlantis           & 1674767.2               & 15.69           &989207              & 9.22       &20992.5            & 0.08    &3803000    & 35.78  &\textbf{3837300}           &\textbf{36.11} \\
 bank heist         & 1278.98                 & 1.54        &1043                & 1.25    &34.2               & 0.02    &\best{1401}           & \best{1.69  } &1380              &1.66   \\
 battle zone        & \textbf{848623         }& \textbf{105.95}      &31225      & 3.87     &4031.2    & 0.47       & 478830                    &  59.77        &824360            &102.92  \\
 beam rider         & \textbf{454993.53}      & \textbf{45.48}      &12413      & 1.21     &621.6     & 0.03    & 162100                        &  16.18        &422390            &42.22  \\
 berzerk            & \textbf{85932.6        }& \textbf{8.11}      &751        & 0.06     &N/A       & N/A     & 7607                           &  0.71         &14649             &1.37  \\
 bowling            & \textbf{260.13         }& \textbf{85.60}      &48         & 8.99     &30        & 2.49    & 202                           &  64.57        &205.2             &65.76 \\
 boxing             & \textbf{100}                     & \textbf{100.00}       &87                  & 86.99       &7.8         & 7.71       & \best{100}          & \best{100.00 }&\textbf{100}       &\textbf{100.00}\\
 breakout           & \textbf{864}                     & \textbf{100.00}          &350                 & 40.39       &16.4     & 1.70       & \best{864}          & \best{100.00} &\textbf{864}     &100.00\\
 centipede          & \textbf{1159049.27}     & \textbf{89.02}     &6601       & 0.35     &N/A       & N/A     & 155830                         & 11.83         &195630    &14.89\\
 chopper command    & 991039.7                & 99.10     &2833                & 0.20     & 979.4             & 0.02    & \best{999999}         & \best{100.00} &\textbf{999999}    &\textbf{100.00}\\
 crazy climber      & \textbf{458315.4}       & \textbf{200.00    }  &141424     & 62.47       & 62583.6  & 24.77   & 201000                    & 90.96         &241170    &110.17\\
 defender           & 839642.95               & 13.93      & N/A                & N/A        & N/A               & N/A      & 893110     &14.82  &\textbf{970540}    &\textbf{16.11}\\
 demon attack       & 143964.26               & 9.24      & 2775              &0.17      & 208.1             & 0.00    & 675530         & 43.40  &\textbf{787985}   &\textbf{50.63}\\
 double dunk        & 23.94          & 107.42     & 22        &102.53        & N/A      & N/A                                               & \textbf{24}& \textbf{107.58} &\textbf{24}&\textbf{107.58} \\
 enduro             & 2382.44                 & 25.08       & 2112              &22.23         & N/A               & N/A     & \best{14330}    & \best{150.84}  &14300     &150.53\\
 fishing derby      & \textbf{91.16}          & \textbf{112.39     }              &93.24          &200.00      &-90.7     & 0.61    & 59        & 92.89         &65        &96.31\\
 freeway            & 33.03                   & 86.92       & \textbf{34}                 &\textbf{89.47}          &16.7               & 43.95   & \best{34}      & \best{89.47}  &\textbf{34}        &\textbf{89.47}\\
 frostbite          & \textbf{631378.53}      & \textbf{138.82}     & 15622    &3.42       &236.9     & 0.04    & 10485                        & 2.29           &11330     &2.48\\
 gopher             & 130345.58               & 36.67      & 53853             &15.11         &596.8              & 0.10       & \best{488830} & \best{137.71}  &473560    &133.41\\
 gravitar           & \textbf{6682.7     }    & \textbf{4.00    }  & 3554     &2.08      &173.4     & 0.00    & 5905                           & 3.52           &5915      &3.53\\
 hero               & \textbf{49244.11}       & \textbf{4.83    }  & 30287    &2.93      &2656.6    & 0.16        & 38330                      & 3.73           &38225     &3.72\\
 ice hockey         & \textbf{67.04      }    & \textbf{165.76  }  & 29        &85.17         &-11.6     & -0.85       &44.92              &118.94        &47.11           &123.54\\
 jamesbond          & 41063.25                & 90.14     & 9269              &20.30         &100.5     & 0.16    &594500              &200.00  &\textbf{620780}          &\textbf{200.00}\\
 kangaroo           & \textbf{16763.6        }& \textbf{1.17}  & 11819     &0.83      &51.2      & 0.00    & 14500                              & 1.01          &14636           &1.02\\
 krull              & 269358.27      & 200.00 & 9687     &7.89       &2204.8    & 0.59    & 97575                        & 93.63          &\textbf{594540}          &\textbf{200.00}\\
 kung fu master     & \textbf{204824         }& \textbf{20.46}  & 66410    &6.62       &14862.5   & 1.46    & 140440                           & 14.02          &1666665          &166.68\\
 montezuma revenge  & 0                       & 0.00         & 1932              &0.16       &N/A       & N/A     & \best{3000}                & \best{0.25  }  &2500            &0.21\\
 ms pacman          & \textbf{243401.1 }      & \textbf{83.89   }  & 5651     &1.84       &1480      & 0.40    & 11536                         & 3.87           &11573           &3.89\\
 name this game     & \textbf{157177.85}      & \textbf{200.00  }  & 14472    &53.12          &2420.7    & 0.56    & 34434                     & 140.19         &36296           &148.31\\
 phoenix            & \textbf{955137.84}      & \textbf{23.78   }  & 13342     &0.31       &N/A       & N/A     & 894460                        & 22.27         &959580          &23.89\\
 pitfall            & \textbf{0}              & \textbf{0.20}               & -1                 &0.20       &N/A       & N/A     & \best{0}             & \best{0.20}   &-4.3            &0.20\\
 pong               & \textbf{21}             & \textbf{100.00}             & 19                 &95.20          & 12.8     & 80.34   & \best{21}        & \best{100.00} &\textbf{21}   &\textbf{100.00}\\
 private eye        & \textbf{15299.98 }      & \textbf{15.01}     & 158       &0.13       & 35       & 0.01    & 15100                         & 14.81         &15100           &14.81\\
 qbert              & \textbf{72276          }& \textbf{3.00}      & 162023    &6.74       & 1288.8   & 0.05    & 27800                         & 1.15          &28657           &1.19\\
 riverraid          & \textbf{323417.18}      & \textbf{32.25}     & 16249    &1.49       & 1957.8   & 0.06    & 28075                         & 2.68           &28349           &2.70\\
 road runner        & 613411.8                & 30.10              & 88772             &4.36       & 5640.6   & 0.28       & 878600     & 43.11   &\textbf{999999}          &\textbf{49.06}\\
 robotank           & \textbf{131.13}         & \textbf{174.70}    & 65        &85.09          & N/A      & N/A     & 108                       & 143.63        &113.4           &150.68\\
 seaquest           & 999976.52               & 100.00             & 45898             &4.58       & 683.3             & 0.06 &943910	             &94.39&\textbf{1000000}          &\textbf{100.00}\\
 skiing             & \textbf{-29968.36}      & \textbf{-93.09}      & -8187    &64.45          & N/A      & N/A     & -6774                   & 74.67          &-6025	          &86.77\\
 solaris            & 56.62                   & -1.07              & 883                &-0.32          & N/A               & N/A   & \best{11074}  & \best{8.93 }&9105            &7.14\\
 space invaders     & 74335.3                 & 11.94                 & 2611               &0.40       & N/A               & N/A    & 140460 & 22.58&\textbf{154380}          &\textbf{24.82}    \\
 star gunner        &549271.7       & 200.00     & 29219    &37.21          & N/A      & N/A     & 465750      & 200.00                         &\textbf{677590}          &\textbf{200.00}\\
 surround           & \textbf{9.99       }    & \textbf{101.99}     & N/A       &N/A         & N/A      & N/A     & -8          & 11.22                            &2.606           &64.32\\
 tennis             & 0       & 53.13      & 23        &104.46        & N/A      & N/A     & \textbf{24}          & \textbf{106.70}       &\textbf{24}           &\textbf{106.70}\\
 time pilot         & \textbf{476763.9}       & \textbf{200.00}         & 32404    &46.71         & N/A      & N/A     & 216770      & 200.00    &450810          &200.00     \\
 tutankham          & \textbf{491.48     }    & \textbf{8.94}   & 238       &4.22         & N/A      & N/A     & 424         & 7.68              &418.2           &7.57\\
 up n down          & 715545.61               & 200.00            & 648363            &200.00        & 3350.3            & 3.42   & \best{986440}        & \best{200.00} &966590        &200.00    \\
 venture            & 0.4                     & 0.00        & 0                  &0.00       & N/A               & N/A     & \best{2000}          & \best{5.23}   &2000            &5.14    \\
 video pinball      & \textbf{981791.88}      & \textbf{1.10}      & 22218    &0.02     & N/A      & N/A     & 925830      & 1.04                                &978190          &1.10\\
 wizard of wor      & \textbf{197126         }& \textbf{49.80}      & 14531    &3.54     & N/A      & N/A     & 64439       & 16.14                             &63735           &16.00\\
 yars revenge       & 553311.46               & 3.67      & 20089             &0.11     & 5664.3            & 0.02    & \best{972000}        & \best{6.46}      &968090          &6.43\\
 zaxxon             & \textbf{725853.9}       & \textbf{200.00}      & 18295    &21.83          & N/A      & N/A     & 109140      & 130.41                     &216020          &200.00\\
\hline
MEAN SABER(\%) &               &\textbf{71.94}  &         &  27.22 &                                          & 4.67  &     & \GDIImeanSABER &      &\GDIHmeanSABER\\
Learning Efficiency &     &  3.60E-11 &         & 1.36E-09  &        &\textbf{ 4.67E-08}&      & 5.90E-09       &      & \GDIHmeanSABERle\\
\hline
MEDIAN SABER(\%) &            & 49.8   &         & 4.22  &                                          & 0.13  &     & \GDIImedianSABER &      & \textbf{\GDIHmedianSABER}  \\
Learning Efficiency &     & 2.49E-11 &         & 2.11E-10  &        & 1.60E-09&      &2.27E-09      &      & \textbf{\GDIHmedianSABERle}\\
\hline
HWRB &            & 19   &         & 3    &                                          & 0  &      & \GDIIHWRB &                  & \textbf{\GDIHHWRB}\\
\bottomrule
\end{tabular}
\caption{Score table of  SOTA  model-based algorithms on SABER.}
\end{center}
\end{table}
\clearpage

\begin{table}[!hb]
\scriptsize
\begin{center}
\setlength{\tabcolsep}{1.0pt}
\begin{tabular}{ c  c c   c c c c c c}
\toprule
 Games           & Muesli              & SABER(\%)          & Go-Explore              & SABER(\%)                     & GDI-I$^3$ & SABER(\%)        & GDI-H$^3$ & SABER(\%)         \\
\midrule
Scale            & 200M           &           & 10B                     &                             & 200M      &        & 200M      &           \\
\midrule    
 alien           &139409          &55.30               &\textbf{959312}       &\textbf{200.00}                & 43384             &17.15     &48735             &19.27         \\
 amidar          &\textbf{21653}  &\textbf{20.78}      &19083                 &18.32                          & 1442              &1.38      &1065              &1.02          \\
 assault         &36963           &200.00              &30773                 &200.00                         & 63876             &200.00  &\textbf{97155}&\textbf{200.00}        \\
 asterix         &316210          &31.61               &999500       &99.95                 & 759910            &75.99      &\textbf{999999}            &\textbf{100.00}     \\
 asteroids       &484609          &4.61                &112952                &1.07                           & 751970     &7.15&\textbf{760005}            &\textbf{7.23}           \\
 atlantis        &1363427         &12.75               &286460                &2.58                           & 3803000    &35.78&\textbf{3837300}           &\textbf{36.11}          \\
 bank heist      &1213            &1.46                &\textbf{3668}         &\textbf{4.45  }                & 1401              &1.69        &1380              &1.66      \\
 battle zone     &414107          &51.68               &\textbf{998800}       &\textbf{124.70}                & 478830            &59.77       &824360            &102.92      \\
 beam rider      &288870          &28.86               &371723       &37.15                 & 162100            &16.18       &\textbf{422390}            &\textbf{42.22}      \\
 berzerk         &44478           &4.19                &\textbf{131417}       &\textbf{12.41 }                & 7607              &0.71        &14649             &1.37      \\
 bowling         &191             &60.64               &\textbf{247}           &\textbf{80.86 }                & 202               &64.57      &205.2             &65.76       \\
 boxing          &99              &99.00               &91                     &90.99                          & \best{100}        &\best{100.00} &\textbf{100}       &\textbf{100.00}           \\
 breakout        &791             &91.53               &774                    &89.56                          & \best{864}        &\best{100.00} &\textbf{864}     &\textbf{100.00}          \\
 centipede       &\textbf{869751} &\textbf{66.76}      &613815                &47.07                          & 155830            &11.83        &195630    &14.89  \\
 chopper command &101289       &10.06               &996220                &99.62                          & \best{999999}     &\best{100.00}   &\textbf{999999}    &\textbf{100.00}       \\
 crazy climber   &175322       &78.68               &235600       &107.51                & 201000            &90.96           &\textbf{241170}    &\textbf{110.17} \\
 defender        &629482       &10.43               &N/A                    &N/A                            & 893110     &14.82   &\textbf{970540}    &\textbf{16.11}        \\
 demon attack    &129544       &8.31                &239895                 &15.41                          & 675530     &43.40   &\textbf{787985}   &\textbf{50.63}       \\
 double dunk     &-3           &39.39               &\textbf{24}            &\textbf{107.58}                         & \best{24}         &\best{107.58}  &\textbf{24}        &\textbf{107.58}         \\
 enduro          &2362         &24.86               &1031                   &10.85                          & \best{14330}      &\best{150.84}  &14300     &150.53         \\
 fishing derby   &51           &87.71               &\textbf{67}            &\textbf{97.54 }                & 59                &92.89          &65        &96.31  \\
 freeway         &33           &86.84               &\textbf{34}            &\textbf{89.47 }                & \best{34}         &\best{89.47}   &\textbf{34}        &\textbf{89.47}         \\
 frostbite       &301694       &66.33               &\textbf{999990}       &\textbf{200.00}                & 10485             &2.29            &11330     &2.48\\
 gopher          &104441       &29.37               &134244                &37.77                          & \best{488830}     &\best{137.71}   &473560    &133.41       \\
 gravitar        &11660        &7.06                &\textbf{13385}        &\textbf{8.12}                  & 5905              &3.52            &5915      &3.53\\
 hero            &37161        &3.62                &37783                  &3.68                           &38330      &3.73    &\textbf{38225}     &\textbf{3.72}       \\
 ice hockey      &25           &76.69               &33                     &93.64                          &44.92              &118.94       &\textbf{47.11}           &\textbf{123.54}          \\
 jamesbond       &19319        &42.38               &200810                &200.00                         &594500              &200.00   &\textbf{620780}          &\textbf{200.00}     \\
 kangaroo        &14096        &0.99                &\textbf{24300}        &\textbf{1.70}                  & 14500             &1.01            &14636              &1.02\\
 krull           &34221        &31.83               &63149                 &60.05                          & 97575      &93.63    &\textbf{594540}    &\textbf{200.00}       \\
 kung fu master  &134689       &13.45               &24320                 &2.41                           & 140440     &14.02    &\textbf{1666665}          &\textbf{166.68}        \\
 montezuma revenge  &2359      &0.19                &\textbf{24758}        &\textbf{2.03}                  & 3000              &0.25            &2500            &0.21\\
 ms pacman          &65278     &22.42               &\textbf{456123}       &\textbf{157.30}                & 11536             &3.87            &11573           &3.89\\
 name this game     &105043    &200.00              &\textbf{212824}       &\textbf{200.00}                & 34434             &140.19          &36296           &148.31\\
 phoenix        &805305        &20.05               &19200                 &0.46                           & 894460     &22.27    &\textbf{959580}          &\textbf{23.89}\\
 pitfall        &0             &0.20                &\textbf{7875}          &\textbf{7.09   }               & 0                 &0.2            &-4.3            &0.20 \\
 pong           &20            &97.60               &\textbf{21}            &\textbf{100.00 }               & \best{21}         &\best{100}     &\textbf{21}              &\textbf{100.00}       \\
 private eye    &10323         &10.12               &\textbf{69976}        &\textbf{68.73  }               & 15100             &14.81           &15100           &14.81  \\
 qbert          &157353        &6.55                &\textbf{999975}       &\textbf{41.66  }               & 27800             &1.15            &28657           &1.19\\
 riverraid      &\textbf{47323}&\textbf{4.60}       &35588                 &3.43                           & 28075             &2.68            &28349           &2.70\\
 road runner    &327025        &16.05               &999900        &49.06                 & 878600            &43.11          &\textbf{999999}          &\textbf{49.06}\\
 robotank       &59            &76.96               &\textbf{143}           &\textbf{190.79 }               & 108               &143.63         &113.4           &150.68\\
 seaquest       &815970        &81.60               &539456                 &53.94                 &943910	             &94.39   &\textbf{1000000}          &\textbf{100.00}\\
 skiing         &-18407        &-9.47               &\textbf{-4185}        &\textbf{93.40  }               & -6774             &74.67           &-6025	         &86.77\\
 solaris        &3031          &1.63                &\textbf{20306}        &\textbf{17.31  }               & 11074             &8.93            &9105            &7.14\\
 space invaders &59602         &9.57                &93147                 &14.97                          & 140460     &22.58    &\textbf{154380}          &\textbf{24.82}\\
 star gunner    &214383        &200.00              &609580      &200.00                         & 465750     &200.00&\textbf{677590}          &\textbf{200.00}     \\
 surround       &\textbf{9}    &\textbf{96.94}      &N/A                    &N/A                            & -8         &11.22                 &2.606           &64.32\\
 tennis         &12            &79.91               &\best{24}              &\best{106.7}                   & \best{24}         &\best{106.70   }&\textbf{24}           &\textbf{106.70}            \\
 time pilot     &359105 &200.00   &183620                &200.00                         & 216770     & 200.00               &\textbf{450810}          &\textbf{200.00}\\
 tutankham      &252           &4.48                &\textbf{528}           &\textbf{9.62}                  & 424               &7.68           &418.2           &7.57\\
 up n down      &649190        &200.00              &553718                &200.00                         & \best{986440}     &\best{11.9785}  &966590        &200.00         \\
 venture        &2104          &5.41                &\textbf{3074}         &\textbf{7.90}                  & 2035              &5.23            &2000            &5.14\\
 video pinball  &685436        &0.77                &\textbf{999999}       &\textbf{1.12}                  & 925830            &1.04            &978190          &1.10\\
 wizard of wor  &93291         &23.49               &\textbf{199900}       &\textbf{50.50}                 & 64293             &16.14           &63735           &16.00\\
 yars revenge   &557818        &3.70                &\textbf{999998}       &\textbf{6.65}                  & 972000            &6.46            &968090          &6.43\\
 zaxxon         &65325         &78.04               &18340                 &21.88                          & 109140         &130.41   &\textbf{216020} &\textbf{200.00}        \\
\hline    
MEAN SABER(\%)  &              & 48.74              &                       & \textbf{71.80}                &                   & \GDIImeanSABER &      &\GDIHmeanSABER\\
Learning Efficiency &    & 2.43E-09 &                       & 7.18E-11                      &      & 3.08E-09       &      & \textbf{\GDIHmeanSABERle}\\
\hline
MEDIAN SABER(\%)&              &  24.86             &                       &50.5                 &                   & \GDIImedianSABER &      & \textbf{\GDIHmedianSABER}  \\
Learning Efficiency &    & 1.24E-09 &                       & 5.05E-11                     &      &1.78E-09      &      & \textbf{\GDIHmedianSABERle}\\
\hline
HWRB           &                       & 5  &                       & 15              &            & \GDIIHWRB &                  & \textbf{\GDIHHWRB}\\
\bottomrule
\end{tabular}
\caption{Score table of other SOTA  algorithms on SABER.}
\end{center}
\end{table}

\clearpage

\end{appendix}

\end{document}